\documentclass[11pt,a4paper]{article}
\usepackage[a4paper, left=1in, right=1in, top=1in, bottom=1in, includehead, includefoot, foot = 0pt, head=0pt]{geometry} 
\parskip=\medskipamount 

\usepackage{booktabs} 
\usepackage{enumitem} 

\usepackage{natbib}

\bibpunct{(}{)}{;}{a}{,}{,}
\newcommand{\biblist}{
\bibliographystyle{apalike}
\bibliography{References}  
}
\usepackage[hyphens,spaces,obeyspaces]{url}
\usepackage[pdftex,colorlinks,citecolor=blue,urlcolor=blue,linkcolor=red]{hyperref} 
\DeclareUrlCommand\doi{\def\UrlLeft##1\UrlRight{doi:\hspace{0.1cm} \href{http://dx.doi.org/##1}{##1}}\urlstyle{rm}}

\usepackage{graphicx} 
\usepackage[small,bf,hang]{caption}	

\usepackage{tabularx}
\usepackage[export]{adjustbox}

\usepackage{amssymb} 
\usepackage{amsmath} 
\usepackage{accents}
\usepackage{bbm} 
\usepackage{amsbsy} 
\usepackage{multirow} 
\usepackage[dvipsnames]{xcolor} 
\usepackage{mathtools}
\usepackage{eurosym}

\usepackage[ruled, lined, longend]{algorithm2e}
\usepackage{float}
\usepackage{rotating}
\usepackage[caption=false]{subfig}
\usepackage{pdflscape}
\usepackage[defaultcolor=red]{changes}

\usepackage{bbm}
\usepackage{changepage}

\usepackage{authblk} 

\usepackage{fancyhdr}
\bibpunct[, ]{(}{)}{;}{a}{,}{,}

\usepackage{tikz,pgfplots}
\usetikzlibrary{fit, positioning,chains,shapes.arrows,fit}
\usetikzlibrary{shapes,arrows, arrows.meta}
\usetikzlibrary{decorations.pathreplacing}
\usetikzlibrary{calc} 
\usetikzlibrary{matrix} 
\usepackage{pgf}
\usetikzlibrary{decorations.text}
\tikzset{cross/.style={cross out, draw=black, minimum size=2*(#1-\pgflinewidth), inner sep=0pt, outer sep=0pt},
cross/.default={1pt}}
\usetikzlibrary{shapes.misc}
\usepackage{environ}
\makeatletter
\newsavebox{\measure@tikzpicture}
\NewEnviron{scaletikzpicturetowidth}[1]{%
  \def\tikz@width{#1}%
  \begin{lrbox}{\measure@tikzpicture}%
  \BODY
  \end{lrbox}%
  \pgfmathparse{#1/\wd\measure@tikzpicture}%
  \BODY
}
\makeatother
\tikzdeclarecoordinatesystem{page}{
    \parsecomma#1\endparsecomma
    \pgfpointanchor{current page}{north east}
    \pgf@xc=\pgf@x%
    \pgf@yc=\pgf@y%
    \pgfpointanchor{current page}{south west}
    \pgf@xb=\pgf@x%
    \pgf@yb=\pgf@y%
    \pgfmathparse{(\pgf@xc-\pgf@xb)/2.*\page@x+(\pgf@xc+\pgf@xb)/2.}
    \expandafter\pgf@x\expandafter=\pgfmathresult pt
    \pgfmathparse{(\pgf@yc-\pgf@yb)/2.*\page@y+(\pgf@yc+\pgf@yb)/2.}
    \expandafter\pgf@y\expandafter=\pgfmathresult pt
}
\makeatother
\usetikzlibrary{arrows,automata}
\usepackage{ragged2e}
\definecolor{arrowcolor}{RGB}{201,216,232}
\definecolor{circlecolor}{RGB}{79,129,189}
\colorlet{textcolor}{white}
\colorlet{bordercolor}{white}

\pgfdeclarelayer{background}
\pgfsetlayers{background,main}

\definecolor{airforceblue}{rgb}{0.36, 0.54, 0.66}
\definecolor{forestgreen}{rgb}{0.13, 0.55, 0.13}\definecolor{fulvous}{rgb}{0.86, 0.52, 0.0}
\definecolor{gray}{rgb}{0.5, 0.5, 0.5}
\definecolor{bistre}{rgb}{0.24, 0.17, 0.12}\definecolor{bostonuniversityred}{rgb}{0.8, 0.0, 0.0}
\definecolor{purpleheart}{rgb}{0.41, 0.21, 0.61}
\definecolor{lightsalmonpink}{rgb}{1.0, 0.6, 0.6}\definecolor{arrowcolor}{rgb}{0.92, 0.92, 0.92}
\tikzset{
inner/.style={
  on chain,
  circle,
  inner sep=4pt,
  fill=circlecolor,
  line width=1.5pt,
  draw=bordercolor,
  text width=1.2em,
  align=center,
  text height=1.25ex,
  text depth=0ex
},
on grid
}

\newcommand\drawarrow{

\node[on chain] (f) {};
\begin{pgfonlayer}{background}
\node[
  inner sep=10pt,
  single arrow,
  single arrow head extend=0.6cm,
  draw=none,
  fill=arrowcolor,
  fit= (c1) (f)
] (arrow) {};
\fill[white] 
  (arrow.before tail) -- (c1|-arrow.west) -- (arrow.after tail) -- cycle;
\end{pgfonlayer}
}

\numberwithin{equation}{section}

\begin{document}
\title{Machine learning in an expectation-maximisation framework for nowcasting}

\author[1,3,*]{Paul Wilsens}
\author[1,2,3,4]{Katrien Antonio}
\author[1,4,5]{Gerda Claeskens}
\affil[1]{Faculty of Economics and Business, KU Leuven, Belgium.}
\affil[2]{Faculty of Economics and Business, University of Amsterdam, The Netherlands.}
\affil[3]{Leuven Research Center on Insurance and Financial Risk Analysis, KU Leuven, Belgium.}
\affil[4]{Leuven Statistics Research Center, KU Leuven, Belgium.}
\affil[5]{Research Centre for Operations Research and Statistics, KU Leuven, Belgium.}
\affil[*]{Corresponding author: \href{mailto:paul.wilsens@kuleuven.be}{paul.wilsens@kuleuven.be}}
\date{\vspace{-1.25cm}} 
\maketitle
\thispagestyle{empty}

\begin{abstract}
\noindent Decision making often occurs in the presence of incomplete information, leading to the under- or overestimation of risk. Leveraging the observable information to learn the complete information is called nowcasting. In practice, incomplete information is often a consequence of reporting or observation delays. In this paper, we propose an expectation-maximisation (EM) framework for nowcasting that uses machine learning techniques to model both the occurrence as well as the reporting process of events. We allow for the inclusion of covariate information specific to the occurrence and reporting periods as well as characteristics related to the entity for which events occurred. We demonstrate how the maximisation step and the information flow between EM iterations can be tailored to leverage the predictive power of neural networks and (extreme) gradient boosting machines (XGBoost). With simulation experiments, we show that we can effectively model both the occurrence and reporting of events when dealing with high-dimensional covariate information. In the presence of non-linear effects, we show that our methodology outperforms existing EM-based nowcasting frameworks that use generalised linear models in the maximisation step. Finally, we apply the framework to the reporting of Argentinian Covid-19 cases, where the XGBoost-based approach again is most performant.

\end{abstract}

\vspace{-0.25cm}
maximisation algorithm, reserving
\vspace{-0.25cm}
\section*{Statements and declarations}
\paragraph{Data and code availability statement:} Data and code are available on \url{https://github.com/PaulWilsens/nowcasting-EM-machinelearning}.
\vspace{-0.25cm}
\paragraph{Funding statement and acknowledgements:} The authors gratefully acknowledge funding from the FWO and Fonds De La Recherche
Scientifique - FNRS (F.R.S.-FNRS) under the Excellence of Science (EOS) program, project
ASTeRISK Research Foundation Flanders [grant number 40007517]. Katrien Antonio gratefully
acknowledges support from the Chaire DIALog sponsored by CNP Assurances and the FWO network W001021N. 

\vspace{-0.25cm}
\paragraph{Conflict of interest disclosure:} The authors declare no conflict of interest.

\pagebreak


\section{Introduction} \label{section:introduction}
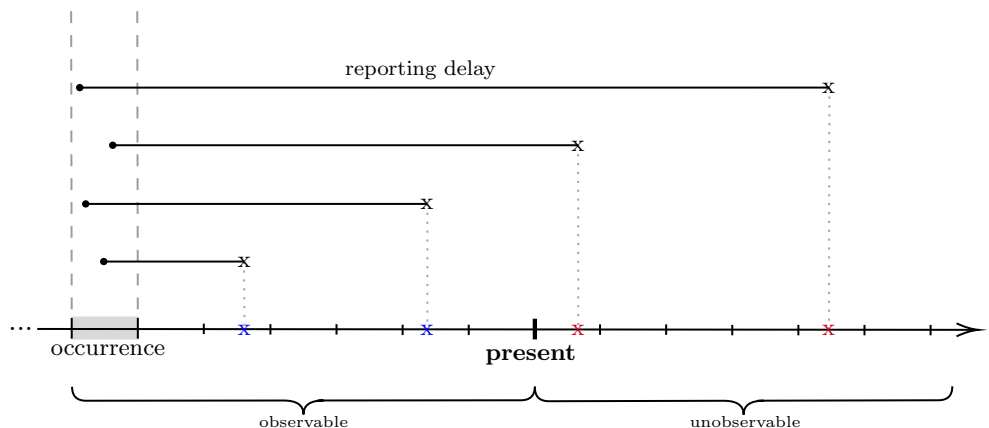
\begin{figure}[H]
    \centering

\tikzset{every picture/.style={line width=0.75pt}} 

\begin{tikzpicture}[x=0.75pt,y=0.75pt,yscale=-1,xscale=1]

\draw    (22.8,165.8) -- (490.6,166.26) ;
\draw [shift={(492.6,166.26)}, rotate = 180.06] [color={rgb, 255:red, 0; green, 0; blue, 0 }  ][line width=0.75]    (10.93,-3.29) .. controls (6.95,-1.4) and (3.31,-0.3) .. (0,0) .. controls (3.31,0.3) and (6.95,1.4) .. (10.93,3.29)   ;
\draw    (73,160.26) -- (73,170.66) ;
\draw[ultra thick] (271.2,160.66) -- (271.2,171.06);
\draw    (40,160.26) -- (40,170.66) ;
\draw [color={rgb, 255:red, 0; green, 0; blue, 0 }  ,draw opacity=1 ][fill={rgb, 255:red, 155; green, 155; blue, 155 }  ,fill opacity=1 ]   (105.94,163.14) -- (105.94,168.39) ;
\draw [color={rgb, 255:red, 0; green, 0; blue, 0 }  ,draw opacity=0.39 ] [dash pattern={on 4.5pt off 4.5pt}]  (39.73,5.95) -- (40,160.26) ;
\draw [color={rgb, 255:red, 0; green, 0; blue, 0 }  ,draw opacity=0.39 ] [dash pattern={on 4.5pt off 4.5pt}]  (72.73,5.95) -- (73,160.26) ;
\draw [color={rgb, 255:red, 0; green, 0; blue, 0 }  ,draw opacity=1 ][fill={rgb, 255:red, 155; green, 155; blue, 155 }  ,fill opacity=1 ]   (139.19,163.39) -- (139.19,168.64) ;
\draw [color={rgb, 255:red, 0; green, 0; blue, 0 }  ,draw opacity=1 ][fill={rgb, 255:red, 155; green, 155; blue, 155 }  ,fill opacity=1 ]   (172.19,163.39) -- (172.19,168.64) ;
\draw [color={rgb, 255:red, 0; green, 0; blue, 0 }  ,draw opacity=1 ][fill={rgb, 255:red, 155; green, 155; blue, 155 }  ,fill opacity=1 ]   (205.19,163.64) -- (205.19,168.89) ;
\draw [color={rgb, 255:red, 0; green, 0; blue, 0 }  ,draw opacity=1 ][fill={rgb, 255:red, 155; green, 155; blue, 155 }  ,fill opacity=1 ]   (238.19,163.64) -- (238.19,168.89) ;
\draw [color={rgb, 255:red, 0; green, 0; blue, 0 }  ,draw opacity=1 ][fill={rgb, 255:red, 155; green, 155; blue, 155 }  ,fill opacity=1 ]   (303.73,163.51) -- (303.73,168.76) ;
\draw [color={rgb, 255:red, 0; green, 0; blue, 0 }  ,draw opacity=1 ][fill={rgb, 255:red, 155; green, 155; blue, 155 }  ,fill opacity=1 ]   (336.73,163.51) -- (336.73,168.76) ;
\draw [color={rgb, 255:red, 0; green, 0; blue, 0 }  ,draw opacity=1 ][fill={rgb, 255:red, 155; green, 155; blue, 155 }  ,fill opacity=1 ]   (369.73,163.76) -- (369.73,169.01) ;
\draw  [color={rgb, 255:red, 0; green, 0; blue, 0 }  ,draw opacity=0.15 ][fill={rgb, 255:red, 0; green, 0; blue, 0 }  ,fill opacity=0.15 ] (40.22,160.06) -- (72.97,160.06) -- (72.97,170.66) -- (40.22,170.66) -- cycle ;
\draw    (56,131.9) -- (126,131.9) ;
\draw [shift={(56,131.9)}, rotate = 0] [color={rgb, 255:red, 0; green, 0; blue, 0 }  ][fill={rgb, 255:red, 0; green, 0; blue, 0 }  ][line width=0.75]      (0, 0) circle [x radius= 1.34, y radius= 1.34]   ;
\draw    (47,102.91) -- (217.54,102.91) ;
\draw [shift={(47,102.91)}, rotate = 0] [color={rgb, 255:red, 0; green, 0; blue, 0 }  ][fill={rgb, 255:red, 0; green, 0; blue, 0 }  ][line width=0.75]      (0, 0) circle [x radius= 1.34, y radius= 1.34]   ;
\draw    (60.5,73.4) -- (292.79,73.4) ;
\draw [shift={(60.5,73.4)}, rotate = 0] [color={rgb, 255:red, 0; green, 0; blue, 0 }  ][fill={rgb, 255:red, 0; green, 0; blue, 0 }  ][line width=0.75]      (0, 0) circle [x radius= 1.34, y radius= 1.34]   ;
\draw    (44,44.41) -- (418.2,44.39) ;
\draw [shift={(44,44.41)}, rotate = 360] [color={rgb, 255:red, 0; green, 0; blue, 0 }  ][fill={rgb, 255:red, 0; green, 0; blue, 0 }  ][line width=0.75]      (0, 0) circle [x radius= 1.34, y radius= 1.34]   ;
\draw [color={rgb, 255:red, 0; green, 0; blue, 0 }  ,draw opacity=0.35 ] [dash pattern={on 0.84pt off 2.51pt}]  (126,131.9) -- (126.29,165.4) ;
\draw [color={rgb, 255:red, 0; green, 0; blue, 0 }  ,draw opacity=0.35 ] [dash pattern={on 0.84pt off 2.51pt}]  (217.54,102.91) -- (217.54,166.15) ;
\draw [color={rgb, 255:red, 0; green, 0; blue, 0 }  ,draw opacity=0.35 ] [dash pattern={on 0.84pt off 2.51pt}]  (292.79,73.4) -- (292.86,165.65) ;
\draw [color={rgb, 255:red, 0; green, 0; blue, 0 }  ,draw opacity=0.35 ] [dash pattern={on 0.84pt off 2.51pt}]  (418.2,44.39) -- (418.02,168.62) ;
\draw   (40.19,194.91) .. controls (40.2,199.58) and (42.53,201.91) .. (47.2,201.91) -- (145.7,201.8) .. controls (152.37,201.79) and (155.7,204.12) .. (155.71,208.79) .. controls (155.7,204.12) and (159.03,201.79) .. (165.7,201.78)(162.7,201.79) -- (264.2,201.68) .. controls (268.87,201.67) and (271.2,199.34) .. (271.19,194.67) ;
\draw   (271.19,195.16) .. controls (271.2,199.83) and (273.54,202.16) .. (278.21,202.15) -- (365.39,201.93) .. controls (372.06,201.92) and (375.4,204.24) .. (375.41,208.91) .. controls (375.4,204.24) and (378.72,201.9) .. (385.39,201.89)(382.39,201.89) -- (472.58,201.67) .. controls (477.25,201.66) and (479.57,199.33) .. (479.56,194.66) ;
\draw [color={rgb, 255:red, 0; green, 0; blue, 0 }  ,draw opacity=1 ][fill={rgb, 255:red, 155; green, 155; blue, 155 }  ,fill opacity=1 ]   (402.73,163.76) -- (402.73,169.01) ;
\draw [color={rgb, 255:red, 0; green, 0; blue, 0 }  ,draw opacity=1 ][fill={rgb, 255:red, 155; green, 155; blue, 155 }  ,fill opacity=1 ]   (435.73,163.76) -- (435.73,169.01) ;
\draw [color={rgb, 255:red, 0; green, 0; blue, 0 }  ,draw opacity=1 ][fill={rgb, 255:red, 155; green, 155; blue, 155 }  ,fill opacity=1 ]   (468.73,163.76) -- (468.73,169.01) ;

\draw (7,163.5) node [anchor=north west][inner sep=0.75pt]   [align=left] {...};
\draw (245,172.4) node [anchor=north west][inner sep=0.75pt]  [font=\footnotesize] [align=left] {\textbf{present}};
\draw (132,208) node [anchor=north west][inner sep=0.75pt]   [align=left] {{\tiny observable}};
\draw (347,208) node [anchor=north west][inner sep=0.75pt]   [align=left] {{\tiny unobservable}};
\draw (121.4,162.3) node [anchor=north west][inner sep=0.75pt]   [align=left] {{\footnotesize \textcolor[rgb]{0,0,1}{x}}};
\draw (288,162.3) node [anchor=north west][inner sep=0.75pt]   [align=left] {{\footnotesize \textcolor[rgb]{0.82,0.01,0.11}{x}}};
\draw (413.1,162.3) node [anchor=north west][inner sep=0.75pt]   [align=left] {{\footnotesize \textcolor[rgb]{0.82,0.01,0.11}{x}}};
\draw (212.6,162.3) node [anchor=north west][inner sep=0.75pt]   [align=left] {{\footnotesize \textcolor[rgb]{0,0,1}{x}}};
\draw (28,172.4) node [anchor=north west][inner sep=0.75pt]  [font=\footnotesize] [align=left] {occurrence};
\draw (175,29) node [anchor=north west][inner sep=0.75pt]  [font=\scriptsize] [align=left] {reporting delay};
\draw (121.4,128) node [anchor=north west][inner sep=0.75pt]  [font=\footnotesize] [align=left] {x};
\draw (212.6,99) node [anchor=north west][inner sep=0.75pt]  [font=\footnotesize] [align=left] {x};
\draw (288,69.5) node [anchor=north west][inner sep=0.75pt]  [font=\footnotesize] [align=left] {x};
\draw (413.1,40.5) node [anchor=north west][inner sep=0.75pt]  [font=\footnotesize] [align=left] {x};

\end{tikzpicture}
    
    \caption{Occurrence and reporting timing of four events that happened in the same past occurrence period (highlighted in grey). Event occurrence is indicated with a black dot, event reporting with an x-mark. The reporting delay is visualised by the black line between a dot and an x-mark. A blue or red x-mark on the time axis indicates whether the occurrence of an event is observable or unobservable at present time, respectively.}
    \label{fig:timeline}
\end{figure}

When an event happens, its occurrence might not be immediately observable. Decision making often happens in dynamic environments where only partial information is available. For example, when an epidemic occurs, health instances have to account for all infected cases, while only having access to information about the diagnosed cases. An ill-defined model for the undiagnosed cases can lead to the under- or overestimation of the underlying risk. Leveraging the observed information to model the complete information, which is only partially available, is commonly referred to in the literature as nowcasting, see e.g.~\cite{browning1989nowcasting}. It is not the same as forecasting because we are dealing with events that have already happened or are happening in real-time, but are not (yet) known because of observation or reporting delays. Figure~\ref{fig:timeline} shows four events that have occurred in the same period but are reported at different times. The black dots represent the occurrence of the events, while the x-marks denote their reporting. At present, we have incomplete information since we can only observe events that have already been reported, which are denoted by blue x-marks. In this case, the goal of nowcasting is to learn the occurrence and reporting dynamics based on historical data and event-specific information. This allows for the prediction of unobserved events (denoted by red x-marks) that have already occurred in a specific past occurrence period.

Nowcasting is relevant to different scientific literatures. In \textit{economics}, macroeconomic information is released at different time intervals, e.g.~monthly versus quarterly. As a result of this publication delay, at a given point in time, only the macroeconomic indicators of the latest reporting period are available. In this context, nowcasting often refers to obtaining an estimate of economic growth within the current quarter, measured by the gross domestic product (GDP) \citep{giannone2008nowcasting}. For this purpose, macroeconomic indicators are nowcasted to obtain a realistic estimate of the GDP at any given point in time. Several model types are prevalent in the literature, including  dynamic factor models (DFM), mixed-data sampling approach (MIDAS) models and vector autoregression (VAR) models \citep{stundziene2024future}. Recent studies compare these traditional models with more advanced machine learning approaches like neural networks and support vector machines, see e.g.~\cite{richardson2018nowcasting}. In \textit{meteorology}, nowcasting usually refers to obtaining real-time estimates of weather phenomena over a very short period of time. The delay in receiving information from weather stations causes a partial availability of information. Historically, weather variable extrapolation based on radar imagery has been used to obtain nowcasts of measurements, such as rainfall \citep{browning1989nowcasting}. Recent studies leverage the flexibility of machine learning techniques in obtaining nowcasts. For instance, \cite{meo2024extreme} introduced a transformer-based generative model, which they refer to as NowcastingGPT-EVL, to obtain extreme precipitation nowcasts. \cite{upadhyay2024theoretical} explored next-frame prediction of radar images using computer vision techniques. In \textit{epidemiology}, accurately estimating the number of infections is crucial for decision surveillance systems \citep{menkir2021nowcasting}. However, due to several factors, e.g.~the incubation period as well as logistical reporting delays such as communication lag, the number of diagnosed infections underestimates the true number of infected cases. \Citet{van2019nowcasting} proposed a nowcasting method based on bivariate P-spline smoothing to estimate the number of unreported cases. \cite{hohle2014bayesian} jointly modelled the occurrence and reporting delay process of infected cases using a Bayesian approach. Estimating the number of unreported infected cases is an example of nowcasting events that have occurred but are not yet reported \citep{lawless1994adjustments}. A similar problem is discussed in the \textit{actuarial} literature. A distinction is made between the reported-but-not-settled (RBNS) claims and the incurred-but-not-reported (IBNR) claims. Having an estimation of the IBNR claims is needed for the insurer to calculate the reserve, i.e.~the funds set aside to cover future claim payments. The chain ladder method \citep{mack1993distribution} models the number of IBNR claims by estimating development factors that reflect historical patterns in the reporting of claims. \cite{bastos2019modelling} extended the chain ladder method by specifying a negative binomial distribution for the event counts, to allow for spatiotemporal variation in the event counts. They allow for variation in the reporting delays for each administrative reporting region in (Brazilian) disease surveillance data. Their approach illustrates the similarity between the IBNR problem and the estimation of the number of infected cases in epidemiology. Recently, \cite{buchardt2023estimation} proposed a two-step M-estimator for multistate models that corrects for reporting delays and incomplete event adjudication, and apply their methodology to a portfolio of disability insurance policies.

A common approach to deal with an incomplete data problem such as nowcasting or mixture models, is an expectation-maximisation (EM) algorithm \citep{dempster1977maximum}. An EM algorithm alternates between two steps: the expectation step (E-step) and the maximisation step (M-step). In the E-step, the missing part of the data is estimated based on parameter estimates obtained through likelihood optimisation in the M-step. Since only an estimate is available for the missing data, the M-step optimises an expected likelihood. The idea is that by alternating between both steps, the algorithm converges to the optimal parameter values. \cite{verbelen2022modeling} proposed an expectation-maximisation (EM) framework to jointly model the occurrence and the reporting process of events that occur in a specific period (day, month,..). They rely on generalised linear models (GLMs) in the maximisation step of the EM algorithm. They assume a Poisson distribution for the number of events in each occurrence period and a multinomial distribution to model the corresponding reporting delay structure. Subsequently, the parameters governing these distributions, referred to as the occurrence intensities and the reporting probabilities, are modelled as a function of covariate information related to the occurrence and reporting periods. For events that occurred on a specific day, this may include, for example, indicators of whether that day happened to be in a weekend or was a holiday. Additionally, they applied their methodology to a general liability insurance portfolio and showed how the parameter estimates of the chain ladder approach \citep{mack1993distribution} can be obtained using an EM approach. \cite{crevecoeur2023bridging} proposed a hierarchical reserving model for the development of individual claims over discrete time. To model the occurrence and reporting of claims, they extended the EM framework (with GLMs) from \cite{verbelen2022modeling} to model occurrence intensities and reporting probabilities at the level of the individual policyholder (for a given occurrence period) in an actuarial setting. They allow for the inclusion of policyholder-specific information in addition to date-specific information. Their extension allows the modelling framework to account for individual policy characteristics, such as coverage details or demographic information, alongside the temporal covariates related to the occurrence and reporting periods. Another EM framework for reserving that also relies on GLMs and allows for the inclusion of (granular) policyholder information was proposed by \cite{wang2022stochastic}. 

Generalised linear models are an effective model choice within an EM scheme because of their fast training speed, and because the properties of the EM algorithm ensure that the training likelihood increases with each iteration of the EM algorithm \citep{dempster1977maximum}. GLMs either allow for exact optimisation of a (log-)likelihood or there exist mathematical results that guarantee convergence to the global optimum in the case of numerical optimisation. For machine learning models, which rely on gradient-descent based methods, this is not the case. They can get stuck in local optima and are therefore prone to numerical instability via oscillation in the training likelihood, depending on e.g.~the hyperparameter choices and the initialisation values for the trainable parameters. Consequently, it is not straightforward to stabilise the training process for such models, as e.g.~discussed for neural networks in \cite{delong2021gamma}. Different applications, which are suitable for an EM algorithm, have recently been tackled with machine learning. In mixture modelling, latent variables describe to which mixture component an observation belongs. This is a missing data problem since the latent variables are unobservable. Consequently, an EM algorithm is typically applied to calibrate the mixture component's parameters, given a known number of components. \cite{delong2021gamma} used an EM algorithm with neural networks, so-called gamma mixture density networks (GMDNs), in the maximisation step to fit a mixture of gamma distributions. They proposed two approaches to implement the algorithm: the EM network boosting algorithm and the EM forward network algorithm. \cite{hou2025insurance} modelled insurance loss data using a mixture model and proposed an expectation-boosting (EB) algorithm, which relies on gradient boosting machines with regression trees in the maximisation step. They apply their algorithm to a zero-inflated Poisson model, i.e.~a mixture of unit probability mass at zero and a Poisson distribution, and a mixture of Gaussians. \cite{shobha2021clustering} proposed an imputation algorithm for missing values in healthcare datasets that integrates neural networks and decision trees in an EM algorithm. In the framework of latent-state models, EM algorithms are commonly used to model the latent variables. \cite{adam2022gradient} proposed a class of hidden Markov models, called Markov-switching generalized additive models for location, scale and shape (MS-GAMLSS), which use gradient boosting in an EM algorithm to estimate the state-dependent parameters of an observed process.

As discussed above, GLMs are suitable to use in the M-step because of their stability and speed during training. However, they require an explicit specification of the functional relationship between the covariates and the response. In the presence of high-dimensionality within the covariates, capturing non-linear effects is often infeasible because the number of potential interactions and transformations grows rapidly. In this paper, we generalise the GLM-based EM framework for nowcasting of \cite{verbelen2022modeling} and \cite{crevecoeur2023bridging} by relaxing the requirement of an explicit functional form. Specifically, we use machine learning models, i.e.~an XGBoost model or a neural network, within the maximisation step of the EM algorithm. For both models, we explore how their structure and the information flow between EM iterations can be tailored to ensure stable training for the occurrence and reporting process similar to the GLM-based approach. By not requiring an explicit functional form (in terms of covariate effects used in the predictor), our framework can handle high-dimensional covariate information without the need to predefine linear or non-linear relationships, as is necessary in GLMs (e.g. via a second-order term). Using simulated data, we demonstrate that our framework can capture specified (non-)linear effects. We consider a setting with only linear effects and a setting including both linear and non-linear effects. For both experiments, we compare the use of a GLM, an XGBoost model and a neural network in the M-step. Additionally, we investigate whether implementing an EM scheme, i.e.~incrementally adapting the model to changing input data, improves the fit for the occurrence and reporting processes compared to using a model with static input data.

In Section~\ref{sec:notation}, we introduce notation as well as the model assumptions. Section~\ref{sec:method} gives a general as well as detailed overview of our proposed expectation-maximisation framework. In Section~\ref{sec:simul}, we simulate event count data and assess how well the specified effects are retrieved when using a GLM, an XGBoost model or a neural network. Section~\ref{sec:covid} illustrates the application of the framework to analyse the reporting and nowcasting of Argentinian Covid-19 cases. Finally, Section~\ref{sec:conclus} concludes the paper.

\section{Notation and setup} \label{sec:notation}
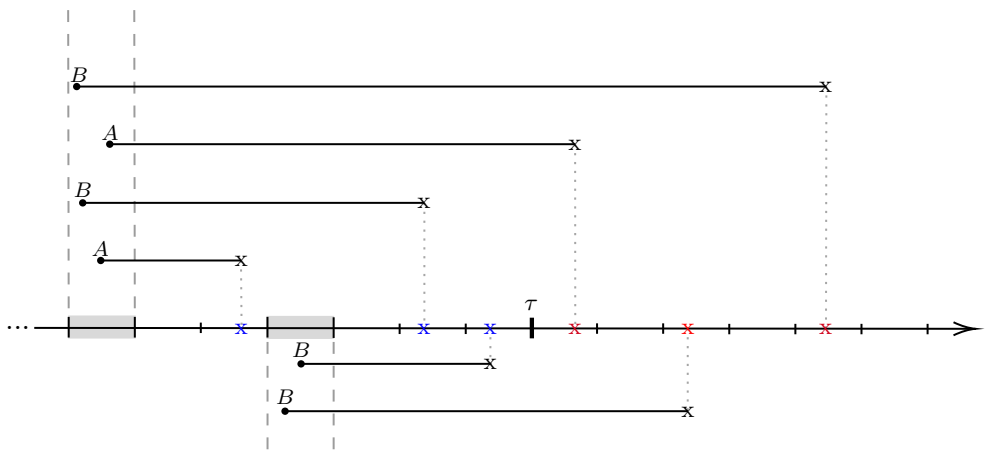
\begin{figure}[H]
    \centering

\tikzset{every picture/.style={line width=0.75pt}} 

\begin{tikzpicture}[x=0.75pt,y=0.75pt,yscale=-1,xscale=1]

\draw  [color={rgb, 255:red, 0; green, 0; blue, 0 }  ,draw opacity=0.15 ][fill={rgb, 255:red, 0; green, 0; blue, 0 }  ,fill opacity=0.15 ] (40.22,160.06) -- (72.97,160.06) -- (72.97,170.66) -- (40.22,170.66) -- cycle ;
\draw    (139.22,160.26) -- (139.22,170.66) ;
\draw    (22.8,165.8) -- (490.6,166.26) ;
\draw [shift={(492.6,166.26)}, rotate = 180.06] [color={rgb, 255:red, 0; green, 0; blue, 0 }  ][line width=0.75]    (10.93,-3.29) .. controls (6.95,-1.4) and (3.31,-0.3) .. (0,0) .. controls (3.31,0.3) and (6.95,1.4) .. (10.93,3.29)   ;
\draw    (73,160.26) -- (73,170.66) ;
\draw[ultra thick] (271.2,160.66) -- (271.2,171.06) ;
\draw    (40,160.26) -- (40,170.66) ;
\draw [color={rgb, 255:red, 0; green, 0; blue, 0 }  ,draw opacity=1 ][fill={rgb, 255:red, 155; green, 155; blue, 155 }  ,fill opacity=1 ]   (105.94,163.14) -- (105.94,168.39) ;
\draw [color={rgb, 255:red, 0; green, 0; blue, 0 }  ,draw opacity=0.39 ] [dash pattern={on 4.5pt off 4.5pt}]  (39.73,5.95) -- (40,160.26) ;
\draw [color={rgb, 255:red, 0; green, 0; blue, 0 }  ,draw opacity=0.39 ] [dash pattern={on 4.5pt off 4.5pt}]  (72.73,5.95) -- (73,160.26) ;
\draw [color={rgb, 255:red, 0; green, 0; blue, 0 }  ,draw opacity=1 ][fill={rgb, 255:red, 155; green, 155; blue, 155 }  ,fill opacity=1 ]   (205.19,163.64) -- (205.19,168.89) ;
\draw [color={rgb, 255:red, 0; green, 0; blue, 0 }  ,draw opacity=1 ][fill={rgb, 255:red, 155; green, 155; blue, 155 }  ,fill opacity=1 ]   (238.19,163.64) -- (238.19,168.89) ;
\draw [color={rgb, 255:red, 0; green, 0; blue, 0 }  ,draw opacity=1 ][fill={rgb, 255:red, 155; green, 155; blue, 155 }  ,fill opacity=1 ]   (303.73,163.51) -- (303.73,168.76) ;
\draw [color={rgb, 255:red, 0; green, 0; blue, 0 }  ,draw opacity=1 ][fill={rgb, 255:red, 155; green, 155; blue, 155 }  ,fill opacity=1 ]   (336.73,163.51) -- (336.73,168.76) ;
\draw [color={rgb, 255:red, 0; green, 0; blue, 0 }  ,draw opacity=1 ][fill={rgb, 255:red, 155; green, 155; blue, 155 }  ,fill opacity=1 ]   (369.73,163.76) -- (369.73,169.01) ;
\draw    (56,131.9) -- (126,131.9) ;
\draw [shift={(56,131.9)}, rotate = 0] [color={rgb, 255:red, 0; green, 0; blue, 0 }  ][fill={rgb, 255:red, 0; green, 0; blue, 0 }  ][line width=0.75]      (0, 0) circle [x radius= 1.34, y radius= 1.34]   ;
\draw    (47,102.91) -- (217.54,102.91) ;
\draw [shift={(47,102.91)}, rotate = 0] [color={rgb, 255:red, 0; green, 0; blue, 0 }  ][fill={rgb, 255:red, 0; green, 0; blue, 0 }  ][line width=0.75]      (0, 0) circle [x radius= 1.34, y radius= 1.34]   ;
\draw    (60.5,73.4) -- (292.79,73.4) ;
\draw [shift={(60.5,73.4)}, rotate = 0] [color={rgb, 255:red, 0; green, 0; blue, 0 }  ][fill={rgb, 255:red, 0; green, 0; blue, 0 }  ][line width=0.75]      (0, 0) circle [x radius= 1.34, y radius= 1.34]   ;
\draw    (44,44.41) -- (418.2,44.39) ;
\draw [shift={(44,44.41)}, rotate = 360] [color={rgb, 255:red, 0; green, 0; blue, 0 }  ][fill={rgb, 255:red, 0; green, 0; blue, 0 }  ][line width=0.75]      (0, 0) circle [x radius= 1.34, y radius= 1.34]   ;
\draw [color={rgb, 255:red, 0; green, 0; blue, 0 }  ,draw opacity=0.35 ] [dash pattern={on 0.84pt off 2.51pt}]  (126,131.9) -- (126.29,165.4) ;
\draw [color={rgb, 255:red, 0; green, 0; blue, 0 }  ,draw opacity=0.35 ] [dash pattern={on 0.84pt off 2.51pt}]  (217.54,102.91) -- (217.54,166.15) ;
\draw [color={rgb, 255:red, 0; green, 0; blue, 0 }  ,draw opacity=0.35 ] [dash pattern={on 0.84pt off 2.51pt}]  (292.79,73.4) -- (292.86,165.65) ;
\draw [color={rgb, 255:red, 0; green, 0; blue, 0 }  ,draw opacity=0.35 ] [dash pattern={on 0.84pt off 2.51pt}]  (418.2,44.39) -- (418.02,168.62) ;
\draw [color={rgb, 255:red, 0; green, 0; blue, 0 }  ,draw opacity=1 ][fill={rgb, 255:red, 155; green, 155; blue, 155 }  ,fill opacity=1 ]   (402.73,163.76) -- (402.73,169.01) ;
\draw [color={rgb, 255:red, 0; green, 0; blue, 0 }  ,draw opacity=1 ][fill={rgb, 255:red, 155; green, 155; blue, 155 }  ,fill opacity=1 ]   (435.73,163.76) -- (435.73,169.01) ;
\draw [color={rgb, 255:red, 0; green, 0; blue, 0 }  ,draw opacity=1 ][fill={rgb, 255:red, 155; green, 155; blue, 155 }  ,fill opacity=1 ]   (468.73,163.76) -- (468.73,169.01) ;
\draw  [color={rgb, 255:red, 0; green, 0; blue, 0 }  ,draw opacity=0.15 ][fill={rgb, 255:red, 0; green, 0; blue, 0 }  ,fill opacity=0.15 ] (139.25,160.26) -- (172,160.26) -- (172,170.86) -- (139.25,170.86) -- cycle ;
\draw    (172.25,160.46) -- (172.25,170.86) ;
\draw [color={rgb, 255:red, 0; green, 0; blue, 0 }  ,draw opacity=0.39 ] [dash pattern={on 4.5pt off 4.5pt}]  (139.25,172.86) -- (139.25,227.51) ;
\draw [color={rgb, 255:red, 0; green, 0; blue, 0 }  ,draw opacity=0.39 ] [dash pattern={on 4.5pt off 4.5pt}]  (172.25,172.86) -- (172.25,227.51) ;
\draw    (156,183.9) -- (251,183.88) ;
\draw [shift={(156,183.9)}, rotate = 359.99] [color={rgb, 255:red, 0; green, 0; blue, 0 }  ][fill={rgb, 255:red, 0; green, 0; blue, 0 }  ][line width=0.75]      (0, 0) circle [x radius= 1.34, y radius= 1.34]   ;
\draw    (148,207.7) -- (349.11,207.7) ;
\draw [shift={(148,207.7)}, rotate = 0] [color={rgb, 255:red, 0; green, 0; blue, 0 }  ][fill={rgb, 255:red, 0; green, 0; blue, 0 }  ][line width=0.75]      (0, 0) circle [x radius= 1.34, y radius= 1.34]   ;
\draw [color={rgb, 255:red, 0; green, 0; blue, 0 }  ,draw opacity=0.35 ] [dash pattern={on 0.84pt off 2.51pt}]  (250.5,166) -- (250.5,182) ;
\draw [color={rgb, 255:red, 0; green, 0; blue, 0 }  ,draw opacity=0.35 ] [dash pattern={on 0.84pt off 2.51pt}]  (349,166) -- (349,210) ;

\draw (7,163.5) node [anchor=north west][inner sep=0.75pt]   [align=left] {...};
\draw (50.1,120.83) node [anchor=north west][inner sep=0.75pt]  [font=\scriptsize] [align=left] {$A$};
\draw (54.8,62.33) node [anchor=north west][inner sep=0.75pt]  [font=\scriptsize] [align=left] {$A$};
\draw (41,91.33) node [anchor=north west][inner sep=0.75pt]  [font=\scriptsize] [align=left] {$B$};
\draw (39,33.08) node [anchor=north west][inner sep=0.75pt]  [font=\scriptsize] [align=left] {$B$};
\draw (150,171.66) node [anchor=north west][inner sep=0.75pt]  [font=\scriptsize] [align=left] {$B$};
\draw (142,195.33) node [anchor=north west][inner sep=0.75pt]  [font=\scriptsize] [align=left] {$B$};
\draw (121.4,162.3) node [anchor=north west][inner sep=0.75pt]   [align=left] {{\footnotesize \textcolor[rgb]{0,0,1}{x}}};
\draw (288,162.3) node [anchor=north west][inner sep=0.75pt]   [align=left] {{\footnotesize \textcolor[rgb]{0.82,0.01,0.11}{x}}};
\draw (413.1,162.3) node [anchor=north west][inner sep=0.75pt]   [align=left] {{\footnotesize \textcolor[rgb]{0.82,0.01,0.11}{x}}};
\draw (212.6,162.3) node [anchor=north west][inner sep=0.75pt]   [align=left] {{\footnotesize \textcolor[rgb]{0,0,1}{x}}};
\draw (121.4,128) node [anchor=north west][inner sep=0.75pt]  [font=\footnotesize] [align=left] {x};
\draw (212.6,99) node [anchor=north west][inner sep=0.75pt]  [font=\footnotesize] [align=left] {x};
\draw (288,69.5) node [anchor=north west][inner sep=0.75pt]  [font=\footnotesize] [align=left] {x};
\draw (413.1,40.5) node [anchor=north west][inner sep=0.75pt]  [font=\footnotesize] [align=left] {x};
\draw (245.7,162.3) node [anchor=north west][inner sep=0.75pt]   [align=left] {{\footnotesize \textcolor[rgb]{0,0,1}{x}}};
\draw (344.4,162.3) node [anchor=north west][inner sep=0.75pt]   [align=left] {{\footnotesize \textcolor[rgb]{1,0,0}{x}}};
\draw (266,150) node [anchor=north west][inner sep=0.75pt]  [font=\footnotesize] [align=left] {{\footnotesize $\tau$}};
\draw (245.7,180) node [anchor=north west][inner sep=0.75pt]  [font=\footnotesize] [align=left] {x};
\draw (344.4,204) node [anchor=north west][inner sep=0.75pt]  [font=\footnotesize] [align=left] {x};

\end{tikzpicture}
    
    \caption{Occurrence and reporting timing of six events that happened in two different occurrence periods for two different entities, denoted $A$ and $B$. The two occurrence periods are highlighted in grey. A blue or red x-mark indicates whether an event is observable or unobservable at present time $\tau$, respectively.}
    \label{fig:timeline2}
\end{figure}

Let $N_i$ denote the number of events that occurred for entity $\text{ent}(i)$ in a discrete time period $\text{occ}(i)$. In epidemiology, for example, this can be the number of infected cases $N_i$ that occurred for a specific region $\text{ent}(i)$ in a specific month $\text{occ}(i)$. In Figure~\ref{fig:timeline2}, we show a toy example for two entities, denoted as $A$ and $B$, with events occurring in two different time periods (highlighted with a grey bar). Events for entity $A$ are recorded only in the first occurrence period, while events for $B$ are recorded in both occurrence periods. In our proposed framework, each combination of an entity and an occurrence period constitutes a separate observation. At present time $\tau$, only partial information is available, since some events are not yet reported. For the example sketched in Figure~\ref{fig:timeline2}, this results in three observations, i.e.~one observation for entity $A$ and two observations for entity $B$, each consisting of one observable event occurrence.

When faced with a reporting delay, we can not directly observe the number of events $N_i$. At present time $\tau$, we can only observe partial information about $N_i$. We use $N_{i,j}$ to denote the number of events for observation $i$ that are reported $j-1$ time periods after the occurrence, i.e.~in period $\text{occ}(i)+j-1$. We use $\boldsymbol{x}_i$ to denote the covariate vector for observation $i$. Let $\lambda(\boldsymbol{x}_i)$ denote the mean number of events that occurs for entity $\text{ent}(i)$ in occurrence period $\text{occ}(i)$, i.e.~the occurrence intensity for observation $i$. The symbol $p_j(\boldsymbol{x}_i)$ denotes the probability that an event occurring for entity $\text{ent}(i)$ in occurrence period $\text{occ}(i)$ is reported $j-1$ periods later, i.e.~the reporting probability for observation $i$ in period $\text{occ}(i)+j-1$. All periods are assumed to be of the same length. We adopt the same assumptions as in \cite{crevecoeur2023bridging}:

\begin{itemize}
    \item[1.] events can be reported at most $d-1$ time units after occurrence with $d$ smaller or equal to the observation window $\tau$, i.e.~$d \le \tau$,
    \item[2.] $N_i \sim \text{Poisson}(\lambda(\boldsymbol{x}_i))$ with $\lambda(\boldsymbol{x}_i) \ge 0$,
    \item[3.] $N_{i,j}|N_i \sim \text{Multinomial}(p_j(\boldsymbol{x}_i))$ with $j=1,\ldots,d$ and $\sum_{j=1}^{d} p_j(\boldsymbol{x}_i)=1$. 
\end{itemize}
Let $\tau_{i}=\min(d,\tau-\text{occ}(i)+1)$ denote the number of observable periods for observation $i$ within the maximal observation delay $d$. We define $\boldsymbol{N}_i=(N_{i,1},\ldots,N_{i,d})$ and $\boldsymbol{N}_i^{r}=(N_{i,1},\ldots,N_{i,\tau_{i}})$ as the complete and the observed event information for observation $i$, respectively. We assume the availability of an observed dataset $D=(\boldsymbol{N}_i^{r},\boldsymbol{x}_i)_{i=1}^{n}$ consisting of $n$ observations, where $\boldsymbol{x}_i=(\boldsymbol{x}_{\text{ent}(i)},\boldsymbol{x}_{\text{per}(i)})$ is the covariate vector that contains the entity-specific information $\boldsymbol{x}_{\text{ent}(i)}$ as well as the period-related information $\boldsymbol{x}_{\text{per}(i)}=(\boldsymbol{x}_{\text{occ}(i)},\ldots,\boldsymbol{x}_{\text{occ}(i)+d-1})$ for the occurrence period $\text{occ}(i)$ and the other possible reporting periods $\text{occ}(i)+1,\ldots,\text{occ}(i)+d-1$. Our objective is to build a framework that can estimate the occurrence intensities $\lambda(\boldsymbol{x}_i)$ as well as the reporting probabilities $p_j(\boldsymbol{x}_i)$ with $j=1,\ldots,d$, without requiring an explicit functional form. By incorporating machine learning models such as XGBoost or neural networks in the maximisation step, we extend the work of \cite{crevecoeur2023bridging} and \cite{verbelen2022modeling}. We can then capture the (non-)linear dynamics within the occurrence as well as the reporting process at a granular level. This will allows us to obtain more accurate nowcasts for the event counts $N_{i,\tau_{i}+1},\ldots,N_{i,d}$.

\section{Expectation-maximisation algorithm} \label{sec:method}
\subsection{Occurrence and reporting model}
Following the assumptions made in Section~\ref{sec:notation} and the Poisson thinning property, we have that $N_i = \sum^{d}_{j=1} N_{i,j}$ and $N_{i,j} \sim \text{Poisson}(\lambda(\boldsymbol{x}_i) \cdot p_j(\boldsymbol{x}_i))$ \citep{verbelen2022modeling}. Consequently, the log-likelihood of the observed dataset $D$ is
\begin{align*}
    LL(\lambda_i, p_{i,j};D)=\sum_{i=1}^n \sum_{j=1}^{\tau_i}\bigg[-\lambda_i \cdot p_{i,j}+N_{i,j} \cdot \log \left(\lambda_i\right)+N_{i,j} \cdot \log \left(p_{i,j}\right)-\log \left(N_{i,j} !\right)\bigg],
\end{align*}
where we use the shorthand notation $\lambda_i \overset{\text{not.}}{=} \lambda\left(\boldsymbol{x}_i\right)$ and $p_{i,j} \overset{\text{not.}}{=} p_j\left(\boldsymbol{x}_i\right)$. The direct maximisation of this log-likelihood is not straightforward due to the presence of the interaction term $-\lambda_i \cdot p_{i,j}$. However, this difficulty can be addressed by viewing the problem as one of incomplete data, which allows the occurrence and the reporting parameters to be decoupled. Let $\mathcal{D}=(\boldsymbol{N}_{i},\boldsymbol{x}_i)_{i=1}^{n}$ denote the dataset containing complete information, which is only partially observed at present time $\tau$. Following \cite{crevecoeur2023bridging} and \cite{verbelen2022modeling}, the (full or complete) log-likelihood then becomes
\begin{align*}
    LL_{c}(\lambda_i, p_{i,j};\mathcal{D})=\sum_{i=1}^n\bigg[-\lambda_i+N_i \cdot \log \left(\lambda_i\right)+\sum_{j=1}^d\left\{N_{i, j} \cdot \log \left(p_{i,j}\right)-\log \left(N_{i,j} !\right)\right\}\bigg].
\end{align*}
The interaction term is no longer present, allowing for the estimation of the occurrence intensities $\lambda\left(\boldsymbol{x}_i\right)$ and the reporting probabilities $p_j\left(\boldsymbol{x}_i\right)$ by separately maximising
\begin{align*}
    LL_{c}^{\text{occ}}(\lambda_i;\mathcal{D}) &= \sum_{i=1}^n\left[ -\lambda_i+N_i \cdot \log \left(\lambda_i\right) \right] \text{, and}\\
    LL_{c}^{\text{rep}}(p_{i,j};\mathcal{D}) &= \sum_{i=1}^n \sum_{j=1}^d N_{i, j} \cdot \log \left(p_{i,j}\right), \:\:\: \text{subject to} \:\:\: \sum_{j=1}^d p_{i,j} = 1 \:\:\: \text{with} \:\:\: i=1,\ldots,n.
\end{align*}
However, since we cannot directly observe the full event information, optimising these log-likelihoods is not feasible. As elaborated in the next section, an expectation-maximisation approach provides a solution by replacing the missing information with its expected value (under the modelling assumptions of Section~\ref{sec:notation}), allowing for the optimisation of the log-likelihoods $LL_{c}^{\text{occ}}(\lambda_i;\mathcal{D})$ and $LL_{c}^{\text{rep}}(p_{i,j};\mathcal{D})$ given those expected values.

\subsection{Expectation-maximisation framework}
\begin{figure}[H] 
\centering

\tikzset{every picture/.style={line width=0.75pt}} 

\begin{adjustbox}{width=\textwidth}

\tikzset{every picture/.style={line width=0.75pt}} 

\begin{tikzpicture}[x=0.75pt,y=0.75pt,yscale=-1,xscale=1]

\draw  [fill={rgb, 255:red, 193; green, 190; blue, 190 }  ,fill opacity=0.43 ] (11.75,122.78) -- (245,122.78) -- (245,162.78) -- (11.75,162.78) -- cycle ;
\draw  [fill={rgb, 255:red, 193; green, 190; blue, 190 }  ,fill opacity=0.43 ] (309,122.78) -- (542.25,122.78) -- (542.25,162.78) -- (309,162.78) -- cycle ;
\draw  [fill={rgb, 255:red, 193; green, 190; blue, 190 }  ,fill opacity=0.43 ] (631.25,122.78) -- (864.5,122.78) -- (864.5,162.78) -- (631.25,162.78) -- cycle ;
\draw    (267,142.81) -- (283,142.81) ;
\draw [shift={(286,142.81)}, rotate = 180] [fill={rgb, 255:red, 0; green, 0; blue, 0 }  ][line width=0.08]  [draw opacity=0] (8.93,-4.29) -- (0,0) -- (8.93,4.29) -- cycle    ;
\draw    (553,142.81) -- (569,142.81) ;
\draw [shift={(572,142.81)}, rotate = 180] [fill={rgb, 255:red, 0; green, 0; blue, 0 }  ][line width=0.08]  [draw opacity=0] (8.93,-4.29) -- (0,0) -- (8.93,4.29) -- cycle    ;
\draw    (603,142.81) -- (619,142.81) ;
\draw [shift={(622,142.81)}, rotate = 180] [fill={rgb, 255:red, 0; green, 0; blue, 0 }  ][line width=0.08]  [draw opacity=0] (8.93,-4.29) -- (0,0) -- (8.93,4.29) -- cycle    ;
\draw    (127.75,177.28) -- (127.75,210.28) ;
\draw [shift={(127.75,213.28)}, rotate = 270] [fill={rgb, 255:red, 0; green, 0; blue, 0 }  ][line width=0.08]  [draw opacity=0] (8.93,-4.29) -- (0,0) -- (8.93,4.29) -- cycle    ;
\draw    (425.75,177.28) -- (425.75,184.49) -- (425.75,210.28) ;
\draw [shift={(425.75,213.28)}, rotate = 270] [fill={rgb, 255:red, 0; green, 0; blue, 0 }  ][line width=0.08]  [draw opacity=0] (8.93,-4.29) -- (0,0) -- (8.93,4.29) -- cycle    ;
\draw    (746.75,177.28) -- (746.75,210.28) ;
\draw [shift={(746.75,213.28)}, rotate = 270] [fill={rgb, 255:red, 0; green, 0; blue, 0 }  ][line width=0.08]  [draw opacity=0] (8.93,-4.29) -- (0,0) -- (8.93,4.29) -- cycle    ;
\draw  [dash pattern={on 4.5pt off 4.5pt}]  (256.75,213.28) -- (295.01,179.27) ;
\draw [shift={(297.25,177.28)}, rotate = 138.37] [fill={rgb, 255:red, 0; green, 0; blue, 0 }  ][line width=0.08]  [draw opacity=0] (8.93,-4.29) -- (0,0) -- (8.93,4.29) -- cycle    ;
\draw  [dash pattern={on 4.5pt off 4.5pt}]  (551.75,211.28) -- (574.02,179.73) ;
\draw [shift={(575.75,177.28)}, rotate = 125.22] [fill={rgb, 255:red, 0; green, 0; blue, 0 }  ][line width=0.08]  [draw opacity=0] (8.93,-4.29) -- (0,0) -- (8.93,4.29) -- cycle    ;
\draw  [dash pattern={on 4.5pt off 4.5pt}]  (599.75,211.28) -- (622.02,179.73) ;
\draw [shift={(623.75,177.28)}, rotate = 125.22] [fill={rgb, 255:red, 0; green, 0; blue, 0 }  ][line width=0.08]  [draw opacity=0] (8.93,-4.29) -- (0,0) -- (8.93,4.29) -- cycle    ;
\draw  [fill={rgb, 255:red, 255; green, 255; blue, 255 }  ,fill opacity=1 ] (11.75,227.78) -- (245,227.78) -- (245,267.78) -- (11.75,267.78) -- cycle ;
\draw  [fill={rgb, 255:red, 255; green, 255; blue, 255 }  ,fill opacity=1 ] (11.75,279.78) -- (245,279.78) -- (245,319.78) -- (11.75,319.78) -- cycle ;
\draw  [fill={rgb, 255:red, 255; green, 255; blue, 255 }  ,fill opacity=1 ] (308.75,227.78) -- (542,227.78) -- (542,267.78) -- (308.75,267.78) -- cycle ;
\draw  [fill={rgb, 255:red, 255; green, 255; blue, 255 }  ,fill opacity=1 ] (308.75,279.78) -- (542,279.78) -- (542,319.78) -- (308.75,319.78) -- cycle ;
\draw  [fill={rgb, 255:red, 255; green, 255; blue, 255 }  ,fill opacity=1 ] (631,227.78) -- (864.25,227.78) -- (864.25,267.78) -- (631,267.78) -- cycle ;
\draw  [fill={rgb, 255:red, 255; green, 255; blue, 255 }  ,fill opacity=1 ] (631,279.78) -- (864.25,279.78) -- (864.25,319.78) -- (631,319.78) -- cycle ;
\draw (193.35,63.4) -- (193.35,115.77) ;
\draw [shift={(193.35,118.77)}, rotate = 270] [fill={rgb, 255:red, 0; green, 0; blue, 0 }  ,fill opacity=1 ][line width=0.08]  [draw opacity=0] (8.93,-4.29) -- (0,0) -- (8.93,4.29) -- cycle    ;
\draw  [fill={rgb, 255:red, 193; green, 190; blue, 190 }  ,fill opacity=0.43 ] (11.75,18.78) -- (245,18.78) -- (245,58.78) -- (11.75,58.78) -- cycle ;
\draw    (886.29,247.88) -- (902.29,247.88) ;
\draw [shift={(905.29,247.88)}, rotate = 180] [fill={rgb, 255:red, 0; green, 0; blue, 0 }  ][line width=0.08]  [draw opacity=0] (8.93,-4.29) -- (0,0) -- (8.93,4.29) -- cycle    ;
\draw    (886.29,299.88) -- (902.29,299.88) ;
\draw [shift={(905.29,299.88)}, rotate = 180] [fill={rgb, 255:red, 0; green, 0; blue, 0 }  ][line width=0.08]  [draw opacity=0] (8.93,-4.29) -- (0,0) -- (8.93,4.29) -- cycle    ;
\draw (63.4,103.59) -- (63.4,115.77) ;
\draw [shift={(63.4,118.77)}, rotate = 270] [fill={rgb, 255:red, 0; green, 0; blue, 0 }  ,fill opacity=1 ][line width=0.08]  [draw opacity=0] (8.93,-4.29) -- (0,0) -- (8.93,4.29) -- cycle    ;
\draw (63.4,63.4) -- (63.4,75.59) ;
\draw [shift={(63.4,78.59)}, rotate = 270] [fill={rgb, 255:red, 0; green, 0; blue, 0 }  ,fill opacity=1 ][line width=0.08]  [draw opacity=0] (8.93,-4.29) -- (0,0) -- (8.93,4.29) -- cycle    ;

\draw (111.75,132.4) node [anchor=north west][inner sep=0.75pt] [font=\Large]  [align=left] {$ \mathcal{D}^{( 1)}$};
\draw (411.08,132.4) node [anchor=north west][inner sep=0.75pt] [font=\Large]  [align=left] {$\mathcal{D}^{( 2)}$};
\draw (732.42,132.4) node [anchor=north west][inner sep=0.75pt]  [font=\Large] [align=left] {$\mathcal{D}^{( K)}$};
\draw (582.89,140) node [anchor=north west][inner sep=0.75pt]  [font=\small] [align=left] {...};
\draw (99,234) node [anchor=north west][inner sep=0.75pt]  [font=\Large] [align=left] {$f_{\text{occ}}^{( 1)}(\boldsymbol{x}_{i})$};
\draw (99,286) node [anchor=north west][inner sep=0.75pt]  [font=\Large] [align=left] {$f_{\text{rep}}^{( 1)}(\boldsymbol{x}_{i})$};
\draw (398,234) node [anchor=north west][inner sep=0.75pt] [font=\Large]  [align=left] {$f_{\text{occ}}^{( 2)}(\boldsymbol{x}_{i})$};
\draw (398,286) node [anchor=north west][inner sep=0.75pt]  [font=\Large] [align=left] {$ f_{\text{rep}}^{( 2)}(\boldsymbol{x}_{i})$};
\draw (718,234) node [anchor=north west][inner sep=0.75pt] [font=\Large]  [align=left] {$ f_{\text{occ}}^{( K)}(\boldsymbol{x}_{i})$};
\draw (718,286) node [anchor=north west][inner sep=0.75pt] [font=\Large]  [align=left] {$f_{\text{rep}}^{( K)}(\boldsymbol{x}_{i})$};
\draw (582.89,271) node [anchor=north west][inner sep=0.75pt]  [font=\small] [align=left] {...};
\draw (65,28) node [anchor=north west][inner sep=0.75pt]  [font=\Large] [align=left] {$D=(\boldsymbol{N}_i^{r},\boldsymbol{x}_i)_{i=1}^{n}$};
\draw (928.75,235) node [anchor=north west][inner sep=0.75pt] [font=\Large]   [align=left] {$ \widehat{\lambda}^{(K)}(\boldsymbol{x}_{i})$};
\draw (928.75,284) node [anchor=north west][inner sep=0.75pt]  [font=\Large]  [align=left] {$ \widehat{p}^{(K)}_{j}(\boldsymbol{x}_{i})$};
\draw (27,78.74) node [anchor=north west][inner sep=0.75pt] [align=left] {$\left( \widehat{\lambda}^{(0)}_{i} ,\widehat{p}_{i,j}^{(0)}\right)$};
\draw (280,190) node [anchor=north west][inner sep=0.75pt] [align=left] {$\left( \widehat{\lambda}^{(1)}_{i} ,\widehat{p}_{i,j}^{(1)}\right)$};

\end{tikzpicture}

\end{adjustbox}

\caption{Visualisation of the consecutive steps for the model-agnostic expectation-maximisation framework. No information is passed between EM iterations except for the estimates for the occurrence intensities and the reporting probabilities.} 
\label{fig:summary}
\end{figure}
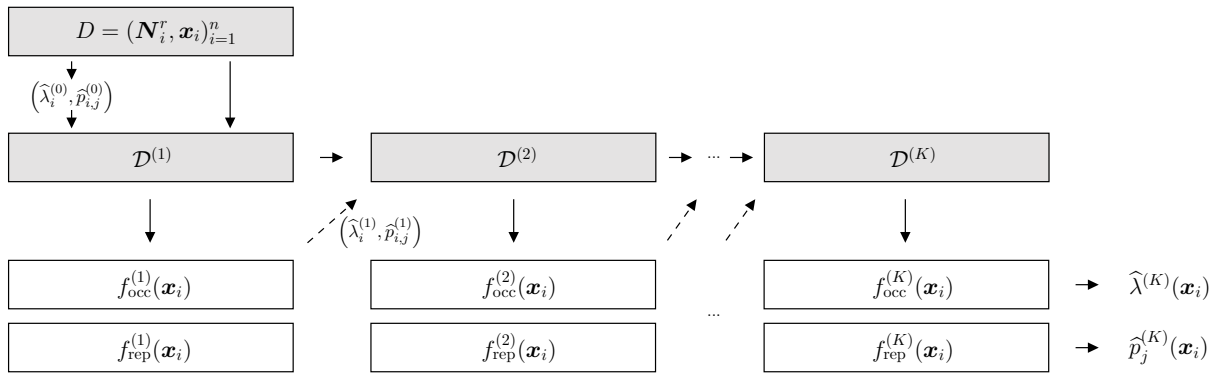

To mitigate the impact of the missing data, we use an EM algorithm to obtain estimates for the occurrence intensities and the reporting probabilities. The EM algorithm alternates between an expectation- and maximisation step. The algorithm steps are similar to \cite{verbelen2022modeling} and \cite{crevecoeur2023bridging}. In the $k$th expectation step, we compute an estimate for the partially observable dataset $\mathcal{D}$ under the model assumptions specified in Section~\ref{sec:notation}. We denote this estimate as $\mathcal{D}^{(k)}=(\boldsymbol{N}_{i}^{(k)},\boldsymbol{x}_i)_{i=1}^{n}$ with $\boldsymbol{N}^{(k)}_i=(N^{(k)}_{i,1},\ldots,N^{(k)}_{i,d})$. Specifically, we replace the unobserved event counts $N_{i,j}$ with estimates based on the current values for the occurrence intensities $\widehat{\lambda}^{(k-1)}(\boldsymbol{x}_i)$ and the reporting probabilities $\widehat{p}_j^{(k-1)}(\boldsymbol{x}_i)$. We define
\begin{align*}
    N_{i,j}^{(k)}= \begin{cases}N_{i,j} & j \leq \tau_i \\ \widehat{\lambda}^{(k-1)}(\boldsymbol{x}_i) \cdot \widehat{p}_{j}^{(k-1)}(\boldsymbol{x_i}) & \tau_i<j \leq d\end{cases},
\end{align*}
with $\tau_i$ denoting the observable time window for observation $i$. In the $k$th maximisation step, we obtain updated parameter estimates for the occurrence intensities and the reporting probabilities, denoted as $\widehat{\lambda}^{(k)}(\boldsymbol{x}_i)$ and $\widehat{p}_j^{(k)}(\boldsymbol{x}_i)$ for $i=1,\ldots,n$ and $j=1,\ldots,d$. We define 
\begin{align*}
    \widehat{\lambda}^{(k)}(\boldsymbol{x}_i) = \exp{(\widehat{f}^{(k)}_{\text{occ}}(\boldsymbol{x}_i))} \:\: \text{and} \:\: \widehat{p}_{j}^{(k)}(\boldsymbol{x}_i)= \frac{\exp(\widehat{f}_{\text{rep}}^{(k)[j]}(\boldsymbol{x}_{i}))}{\sum_{z=1}^{d} \exp(\widehat{f}_{\text{rep}}^{(k)[z]}(\boldsymbol{x}_{i}))},
\end{align*}
where $f^{(k)}_{\text{occ}}(.)$ and $f^{(k)}_{\text{rep}}(.)=(f^{(k)[1]}_{\text{rep}}(.),\ldots,f^{(k)[d]}_{\text{rep}}(.))$ are predictive functions learned in the $k$th EM iteration for the occurrence and reporting model, respectively. Following our assumptions in Section~\ref{sec:notation}, we apply an exponential link function for the occurrence intensities and a softmax transformation for the reporting probabilities. To learn these predictive functions, we maximise the expectation of the (complete) log-likelihoods $LL_{c}^{\text{occ}}(\lambda_i;\mathcal{D})$ and $ LL_{c}^{\text{rep}}(p_{i,j};\mathcal{D})$ given the current estimate of the complete data $\mathcal{D}^{(k)}$:
\begin{align*}
    Q^{\text{occ}}(\lambda_{i}^{(k)};\mathcal{D}^{(k)}) &= \mathbb{E}\left(LL_{c}^{\text{occ}}(\lambda^{(k)}_i;\mathcal{D}) | \mathcal{D}^{(k)}\right)\\
    &= \sum_{i=1}^n\left[ -\lambda_{i}^{(k)}+N^{(k)}_i \cdot \log \left(\lambda_{i}^{(k)}\right) \right], \:\:\: \text{and}\\
    Q^{\text{rep}}(p_{i,j}^{(k)};\mathcal{D}^{(k)})&= \mathbb{E}\left( LL_{c}^{\text{rep}}(p_{i,j}^{(k)};\mathcal{D}) | \mathcal{D}^{(k)}\right)\\ &= \sum_{i=1}^n \sum_{j=1}^d N^{(k)}_{i, j} \cdot \log \left(p_{i,j}^{(k)}\right), \:\:\: \text{subject to} \:\:\: \sum_{j=1}^d p_{i,j}^{(k)} = 1 \:\:\: \text{for} \:\:\: i=1,\ldots,n,
\end{align*}
respectively. By alternating between the expectation and the maximisation step, the parameters are updated in a stepwise manner. Figure~\ref{fig:summary} visualises the consecutive steps. The predictive functions $f^{(k)}_{\text{occ}}(.)$ and $f^{(k)}_{\text{rep}}(.)$ can represent any type of predictive model, as long as the predictions adhere to the assumptions made in Section~\ref{sec:notation}.

After each EM iteration, we calculate the observed log-likelihood $LL(\widehat{\lambda}^{(k)}_i, \widehat{p}^{(k)}_{i,j};D)$ using the most recent estimates for the occurrence intensities and reporting probabilities. When generalised linear models are utilised in the maximisation step, the observed log-likelihood $LL(\widehat{\lambda}^{(k)}_i, \widehat{p}^{(k)}_{i,j};D)$ increases over the iterations of the EM algorithm \citep{dempster1977maximum}, which is not the case for machine learning models. For GLMs, optimisation is exact as in the Gaussian case or there are mathematical results that guarantee convergence to the global optimum for numerical optimisation. Conversely, machine learning models are optimised using gradient descent-based methods, which do not have the same mathematical convergence guarantees and therefore can get stuck in local optima. Consequently, as discussed in \cite{delong2021gamma}, we need to ensure that information properly passes from one EM iteration to the next when applying machine learning models in an EM algorithm. For different (machine learning) models, this issue needs to be addressed carefully in different ways when implementing the model in the maximisation step. In the remainder of this section, we sketch approaches for neural networks and XGBoost models.

\begin{figure}[H]
    \centering

    \scalebox{0.7}{

\tikzset{every picture/.style={line width=0.75pt}} 

\begin{tikzpicture}[x=0.75pt,y=0.75pt,yscale=-1,xscale=1]

\draw   (11.75,113.78) -- (245,113.78) -- (245,220.33) -- (11.75,220.33) -- cycle ;
\draw   (288.75,113.78) -- (522,113.78) -- (522,220.33) -- (288.75,220.33) -- cycle ;
\draw   (611,114.03) -- (844.25,114.03) -- (844.25,220.58) -- (611,220.58) -- cycle ;
\draw  [fill={rgb, 255:red, 193; green, 190; blue, 190 }  ,fill opacity=0.43 ] (11.75,12.78) -- (245,12.78) -- (245,52.78) -- (11.75,52.78) -- cycle ;
\draw  [fill={rgb, 255:red, 193; green, 190; blue, 190 }  ,fill opacity=0.43 ] (289,12.78) -- (522.25,12.78) -- (522.25,52.78) -- (289,52.78) -- cycle ;
\draw  [fill={rgb, 255:red, 193; green, 190; blue, 190 }  ,fill opacity=0.43 ] (611.25,12.78) -- (844.5,12.78) -- (844.5,52.78) -- (611.25,52.78) -- cycle ;
\draw    (257,32.81) -- (273,32.81) ;
\draw [shift={(276,32.81)}, rotate = 180] [fill={rgb, 255:red, 0; green, 0; blue, 0 }  ][line width=0.08]  [draw opacity=0] (8.93,-4.29) -- (0,0) -- (8.93,4.29) -- cycle    ;
\draw    (533,32.81) -- (549,32.81) ;
\draw [shift={(552,32.81)}, rotate = 180] [fill={rgb, 255:red, 0; green, 0; blue, 0 }  ][line width=0.08]  [draw opacity=0] (8.93,-4.29) -- (0,0) -- (8.93,4.29) -- cycle    ;
\draw    (583,32.81) -- (599,32.81) ;
\draw [shift={(602,32.81)}, rotate = 180] [fill={rgb, 255:red, 0; green, 0; blue, 0 }  ][line width=0.08]  [draw opacity=0] (8.93,-4.29) -- (0,0) -- (8.93,4.29) -- cycle    ;
\draw    (127.75,67.28) -- (127.75,100.28) ;
\draw [shift={(127.75,103.28)}, rotate = 270] [fill={rgb, 255:red, 0; green, 0; blue, 0 }  ][line width=0.08]  [draw opacity=0] (8.93,-4.29) -- (0,0) -- (8.93,4.29) -- cycle    ;
\draw    (404.75,67.28) -- (404.75,100.28) ;
\draw [shift={(404.75,103.28)}, rotate = 270] [fill={rgb, 255:red, 0; green, 0; blue, 0 }  ][line width=0.08]  [draw opacity=0] (8.93,-4.29) -- (0,0) -- (8.93,4.29) -- cycle    ;
\draw    (726.75,67.28) -- (726.75,100.28) ;
\draw [shift={(726.75,103.28)}, rotate = 270] [fill={rgb, 255:red, 0; green, 0; blue, 0 }  ][line width=0.08]  [draw opacity=0] (8.93,-4.29) -- (0,0) -- (8.93,4.29) -- cycle    ;
\draw  [dash pattern={on 4.5pt off 4.5pt}]  (254.75,101.28) -- (277.02,69.73) ;
\draw [shift={(278.75,67.28)}, rotate = 125.22] [fill={rgb, 255:red, 0; green, 0; blue, 0 }  ][line width=0.08]  [draw opacity=0] (8.93,-4.29) -- (0,0) -- (8.93,4.29) -- cycle    ;
\draw  [dash pattern={on 4.5pt off 4.5pt}]  (531.75,101.28) -- (554.02,69.73) ;
\draw [shift={(555.75,67.28)}, rotate = 125.22] [fill={rgb, 255:red, 0; green, 0; blue, 0 }  ][line width=0.08]  [draw opacity=0] (8.93,-4.29) -- (0,0) -- (8.93,4.29) -- cycle    ;
\draw  [dash pattern={on 4.5pt off 4.5pt}]  (579.75,101.28) -- (602.02,69.73) ;
\draw [shift={(603.75,67.28)}, rotate = 125.22] [fill={rgb, 255:red, 0; green, 0; blue, 0 }  ][line width=0.08]  [draw opacity=0] (8.93,-4.29) -- (0,0) -- (8.93,4.29) -- cycle    ;
\draw    (11.75,284.33) -- (245,284.33) ;
\draw    (11.75,277.33) -- (11.75,284.33) ;
\draw    (245,277.33) -- (245,284.33) ;
\draw  [fill={rgb, 255:red, 155; green, 155; blue, 155 }  ,fill opacity=0.56 ] (51.72,127.16) .. controls (51.72,123.35) and (55.57,120.26) .. (60.31,120.26) .. controls (65.06,120.26) and (68.9,123.35) .. (68.9,127.16) .. controls (68.9,130.97) and (65.06,134.06) .. (60.31,134.06) .. controls (55.57,134.06) and (51.72,130.97) .. (51.72,127.16) -- cycle ;
\draw  [fill={rgb, 255:red, 155; green, 155; blue, 155 }  ,fill opacity=0.56 ] (51.72,154.48) .. controls (51.72,150.67) and (55.57,147.58) .. (60.31,147.58) .. controls (65.06,147.58) and (68.9,150.67) .. (68.9,154.48) .. controls (68.9,158.29) and (65.06,161.38) .. (60.31,161.38) .. controls (55.57,161.38) and (51.72,158.29) .. (51.72,154.48) -- cycle ;
\draw [line width=1.5]  [dash pattern={on 1.69pt off 2.76pt}]  (60.31,137.37) -- (60.32,145.27) ;
\draw  [fill={rgb, 255:red, 155; green, 155; blue, 155 }  ,fill opacity=0.56 ] (51.72,179.32) .. controls (51.72,175.51) and (55.57,172.42) .. (60.31,172.42) .. controls (65.06,172.42) and (68.9,175.51) .. (68.9,179.32) .. controls (68.9,183.13) and (65.06,186.21) .. (60.31,186.21) .. controls (55.57,186.21) and (51.72,183.13) .. (51.72,179.32) -- cycle ;
\draw  [fill={rgb, 255:red, 155; green, 155; blue, 155 }  ,fill opacity=0.56 ] (51.72,206.64) .. controls (51.72,202.83) and (55.57,199.74) .. (60.31,199.74) .. controls (65.06,199.74) and (68.9,202.83) .. (68.9,206.64) .. controls (68.9,210.45) and (65.06,213.53) .. (60.31,213.53) .. controls (55.57,213.53) and (51.72,210.45) .. (51.72,206.64) -- cycle ;
\draw [line width=1.5]  [dash pattern={on 1.69pt off 2.76pt}]  (60.31,189.53) -- (60.32,197.43) ;
\draw  [fill={rgb, 255:red, 248; green, 231; blue, 28 }  ,fill opacity=0.39 ] (92.62,141.23) .. controls (92.62,137.42) and (96.47,134.33) .. (101.21,134.33) .. controls (105.96,134.33) and (109.8,137.42) .. (109.8,141.23) .. controls (109.8,145.04) and (105.96,148.13) .. (101.21,148.13) .. controls (96.47,148.13) and (92.62,145.04) .. (92.62,141.23) -- cycle ;
\draw  [fill={rgb, 255:red, 248; green, 231; blue, 28 }  ,fill opacity=0.39 ] (92.62,193.67) .. controls (92.62,189.86) and (96.47,186.77) .. (101.21,186.77) .. controls (105.96,186.77) and (109.8,189.86) .. (109.8,193.67) .. controls (109.8,197.48) and (105.96,200.56) .. (101.21,200.56) .. controls (96.47,200.56) and (92.62,197.48) .. (92.62,193.67) -- cycle ;
\draw    (72.82,127.93) -- (89.79,138.32) ;
\draw    (90.07,145.16) -- (72.87,155.32) ;
\draw    (72.82,179.31) -- (89.79,189.7) ;
\draw    (90.07,196.55) -- (72.87,206.7) ;
\draw [line width=1.5]  [dash pattern={on 1.69pt off 2.76pt}]  (101.34,164.08) -- (101.35,171.98) ;
\draw    (72.74,174.05) -- (90.07,149.55) ;
\draw    (89.93,183.88) -- (72.88,160.03) ;
\draw    (72.82,132.75) -- (89.79,176.7) ;
\draw    (72.82,200.09) -- (90.07,156.57) ;
\draw  [fill={rgb, 255:red, 248; green, 231; blue, 28 }  ,fill opacity=0.39 ] (152.42,141.16) .. controls (152.42,137.35) and (156.27,134.26) .. (161.02,134.26) .. controls (165.76,134.26) and (169.61,137.35) .. (169.61,141.16) .. controls (169.61,144.97) and (165.76,148.06) .. (161.02,148.06) .. controls (156.27,148.06) and (152.42,144.97) .. (152.42,141.16) -- cycle ;
\draw  [fill={rgb, 255:red, 248; green, 231; blue, 28 }  ,fill opacity=0.39 ] (152.42,193.59) .. controls (152.42,189.78) and (156.27,186.69) .. (161.02,186.69) .. controls (165.76,186.69) and (169.61,189.78) .. (169.61,193.59) .. controls (169.61,197.4) and (165.76,200.49) .. (161.02,200.49) .. controls (156.27,200.49) and (152.42,197.4) .. (152.42,193.59) -- cycle ;
\draw [line width=1.5]  [dash pattern={on 1.69pt off 2.76pt}]  (161.49,164.08) -- (161.5,171.98) ;
\draw    (114.07,141.24) -- (124.16,141.25) ;
\draw    (114.07,193.67) -- (124.16,193.69) ;
\draw    (114.07,146.64) -- (124.44,184.97) ;
\draw    (124.16,146.59) -- (113.79,184.91) ;
\draw  [dash pattern={on 4.5pt off 4.5pt}] (50,119.39) -- (70.54,119.39) -- (70.54,162.42) -- (50,162.42) -- cycle ;
\draw  [dash pattern={on 4.5pt off 4.5pt}] (50,171.53) -- (70.54,171.53) -- (70.54,214.56) -- (50,214.56) -- cycle ;
\draw    (137.44,141.24) -- (147.53,141.25) ;
\draw    (137.44,193.67) -- (147.53,193.69) ;
\draw    (137.44,146.64) -- (147.81,184.97) ;
\draw    (147.53,146.59) -- (137.16,184.91) ;
\draw [line width=1.5]  [dash pattern={on 1.69pt off 2.76pt}]  (133.24,141.28) -- (127.88,141.34) ;
\draw [line width=1.5]  [dash pattern={on 1.69pt off 2.76pt}]  (133.24,193.72) -- (127.88,193.78) ;
\draw  [fill={rgb, 255:red, 184; green, 233; blue, 134 }  ,fill opacity=0.55 ] (187.48,167.65) .. controls (187.48,163.84) and (191.33,160.75) .. (196.07,160.75) .. controls (200.82,160.75) and (204.67,163.84) .. (204.67,167.65) .. controls (204.67,171.46) and (200.82,174.55) .. (196.07,174.55) .. controls (191.33,174.55) and (187.48,171.46) .. (187.48,167.65) -- cycle ;
\draw    (172.15,142.11) -- (185.85,160.51) ;
\draw    (172.15,193.25) -- (185.85,174.86) ;
\draw  [fill={rgb, 255:red, 155; green, 155; blue, 155 }  ,fill opacity=0.56 ] (328.8,126.91) .. controls (328.8,123.1) and (332.65,120.01) .. (337.39,120.01) .. controls (342.14,120.01) and (345.99,123.1) .. (345.99,126.91) .. controls (345.99,130.72) and (342.14,133.81) .. (337.39,133.81) .. controls (332.65,133.81) and (328.8,130.72) .. (328.8,126.91) -- cycle ;
\draw  [fill={rgb, 255:red, 155; green, 155; blue, 155 }  ,fill opacity=0.56 ] (328.8,154.23) .. controls (328.8,150.42) and (332.65,147.33) .. (337.39,147.33) .. controls (342.14,147.33) and (345.99,150.42) .. (345.99,154.23) .. controls (345.99,158.04) and (342.14,161.13) .. (337.39,161.13) .. controls (332.65,161.13) and (328.8,158.04) .. (328.8,154.23) -- cycle ;
\draw [line width=1.5]  [dash pattern={on 1.69pt off 2.76pt}]  (337.39,137.12) -- (337.41,145.02) ;
\draw  [fill={rgb, 255:red, 155; green, 155; blue, 155 }  ,fill opacity=0.56 ] (328.8,179.07) .. controls (328.8,175.26) and (332.65,172.17) .. (337.39,172.17) .. controls (342.14,172.17) and (345.99,175.26) .. (345.99,179.07) .. controls (345.99,182.88) and (342.14,185.96) .. (337.39,185.96) .. controls (332.65,185.96) and (328.8,182.88) .. (328.8,179.07) -- cycle ;
\draw  [fill={rgb, 255:red, 155; green, 155; blue, 155 }  ,fill opacity=0.56 ] (328.8,206.39) .. controls (328.8,202.58) and (332.65,199.49) .. (337.39,199.49) .. controls (342.14,199.49) and (345.99,202.58) .. (345.99,206.39) .. controls (345.99,210.2) and (342.14,213.28) .. (337.39,213.28) .. controls (332.65,213.28) and (328.8,210.2) .. (328.8,206.39) -- cycle ;
\draw [line width=1.5]  [dash pattern={on 1.69pt off 2.76pt}]  (337.39,189.28) -- (337.41,197.18) ;
\draw  [fill={rgb, 255:red, 248; green, 231; blue, 28 }  ,fill opacity=0.39 ] (369.7,140.98) .. controls (369.7,137.17) and (373.55,134.08) .. (378.3,134.08) .. controls (383.04,134.08) and (386.89,137.17) .. (386.89,140.98) .. controls (386.89,144.79) and (383.04,147.88) .. (378.3,147.88) .. controls (373.55,147.88) and (369.7,144.79) .. (369.7,140.98) -- cycle ;
\draw  [fill={rgb, 255:red, 248; green, 231; blue, 28 }  ,fill opacity=0.39 ] (369.7,193.42) .. controls (369.7,189.61) and (373.55,186.52) .. (378.3,186.52) .. controls (383.04,186.52) and (386.89,189.61) .. (386.89,193.42) .. controls (386.89,197.23) and (383.04,200.31) .. (378.3,200.31) .. controls (373.55,200.31) and (369.7,197.23) .. (369.7,193.42) -- cycle ;
\draw    (349.91,127.68) -- (366.87,138.07) ;
\draw    (367.15,144.91) -- (349.95,155.07) ;
\draw    (349.91,179.06) -- (366.87,189.45) ;
\draw    (367.15,196.3) -- (349.95,206.45) ;
\draw [line width=1.5]  [dash pattern={on 1.69pt off 2.76pt}]  (378.42,163.83) -- (378.43,171.73) ;
\draw    (349.83,173.8) -- (367.15,149.3) ;
\draw    (367.01,183.63) -- (349.96,159.78) ;
\draw    (349.91,132.5) -- (366.87,176.45) ;
\draw    (349.91,199.84) -- (367.15,156.32) ;
\draw  [fill={rgb, 255:red, 248; green, 231; blue, 28 }  ,fill opacity=0.39 ] (429.51,140.91) .. controls (429.51,137.1) and (433.35,134.01) .. (438.1,134.01) .. controls (442.85,134.01) and (446.69,137.1) .. (446.69,140.91) .. controls (446.69,144.72) and (442.85,147.81) .. (438.1,147.81) .. controls (433.35,147.81) and (429.51,144.72) .. (429.51,140.91) -- cycle ;
\draw  [fill={rgb, 255:red, 248; green, 231; blue, 28 }  ,fill opacity=0.39 ] (429.51,193.34) .. controls (429.51,189.53) and (433.35,186.44) .. (438.1,186.44) .. controls (442.85,186.44) and (446.69,189.53) .. (446.69,193.34) .. controls (446.69,197.15) and (442.85,200.24) .. (438.1,200.24) .. controls (433.35,200.24) and (429.51,197.15) .. (429.51,193.34) -- cycle ;
\draw [line width=1.5]  [dash pattern={on 1.69pt off 2.76pt}]  (438.57,163.83) -- (438.58,171.73) ;
\draw    (391.15,140.99) -- (401.24,141) ;
\draw    (391.15,193.42) -- (401.24,193.44) ;
\draw    (391.15,146.39) -- (401.52,184.72) ;
\draw    (401.24,146.34) -- (390.87,184.66) ;
\draw  [dash pattern={on 4.5pt off 4.5pt}] (327.08,119.14) -- (347.63,119.14) -- (347.63,162.17) -- (327.08,162.17) -- cycle ;
\draw  [dash pattern={on 4.5pt off 4.5pt}] (327.08,171.28) -- (347.63,171.28) -- (347.63,214.31) -- (327.08,214.31) -- cycle ;
\draw    (414.52,140.99) -- (424.62,141) ;
\draw    (414.52,193.42) -- (424.62,193.44) ;
\draw    (414.52,146.39) -- (424.89,184.72) ;
\draw    (424.62,146.34) -- (414.25,184.66) ;
\draw [line width=1.5]  [dash pattern={on 1.69pt off 2.76pt}]  (410.32,141.03) -- (404.96,141.09) ;
\draw [line width=1.5]  [dash pattern={on 1.69pt off 2.76pt}]  (410.32,193.47) -- (404.96,193.53) ;
\draw  [fill={rgb, 255:red, 184; green, 233; blue, 134 }  ,fill opacity=0.55 ] (464.56,167.4) .. controls (464.56,163.59) and (468.41,160.5) .. (473.16,160.5) .. controls (477.9,160.5) and (481.75,163.59) .. (481.75,167.4) .. controls (481.75,171.21) and (477.9,174.3) .. (473.16,174.3) .. controls (468.41,174.3) and (464.56,171.21) .. (464.56,167.4) -- cycle ;
\draw    (449.23,141.86) -- (462.94,160.26) ;
\draw    (449.23,193) -- (462.94,174.61) ;
\draw  [fill={rgb, 255:red, 155; green, 155; blue, 155 }  ,fill opacity=0.56 ] (651.27,127.16) .. controls (651.27,123.35) and (655.12,120.26) .. (659.87,120.26) .. controls (664.61,120.26) and (668.46,123.35) .. (668.46,127.16) .. controls (668.46,130.97) and (664.61,134.06) .. (659.87,134.06) .. controls (655.12,134.06) and (651.27,130.97) .. (651.27,127.16) -- cycle ;
\draw  [fill={rgb, 255:red, 155; green, 155; blue, 155 }  ,fill opacity=0.56 ] (651.27,154.48) .. controls (651.27,150.67) and (655.12,147.58) .. (659.87,147.58) .. controls (664.61,147.58) and (668.46,150.67) .. (668.46,154.48) .. controls (668.46,158.29) and (664.61,161.38) .. (659.87,161.38) .. controls (655.12,161.38) and (651.27,158.29) .. (651.27,154.48) -- cycle ;
\draw  [fill={rgb, 255:red, 155; green, 155; blue, 155 }  ,fill opacity=0.56 ] (651.27,179.32) .. controls (651.27,175.51) and (655.12,172.42) .. (659.87,172.42) .. controls (664.61,172.42) and (668.46,175.51) .. (668.46,179.32) .. controls (668.46,183.13) and (664.61,186.21) .. (659.87,186.21) .. controls (655.12,186.21) and (651.27,183.13) .. (651.27,179.32) -- cycle ;
\draw  [fill={rgb, 255:red, 155; green, 155; blue, 155 }  ,fill opacity=0.56 ] (651.27,206.64) .. controls (651.27,202.83) and (655.12,199.74) .. (659.87,199.74) .. controls (664.61,199.74) and (668.46,202.83) .. (668.46,206.64) .. controls (668.46,210.45) and (664.61,213.53) .. (659.87,213.53) .. controls (655.12,213.53) and (651.27,210.45) .. (651.27,206.64) -- cycle ;
\draw  [fill={rgb, 255:red, 248; green, 231; blue, 28 }  ,fill opacity=0.39 ] (692.17,141.23) .. controls (692.17,137.42) and (696.02,134.33) .. (700.77,134.33) .. controls (705.51,134.33) and (709.36,137.42) .. (709.36,141.23) .. controls (709.36,145.04) and (705.51,148.13) .. (700.77,148.13) .. controls (696.02,148.13) and (692.17,145.04) .. (692.17,141.23) -- cycle ;
\draw  [fill={rgb, 255:red, 248; green, 231; blue, 28 }  ,fill opacity=0.39 ] (692.17,193.67) .. controls (692.17,189.86) and (696.02,186.77) .. (700.77,186.77) .. controls (705.51,186.77) and (709.36,189.86) .. (709.36,193.67) .. controls (709.36,197.48) and (705.51,200.56) .. (700.77,200.56) .. controls (696.02,200.56) and (692.17,197.48) .. (692.17,193.67) -- cycle ;
\draw    (672.38,127.93) -- (689.35,138.32) ;
\draw    (689.62,145.16) -- (672.43,155.32) ;
\draw    (672.38,179.31) -- (689.35,189.7) ;
\draw    (689.62,196.55) -- (672.43,206.7) ;
\draw [line width=1.5]  [dash pattern={on 1.69pt off 2.76pt}]  (700.89,164.08) -- (700.91,171.98) ;
\draw    (672.3,174.05) -- (689.62,149.55) ;
\draw    (689.48,183.88) -- (672.44,160.03) ;
\draw    (672.38,132.75) -- (689.35,176.7) ;
\draw    (672.38,200.09) -- (689.62,156.57) ;
\draw  [fill={rgb, 255:red, 248; green, 231; blue, 28 }  ,fill opacity=0.39 ] (751.98,141.16) .. controls (751.98,137.35) and (755.83,134.26) .. (760.57,134.26) .. controls (765.32,134.26) and (769.16,137.35) .. (769.16,141.16) .. controls (769.16,144.97) and (765.32,148.06) .. (760.57,148.06) .. controls (755.83,148.06) and (751.98,144.97) .. (751.98,141.16) -- cycle ;
\draw  [fill={rgb, 255:red, 248; green, 231; blue, 28 }  ,fill opacity=0.39 ] (751.98,193.59) .. controls (751.98,189.78) and (755.83,186.69) .. (760.57,186.69) .. controls (765.32,186.69) and (769.16,189.78) .. (769.16,193.59) .. controls (769.16,197.4) and (765.32,200.49) .. (760.57,200.49) .. controls (755.83,200.49) and (751.98,197.4) .. (751.98,193.59) -- cycle ;
\draw [line width=1.5]  [dash pattern={on 1.69pt off 2.76pt}]  (761.04,164.08) -- (761.06,171.98) ;
\draw    (713.62,141.24) -- (723.72,141.25) ;
\draw    (713.62,193.67) -- (723.72,193.69) ;
\draw    (713.62,146.64) -- (723.99,184.97) ;
\draw    (723.72,146.59) -- (713.35,184.91) ;
\draw  [dash pattern={on 4.5pt off 4.5pt}] (649.56,119.39) -- (670.1,119.39) -- (670.1,162.42) -- (649.56,162.42) -- cycle ;
\draw  [dash pattern={on 4.5pt off 4.5pt}] (649.56,171.53) -- (670.1,171.53) -- (670.1,214.56) -- (649.56,214.56) -- cycle ;
\draw    (736.99,141.24) -- (747.09,141.25) ;
\draw    (736.99,193.67) -- (747.09,193.69) ;
\draw    (736.99,146.64) -- (747.36,184.97) ;
\draw    (747.09,146.59) -- (736.72,184.91) ;
\draw [line width=1.5]  [dash pattern={on 1.69pt off 2.76pt}]  (732.79,141.28) -- (727.43,141.34) ;
\draw [line width=1.5]  [dash pattern={on 1.69pt off 2.76pt}]  (732.79,193.72) -- (727.43,193.78) ;
\draw  [fill={rgb, 255:red, 184; green, 233; blue, 134 }  ,fill opacity=0.55 ] (787.04,167.65) .. controls (787.04,163.84) and (790.88,160.75) .. (795.63,160.75) .. controls (800.38,160.75) and (804.22,163.84) .. (804.22,167.65) .. controls (804.22,171.46) and (800.38,174.55) .. (795.63,174.55) .. controls (790.88,174.55) and (787.04,171.46) .. (787.04,167.65) -- cycle ;
\draw    (771.71,142.11) -- (785.41,160.51) ;
\draw    (771.71,193.25) -- (785.41,174.86) ;
\draw    (288.75,284.33) -- (522,284.33) ;
\draw    (288.75,277.33) -- (288.75,284.33) ;
\draw    (522,277.33) -- (522,284.33) ;
\draw  [dash pattern={on 0.84pt off 2.51pt}]  (127.75,220.43) .. controls (127.67,256.18) and (355.02,263.34) .. (402.75,222.34) ;
\draw [shift={(404.8,220.43)}, rotate = 134.79] [fill={rgb, 255:red, 0; green, 0; blue, 0 }  ][line width=0.08]  [draw opacity=0] (8.93,-4.29) -- (0,0) -- (8.93,4.29) -- cycle    ;
\draw  [dash pattern={on 0.84pt off 2.51pt}]  (522,220.33) .. controls (521.87,223.88) and (538.68,239.69) .. (552.63,222.37) ;
\draw [shift={(554.36,220.03)}, rotate = 124.16] [fill={rgb, 255:red, 0; green, 0; blue, 0 }  ][line width=0.08]  [draw opacity=0] (8.93,-4.29) -- (0,0) -- (8.93,4.29) -- cycle    ;
\draw  [dash pattern={on 0.84pt off 2.51pt}]  (578.36,220.03) .. controls (578.23,223.57) and (595.04,239.38) .. (608.99,222.06) ;
\draw [shift={(610.72,219.72)}, rotate = 124.16] [fill={rgb, 255:red, 0; green, 0; blue, 0 }  ][line width=0.08]  [draw opacity=0] (8.93,-4.29) -- (0,0) -- (8.93,4.29) -- cycle    ;

\draw (111.75,21.4) node [anchor=north west][inner sep=0.75pt] [font=\Large]  [align=left] {$ \mathcal{D}^{( 1)}$};
\draw (391.08,22.4) node [anchor=north west][inner sep=0.75pt] [font=\Large]  [align=left] {$\mathcal{D}^{( 2)}$};
\draw (712.42,24.07) node [anchor=north west][inner sep=0.75pt] [font=\Large]  [align=left] {$\mathcal{D}^{( K)}$};
\draw (560,31) node [anchor=north west][inner sep=0.75pt]  [font=\small] [align=left] {...};
\draw (274.78,78.17) node [anchor=north west][inner sep=0.75pt]   [align=left] {$ \widehat{\lambda }_{i}^{( 1)}$};
\draw (97.75,289.25) node [anchor=north west][inner sep=0.75pt] [font=\Large]   [align=left] {$ f_{\text{occ}}^{( 1)}(\boldsymbol{x}_{i})$};
\draw (381,290.75) node [anchor=north west][inner sep=0.75pt] [font=\Large]  [align=left] {$ f_{\text{occ}}^{( 2)}(\boldsymbol{x}_{i})$};
\draw (560,165) node [anchor=north west][inner sep=0.75pt]  [font=\small] [align=left] {...};
\draw (244,235) node [anchor=north west][inner sep=0.75pt]  [font=\normalsize] [align=left] {weights};

\end{tikzpicture}
    
    }

    \caption{Visualisation of the consecutive steps in the expectation-maximisation algorithm for the occurrence model when using a neural network with weight initialisation in the maximisation step. The transfer of the network weights is indicated with a curved dashed line. The network structure used within each EM iteration corresponds to Figure~\ref{fig:NNocc} in Appendix~\ref{append:modelocc}.}
    \label{fig:convergenceNN}
\end{figure}
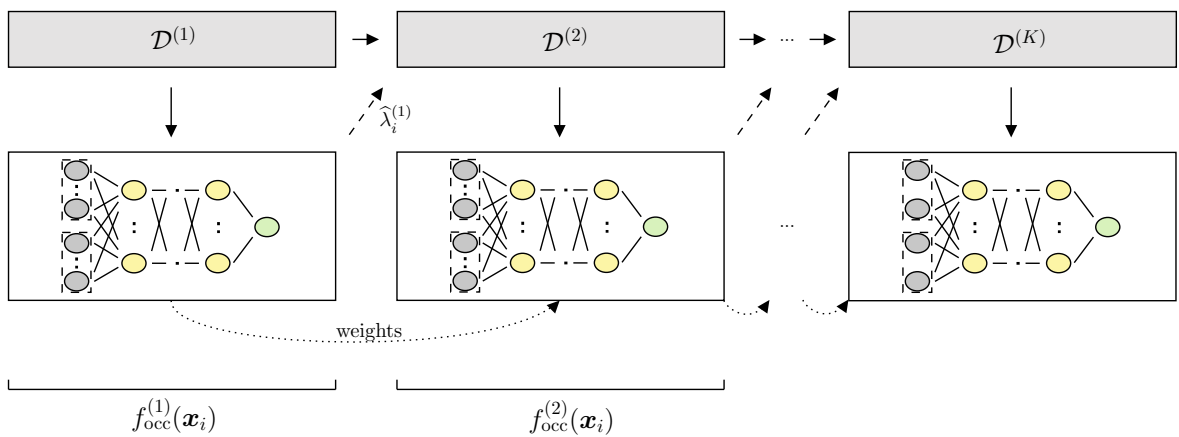

\paragraph{Neural network} We follow a solution proposed by \cite{delong2021gamma} in the context of mixture modelling. We initialise the network weights using the values obtained in the previous EM iteration. Consequently, we use the same model architecture in each iteration of the algorithm. This is reasonable since the data are only slightly modified across iterations via the estimates of the unobserved event counts. Additionally, it is time consuming to tune the model parameters in each iteration. Figure~\ref{fig:convergenceNN} visualises the approach for the occurrence intensities. We apply analogous reasoning for the reporting model. In iteration $k=1,\ldots,K$, separate neural networks for the occurrence intensities and reporting probabilities are trained on the current estimate of the complete information, i.e.~$\mathcal{D}^{(k)}=(\boldsymbol{N}_{i}^{(k)},\boldsymbol{x}_i)_{i=1}^{n}$. The (occurrence or reporting) neural network in the $k$th iteration consists of an input layer, $M^{(k)}$ hidden layers $\theta^{(k)}_{1},\ldots,\theta^{(k)}_{M^{(k)}}$ with $q^{(k)}_1,\ldots,q^{(k)}_{M^{(k)}}$ neurons respectively, and an output layer $\theta^{(k)}_{M^{(k)}+1}$. To maintain the same network structure across the EM algorithm iterations, we fix the number of hidden layers $M^{(k)}=M$ and the number of neurons in the $m$th layer $q^{(k)}_{m}=q_m$ for $k=1,\ldots,K$. The dimension of the input layer $q_0$ is equal to the dimension of the covariate vector $\boldsymbol{x}$. The dimension of the output layer is one- or $d$-dimensional for the occurrence or reporting model. Figures~\ref{fig:NNocc} and \ref{fig:NNrep} in Appendix~\ref{append:model} visualise the (feedforward) network architecture for the occurrence and reporting model, respectively.  Following the notation of \cite{schelldorfer2019nesting}, we denote the $u$th neuron in the $m$th hidden layer (in the $k$th iteration of the EM algorithm) as
\begin{align*}
    \theta_{m,u}^{(k)}(\boldsymbol{z})=\varphi^{(k)}_{m}\left(b_{m,u}^{(k)}+\left\langle\boldsymbol{w}_{m,u}^{(k)}, \boldsymbol{z}\right\rangle\right) \,\, \text{for} \,\, u=1,\ldots ,q_{m} \,\, \text{and} \,\, m=1,\ldots,M,
\end{align*}
where $\boldsymbol{z}$ is $q_{m-1}$-dimensional input. The trainable bias $b_{m,u}^{(k)} \in \mathbb{R}$ and the trainable weights $\boldsymbol{w}_{m,u}^{(k)} = (w_{m,u,1}^{(k)},\ldots,w_{m,u,q_{m-1}}^{(k)}) \in \mathbb{R}^{q_{m-1}}$ are the network parameters for the $m$th hidden layer $\theta^{(m)}$.  The $\varphi^{(k)}_m(.)$ is an activation function, i.e. a transformation that is applied to the output of each neuron in the $m$th hidden layer, allowing the network to learn non-linear effects. Let $\boldsymbol{w}_{m,u}^{(k,0)}$ denote the initialisation values for the trainable weights in the $k$th maximisation step. As suggested by \cite{delong2021gamma}, during network training, we initialise the weights $\boldsymbol{w}_{m,u}^{(k)}$ using the values of the previously calibrated weights $\boldsymbol{w}_{m,u}^{(k-1)}$, i.e. $\boldsymbol{w}_{m,u}^{(k,0)}=\boldsymbol{w}_{m,u}^{(k-1)}$ for $u=1,\ldots,q_{m} \,\, \text{and} \,\, m=1,\ldots,M$. This allows information to pass more effectively across iterations of the EM algorithm, see Figure~\ref{fig:convergenceNN}. We express the estimates for the occurrence intensities and reporting probabilities in the $k$th EM iteration as
\begin{align*}
    \widehat{\lambda}^{(k)}(\boldsymbol{x}_i)&=\exp\left(\widehat{b}^{(k)}_{M+1,1}+\left\langle\widehat{\boldsymbol{w}}^{(k)}_{M+1,1},\left(\widehat{\theta}^{(k)}_M \circ \cdots \circ \widehat{\theta}^{(k)}_1\right)(\boldsymbol{x}_i)\right\rangle\right), \:\:\: \text{and}\\
    \widehat{p}_{j}^{(k)}(\boldsymbol{x}_i)&=\frac{\exp\left(\widehat{b}^{(k)}_{M+1,j}+\left\langle\widehat{\boldsymbol{w}}^{(k)}_{M+1,j},\left(\widehat{\theta}^{(k)}_M \circ \cdots \circ \widehat{\theta}^{(k)}_1\right)(\boldsymbol{x}_i)\right\rangle\right)}{\displaystyle \sum_{z=1}^{d} \exp\left(\widehat{b}^{(k)}_{M+1,z}+\left\langle\widehat{\boldsymbol{w}}^{(k)}_{M+1,z},\left(\widehat{\theta}^{(k)}_M \circ \cdots \circ \widehat{\theta}^{(k)}_1\right)(\boldsymbol{x}_i)\right\rangle\right)},
\end{align*}
where $b^{(k)}_{M+1,1}$ and $\boldsymbol{w}^{(k)}_{M+1,1}$ refer to the bias and weights (network parameters) in the output layer, respectively. Additional details for both the occurrence and reporting model are in Appendix~\ref{append:modelocc} and \ref{append:modelrep}.

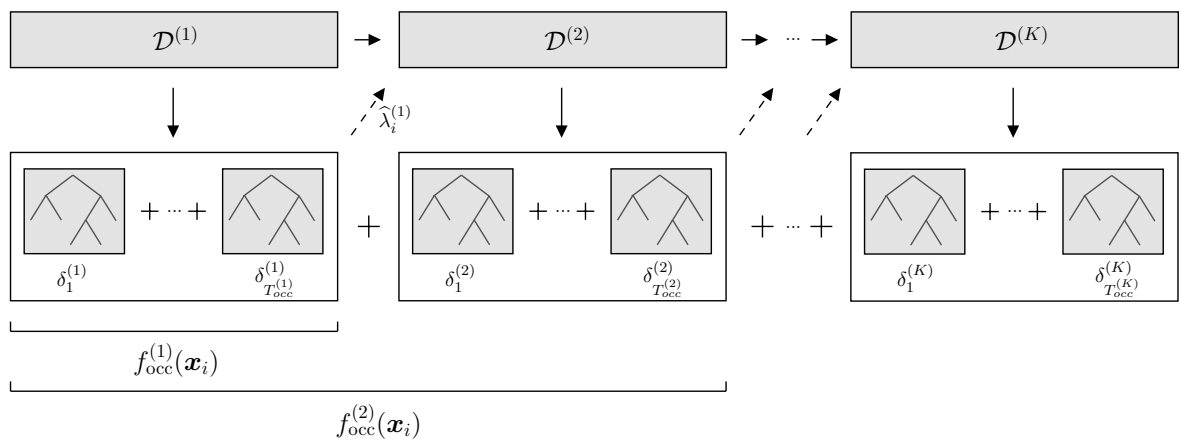
\begin{figure}[H]
    \centering

    \scalebox{0.7}{

\tikzset{every picture/.style={line width=0.75pt}} 

\begin{tikzpicture}[x=0.75pt,y=0.75pt,yscale=-1,xscale=1]

\draw    (56.25,130.06) -- (36.25,145.25) ;
\draw    (56.25,130.06) -- (76.25,145.25) ;

\draw    (76.06,145.06) -- (65.08,162.9) ;
\draw    (76.06,145.06) -- (87.05,162.9) ;

\draw    (37.06,144.56) -- (26.08,162.4) ;
\draw    (37.06,144.56) -- (48.05,162.4) ;

\draw    (65.4,162.31) -- (54.42,180.15) ;
\draw    (65.4,162.31) -- (76.38,180.15) ;

\draw  [fill={rgb, 255:red, 193; green, 190; blue, 190 }  ,fill opacity=0.43 ] (21.5,125.58) -- (93,125.58) -- (93,186.5) -- (21.5,186.5) -- cycle ;
\draw   (104.75,156) -- (117.25,156)(111,149.83) -- (111,162.17) ;
\draw    (197.75,130.06) -- (177.75,145.25) ;
\draw    (197.75,130.06) -- (217.75,145.25) ;

\draw    (217.56,145.06) -- (206.58,162.9) ;
\draw    (217.56,145.06) -- (228.55,162.9) ;

\draw    (178.56,144.56) -- (167.58,162.4) ;
\draw    (178.56,144.56) -- (189.55,162.4) ;

\draw    (206.9,162.31) -- (195.92,180.15) ;
\draw    (206.9,162.31) -- (217.88,180.15) ;

\draw  [fill={rgb, 255:red, 193; green, 190; blue, 190 }  ,fill opacity=0.43 ] (163,125.58) -- (234.5,125.58) -- (234.5,186.5) -- (163,186.5) -- cycle ;
\draw   (138.5,156) -- (151,156)(144.75,149.83) -- (144.75,162.17) ;
\draw   (11.75,113.78) -- (245,113.78) -- (245,220.33) -- (11.75,220.33) -- cycle ;
\draw    (333.25,130.06) -- (313.25,145.25) ;
\draw    (333.25,130.06) -- (353.25,145.25) ;

\draw    (353.06,145.06) -- (342.08,162.9) ;
\draw    (353.06,145.06) -- (364.05,162.9) ;

\draw    (314.06,144.56) -- (303.08,162.4) ;
\draw    (314.06,144.56) -- (325.05,162.4) ;

\draw    (342.4,162.31) -- (331.42,180.15) ;
\draw    (342.4,162.31) -- (353.38,180.15) ;

\draw  [fill={rgb, 255:red, 193; green, 190; blue, 190 }  ,fill opacity=0.43 ] (298.5,125.58) -- (370,125.58) -- (370,186.5) -- (298.5,186.5) -- cycle ;
\draw   (381.75,156) -- (394.25,156)(388,149.83) -- (388,162.17) ;
\draw    (474.75,130.06) -- (454.75,145.25) ;
\draw    (474.75,130.06) -- (494.75,145.25) ;

\draw    (494.56,145.06) -- (483.58,162.9) ;
\draw    (494.56,145.06) -- (505.55,162.9) ;

\draw    (455.56,144.56) -- (444.58,162.4) ;
\draw    (455.56,144.56) -- (466.55,162.4) ;

\draw    (483.9,162.31) -- (472.92,180.15) ;
\draw    (483.9,162.31) -- (494.88,180.15) ;

\draw  [fill={rgb, 255:red, 193; green, 190; blue, 190 }  ,fill opacity=0.43 ] (440,125.58) -- (511.5,125.58) -- (511.5,186.5) -- (440,186.5) -- cycle ;
\draw   (415.5,156) -- (428,156)(421.75,149.83) -- (421.75,162.17) ;
\draw   (288.75,113.78) -- (522,113.78) -- (522,220.33) -- (288.75,220.33) -- cycle ;
\draw   (259.42,166.94) -- (275.03,166.94)(267.22,159.03) -- (267.22,174.86) ;
\draw   (540.17,166.94) -- (555.78,166.94)(547.97,159.03) -- (547.97,174.86) ;
\draw   (582.17,166.94) -- (597.78,166.94)(589.97,159.03) -- (589.97,174.86) ;
\draw    (655.5,130.31) -- (635.5,145.5) ;
\draw    (655.5,130.31) -- (675.5,145.5) ;

\draw    (675.31,145.31) -- (664.33,163.15) ;
\draw    (675.31,145.31) -- (686.3,163.15) ;

\draw    (636.31,144.81) -- (625.33,162.65) ;
\draw    (636.31,144.81) -- (647.3,162.65) ;

\draw    (664.65,162.56) -- (653.67,180.4) ;
\draw    (664.65,162.56) -- (675.63,180.4) ;

\draw  [fill={rgb, 255:red, 193; green, 190; blue, 190 }  ,fill opacity=0.43 ] (620.75,125.83) -- (692.25,125.83) -- (692.25,186.75) -- (620.75,186.75) -- cycle ;
\draw   (704,156.25) -- (716.5,156.25)(710.25,150.08) -- (710.25,162.42) ;
\draw    (797,130.31) -- (777,145.5) ;
\draw    (797,130.31) -- (817,145.5) ;

\draw    (816.81,145.31) -- (805.83,163.15) ;
\draw    (816.81,145.31) -- (827.8,163.15) ;

\draw    (777.81,144.81) -- (766.83,162.65) ;
\draw    (777.81,144.81) -- (788.8,162.65) ;

\draw    (806.15,162.56) -- (795.17,180.4) ;
\draw    (806.15,162.56) -- (817.13,180.4) ;

\draw  [fill={rgb, 255:red, 193; green, 190; blue, 190 }  ,fill opacity=0.43 ] (762.25,125.83) -- (833.75,125.83) -- (833.75,186.75) -- (762.25,186.75) -- cycle ;
\draw   (737.75,156.25) -- (750.25,156.25)(744,150.08) -- (744,162.42) ;
\draw   (611,114.03) -- (844.25,114.03) -- (844.25,220.58) -- (611,220.58) -- cycle ;
\draw  [fill={rgb, 255:red, 193; green, 190; blue, 190 }  ,fill opacity=0.43 ] (11.75,12.78) -- (245,12.78) -- (245,52.78) -- (11.75,52.78) -- cycle ;
\draw  [fill={rgb, 255:red, 193; green, 190; blue, 190 }  ,fill opacity=0.43 ] (289,12.78) -- (522.25,12.78) -- (522.25,52.78) -- (289,52.78) -- cycle ;
\draw  [fill={rgb, 255:red, 193; green, 190; blue, 190 }  ,fill opacity=0.43 ] (611.25,12.78) -- (844.5,12.78) -- (844.5,52.78) -- (611.25,52.78) -- cycle ;
\draw    (257,32.81) -- (273,32.81) ;
\draw [shift={(276,32.81)}, rotate = 180] [fill={rgb, 255:red, 0; green, 0; blue, 0 }  ][line width=0.08]  [draw opacity=0] (8.93,-4.29) -- (0,0) -- (8.93,4.29) -- cycle    ;
\draw    (533,32.81) -- (549,32.81) ;
\draw [shift={(552,32.81)}, rotate = 180] [fill={rgb, 255:red, 0; green, 0; blue, 0 }  ][line width=0.08]  [draw opacity=0] (8.93,-4.29) -- (0,0) -- (8.93,4.29) -- cycle    ;
\draw    (583,32.81) -- (599,32.81) ;
\draw [shift={(602,32.81)}, rotate = 180] [fill={rgb, 255:red, 0; green, 0; blue, 0 }  ][line width=0.08]  [draw opacity=0] (8.93,-4.29) -- (0,0) -- (8.93,4.29) -- cycle    ;
\draw    (127.75,67.28) -- (127.75,100.28) ;
\draw [shift={(127.75,103.28)}, rotate = 270] [fill={rgb, 255:red, 0; green, 0; blue, 0 }  ][line width=0.08]  [draw opacity=0] (8.93,-4.29) -- (0,0) -- (8.93,4.29) -- cycle    ;
\draw    (404.75,67.28) -- (404.75,100.28) ;
\draw [shift={(404.75,103.28)}, rotate = 270] [fill={rgb, 255:red, 0; green, 0; blue, 0 }  ][line width=0.08]  [draw opacity=0] (8.93,-4.29) -- (0,0) -- (8.93,4.29) -- cycle    ;
\draw    (726.75,67.28) -- (726.75,100.28) ;
\draw [shift={(726.75,103.28)}, rotate = 270] [fill={rgb, 255:red, 0; green, 0; blue, 0 }  ][line width=0.08]  [draw opacity=0] (8.93,-4.29) -- (0,0) -- (8.93,4.29) -- cycle    ;
\draw  [dash pattern={on 4.5pt off 4.5pt}]  (254.75,101.28) -- (277.02,69.73) ;
\draw [shift={(278.75,67.28)}, rotate = 125.22] [fill={rgb, 255:red, 0; green, 0; blue, 0 }  ][line width=0.08]  [draw opacity=0] (8.93,-4.29) -- (0,0) -- (8.93,4.29) -- cycle    ;
\draw  [dash pattern={on 4.5pt off 4.5pt}]  (531.75,101.28) -- (554.02,69.73) ;
\draw [shift={(555.75,67.28)}, rotate = 125.22] [fill={rgb, 255:red, 0; green, 0; blue, 0 }  ][line width=0.08]  [draw opacity=0] (8.93,-4.29) -- (0,0) -- (8.93,4.29) -- cycle    ;
\draw  [dash pattern={on 4.5pt off 4.5pt}]  (579.75,101.28) -- (602.02,69.73) ;
\draw [shift={(603.75,67.28)}, rotate = 125.22] [fill={rgb, 255:red, 0; green, 0; blue, 0 }  ][line width=0.08]  [draw opacity=0] (8.93,-4.29) -- (0,0) -- (8.93,4.29) -- cycle    ;
\draw    (11.75,242.33) -- (245,242.33) ;
\draw    (11.75,235.33) -- (11.75,242.33) ;
\draw    (245,235.33) -- (245,242.33) ;
\draw    (11.75,285.33) -- (522,285.33) ;
\draw    (11.75,278.33) -- (11.75,285.33) ;
\draw    (522,278.33) -- (522,285.33) ;

\draw (43.25,192.58) node [anchor=north west][inner sep=0.75pt]   [align=left] {$\displaystyle \delta _{1}^{( 1)}$};
\draw (121.75,154) node [anchor=north west][inner sep=0.75pt]  [font=\small] [align=left] {...};
\draw (184.5,190) node [anchor=north west][inner sep=0.75pt]   [align=left] {$\displaystyle \delta _{T_{occ}^{( 1)}}^{( 1)}$};
\draw (320.25,192.58) node [anchor=north west][inner sep=0.75pt]   [align=left] {$\delta _{1}^{( 2)}$};
\draw (398.75,154) node [anchor=north west][inner sep=0.75pt]  [font=\small] [align=left] {...};
\draw (461.5,190) node [anchor=north west][inner sep=0.75pt]   [align=left] {$\delta _{T_{occ}^{( 2)}}^{( 2)}$};
\draw (562.89,165) node [anchor=north west][inner sep=0.75pt]  [font=\small] [align=left] {...};
\draw (642.5,192.58) node [anchor=north west][inner sep=0.75pt]   [align=left] {$\delta _{1}^{( K)}$};
\draw (721,154) node [anchor=north west][inner sep=0.75pt]  [font=\small] [align=left] {...};
\draw (783.75,190) node [anchor=north west][inner sep=0.75pt]   [align=left] {$\delta _{T_{occ}^{(K)}}^{( K)}$};
\draw (111.75,21.4) node [anchor=north west][inner sep=0.75pt]  [font=\Large] [align=left] {$\mathcal{D}^{( 1)}$};
\draw (391.08,21.4) node [anchor=north west][inner sep=0.75pt]  [font=\Large] [align=left] {$ \mathcal{D}^{( 2)}$};
\draw (712.42,21.4) node [anchor=north west][inner sep=0.75pt]  [font=\Large] [align=left] {$\mathcal{D}^{( K)}$};
\draw (562.89,30) node [anchor=north west][inner sep=0.75pt]  [font=\small] [align=left] {...};
\draw (273,78.17) node [anchor=north west][inner sep=0.75pt]   [align=left] {$\widehat{\lambda}_{i}^{( 1)}$};
\draw (97.75,249) node [anchor=north west][inner sep=0.75pt] [font=\Large]   [align=left] {$ f_{\text{occ}}^{( 1)}(\boldsymbol{x}_{i})$};
\draw (242,293) node [anchor=north west][inner sep=0.75pt] [font=\Large]  [align=left] {$f_{\text{occ}}^{( 2)}(\boldsymbol{x}_{i})$};

\end{tikzpicture}
    
    }

    \caption{Visualisation of the consecutive steps in the expectation-maximisation algorithm for the occurrence model when using an additive XGBoost model in the maximisation step. Regression trees are visualised with a generic tree structure. The `$+$' symbols indicate that regression trees are additive within an EM iteration as well as between EM iterations.}
    \label{fig:convergenceXGB}
\end{figure}

\paragraph{XGBoost} An XGBoost model is an ensemble of (shallow) regression trees \citep{chen2016xgboost}. Splits, i.e.~covariate thresholds that partition the data to optimise an objective function, are learned within each successive regression tree. In a tree ensemble, there are no trainable parameters such as e.g.~the number of hidden layers in a neural network, which we can initialise over successive EM iterations. Therefore, we opt to use an additive model, i.e.~we add additional regression trees to the previous iteration's model instead of calibrating a new model. In Figure~\ref{fig:convergenceXGB}, we illustrate the training steps in the EM algorithm for the occurrence intensities. The full details of the training algorithm for both the occurrence and the reporting model are in Appendix~\ref{append:model}. In the $k$th iteration, for the occurrence model, a sequence of $T_{\text{occ}}^{(k)}$ regression trees $\delta^{(k)}_{1},\ldots,\delta^{(k)}_{T_{\text{occ}}^{(k)}}$ with depth \texttt{tree\_depth\_occ} is trained on $\mathcal{D}^{(k)}=(\boldsymbol{N}_{i}^{(k)},\boldsymbol{x}_i)_{i=1}^{n}$. For the $j$th reporting probability, we learn a sequence of $T_{\text{rep}}^{(k)}$ regression trees $\delta^{(k)}_{1,j},\ldots,\delta^{(k)}_{T_{\text{rep}}^{(k)},j}$ with depth \texttt{tree\_depth\_rep}. Consequently, we can express the estimates for the occurrence intensities and reporting probabilities in the $k$th EM iteration as
\begin{align*}
    \widehat{\lambda}^{(k)}(\boldsymbol{x}_i)&= \exp\left(\widehat{f}^{(k-1)}_{\text{occ}}(\boldsymbol{x}_i) + \sum_{t=1}^{T_{\text{occ}}^{(k)}} \texttt{eta\_occ} 
\cdot \widehat{\delta}^{(k)}_{t}(\boldsymbol{x}_i)\right), \:\:\: \text{and}\\
    \widehat{p}_{j}^{(k)}(\boldsymbol{x}_i)&=\frac{\exp\left(\widehat{f}_{\text{rep}}^{(k-1)[j]}(\boldsymbol{x}_{i})+\displaystyle \sum_{t=1}^{T_{\text{rep}}^{(k)}}\texttt{eta\_rep} \cdot \widehat{\delta}_{t,j}^{(k)}(\boldsymbol{x}_{i})\right)}{\displaystyle \sum_{z=1}^{d}\exp\left(\widehat{f}_{\text{rep}}^{(k-1)[z]}(\boldsymbol{x}_{i})+\displaystyle \sum_{t=1}^{T_{\text{rep}}^{(k)}}\texttt{eta\_rep} \cdot \widehat{\delta}_{t,z}^{(k)}(\boldsymbol{x}_{i})\right)},
\end{align*}
where the learning rates \texttt{eta\_occ} and \texttt{eta\_rep} control the rate at which the estimates change when a new regression tree is added to the occurrence and the reporting model, respectively. As illustrated in Figure~\ref{fig:convergenceXGB}, this allows to incrementally adapt the model to the changing input data.

\begin{figure}[H] 
\centering
\subfloat{\includegraphics[width = 0.50\textwidth]{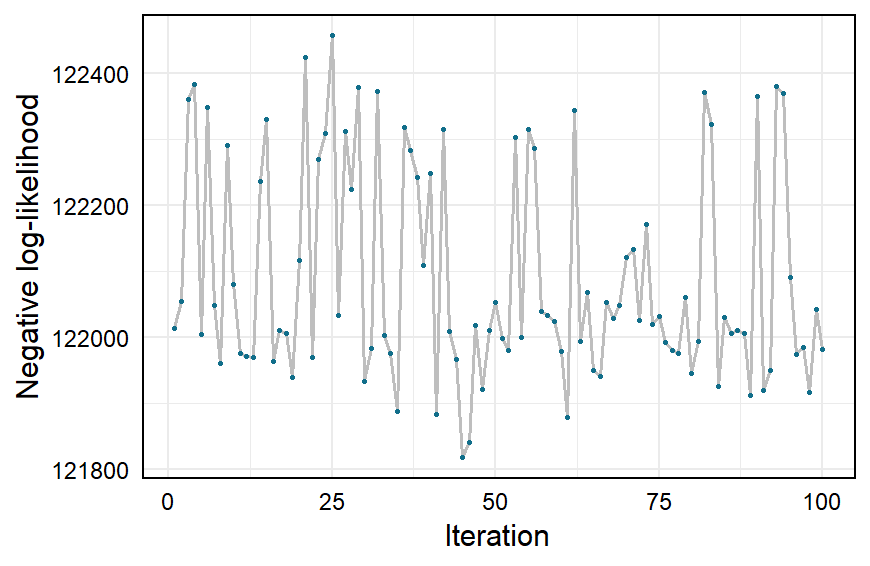}} 
\subfloat{\includegraphics[width = 0.50\textwidth]{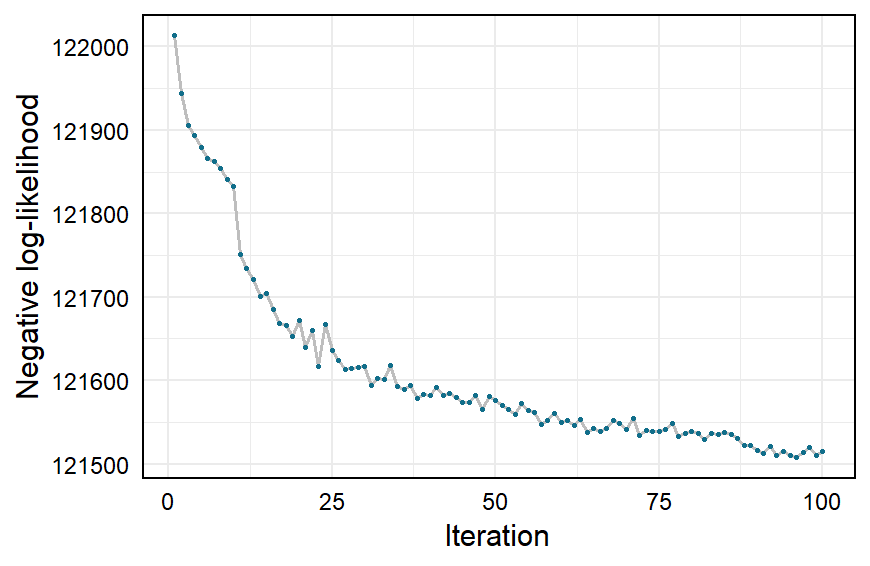}}
\caption{Example of EM training process when using neural networks. The left- and the right-hand side picture the observed negative log-likelihood $-LL(\widehat{\lambda}^{(k)}_i, \widehat{p}^{(k)}_{i,j};D)$ across iterations $k=1,\ldots,K=100$ for an approach with no weight initialisation (left) and weight initialisation (right), respectively.} 
\label{fig:oscillation_nn}
\end{figure}

\begin{figure}[H] 
\centering
\subfloat{\includegraphics[width = 0.50\textwidth]{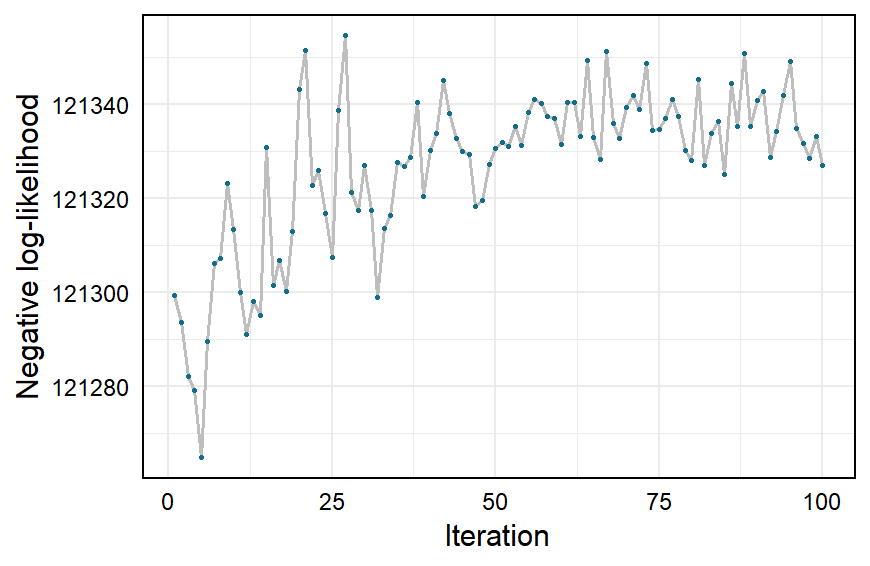}} 
\subfloat{\includegraphics[width = 0.50\textwidth]{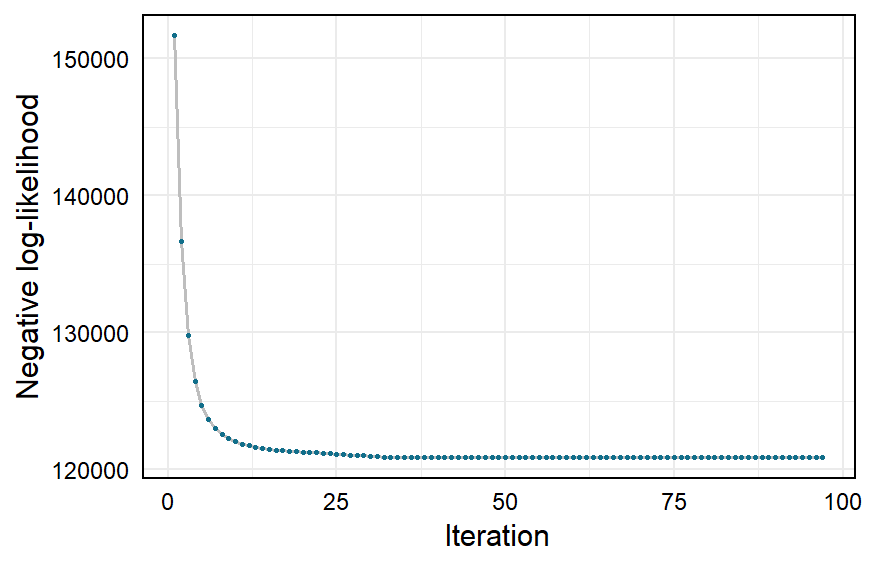}}
\caption{Example of EM training process when using XGBoost. The left- and the right-hand side picture the observed negative log-likelihood $-LL(\widehat{\lambda}^{(k)}_i, \widehat{p}^{(k)}_{i,j};D)$ across iterations $k=1,\ldots,K=100$ for a non-additive and an additive approach, respectively.} 
\label{fig:oscillation}
\end{figure}

\paragraph{Convergence} 
As stated before, we do not have mathematical guarantees that optimisation converges to the global optimum when using machine learning models. As a result, when applying the framework as shown in Figure~\ref{fig:summary}, we may observe oscillation in the observed likelihood over the EM iterations, as illustrated in the left panels of Figures~\ref{fig:oscillation_nn} and \ref{fig:oscillation}. In Figure~\ref{fig:oscillation_nn}, we illustrate the effect of implementing the weight initialisation scheme of \cite{delong2021gamma} when training the neural networks in the maximisation step of the EM algorithm. Specifically, we compute the observed likelihood $LL(\widehat{\lambda}^{(k)}_i, \widehat{p}^{(k)}_{i,j};D)$ for an example from Section~\ref{sec:simul} with $K=100$ iterations. On the left-hand side, we illustrate the results when using neural networks without weight initialisation. On the right-hand side, we follow the approach sketched in Figure~\ref{fig:convergenceNN}. We see that the training process stabilises when including weight initialisation, indicating that training stability is improved. However, there is still some oscillation visible in the observed log-likelihood. Therefore, additional early stopping rules need to be implemented as suggested by \cite{delong2021gamma}.

Figure~\ref{fig:oscillation} shows the same example as Figure~\ref{fig:oscillation_nn} when using XGBoost models in the maximisation step. The left panel shows the observed negative log-likelihood $-LL(\widehat{\lambda}^{(k)}_i, \widehat{p}^{(k)}_{i,j};D)$ under the non-additive approach, i.e. when a new XGBoost model is trained in each iteration. The right panel shows the same values when applying the additive approach, where regression trees are added in each iteration to a single XGBoost model as visualised in Figure~\ref{fig:convergenceXGB}. To allow for a fair comparison, we ensure that the model depth is the same for the final XGBoost model under both approaches for both the occurrence and reporting model, respectively. Specifically, under the additive scheme, we learn the same total number of regression trees across all EM iterations as in a single iteration of the non-additive scheme. This results in reduced training time under the additive scheme. Again, we observe oscillation in the observed log-likelihood when the EM algorithm is not specifically tailored to the model used (left panel). When including the additive scheme illustrated in Figure~\ref{fig:convergenceXGB}, shown in the right panel of Figure~\ref{fig:oscillation}, we observe that there is convergence without oscillation in the log-likelihood. This is because we construct a fully additive XGBoost model, which is not the case for the neural networks.

\vspace{-0.8cm}
\paragraph{Initialisation} The EM algorithm starts with an expectation step, that is: we need (initial) estimates for the missing data. Consequently, we need starting values $\lambda^{(0)}_{i}$ and $p^{(0)}_{i,j}$ for the occurrence intensities and reporting probabilities, respectively. We opt to use the empirical frequency of the event counts within the observed data $D$ to determine these initial values. Specifically, we set
\begin{align*}
    \lambda^{(0)}_{i} = \sum_{j=1}^{\tau_i} N_{i,j} \:\:\: \text{and} \:\:\: p^{(0)}_{i,j} = \frac{\sum_{z=1}^n N_{z,j}}{\sum_{z=1}^n \sum_{j=1}^{\tau_i} N_{z,j}}.
\end{align*}
Subsequently, we initialise the EM algorithm using these parameter estimates by constructing $\mathcal{D}^{(1)}$ in the first expectation step.

\paragraph{Early stopping} Machine learning models are prone to overfitting. Therefore, we split our dataset in a training and validation set. The former is used to learn the model parameters while the latter is used to monitor the training process, i.e.~training is stopped when there is deterioration for an evaluation criterion on that set. In an EM setting, we need to account for deterioration in the log-likelihood within the maximisation step as well as across iterations of the EM algorithm. Consequently, in an approach similar to \cite{delong2021gamma}, we split the data $D$ into three parts: a training set and two validation sets. We use the first validation set for early stopping during model training in the $k$th maximisation step. Specifically, we evaluate $Q^{\text{occ}}(\widehat{\lambda}_{i}^{(k)};\mathcal{D}^{(k)})$ and $Q^{\text{rep}}(\widehat{p}_{i,j}^{(k)};\mathcal{D}^{(k)})$ in each model training step for the occurrence and reporting model, respectively. Model training is stopped if there is no improvement for a specified number of iterations (patience), indicating overfitting. We denote the patience as \texttt{nn\_patience} and \texttt{xgb\_patience} when using neural networks or XGBoost models, respectively. The second validation set is used to monitor the performance of the EM algorithm itself. After each iteration we evaluate the observed log-likelihood $LL(\widehat{\lambda}^{(k)}_i, \widehat{p}^{(k)}_{i,j};D)$. If this log-likelihood deteriorates within a specified number of EM iterations, denoted as \texttt{EM\_patience}, the EM algorithm is stopped. The full details are in Algorithm \ref{algo:EM}, \ref{algo:XGBocc} and \ref{algo:XGBrep} in Appendix~\ref{append:model}.

\paragraph{Parameter tuning}
Machine learning models are flexible compared to generalised linear models, as they do not require to explicitly define the functional form of the covariate effect on the response variable. However, their training is governed by a set of parameters, which we refer to as tuning- and hyperparameters, depending on whether they are predetermined or not, respectively. For the tuning parameters, we opt to select values by conducting a random grid search \citep{bergstra2012random}. Specifically, we choose a set of possible values for each tuning parameter and consider a random subset of all different possible configurations. For both the neural network and XGBoost model, we select the combination of tuning parameter values that results in the highest observed log-likelihood $LL(\widehat{\lambda}_i,\widehat{p}_{i,j};D)$, computed on the part of the data that is used for validation across the EM iterations. The early stopping rules laid out in the previous paragraph are also applied to this dataset. The chosen tuning values are subsequently used for the actual training of the occurrence and reporting model.

\section{Simulation experiments} \label{sec:simul}
Using simulation experiments, we assess the EM framework's performance in retrieving the true underlying occurrence intensities and reporting probabilities under the presence of linear as well as non-linear effects. For both experimental settings, we investigate the use of different statistical and machine learning models, i.e.~GLMs, XGBoost models and neural networks, in the maximisation step. Specifically, for each model, we assess the overall (out-of-sample) performance as well as the performance for the occurrence- and reporting-specific parts of the framework.

\subsection{Model training and evaluation}
For both the experiment including only linear effects and the experiment with both linear and non-linear effects, we repeat the experiment 100 times, i.e.~we simulate 100 (independent) datasets $\mathcal{D}_{1},\ldots,\mathcal{D}_{100}$ with $n=10\ 000$ observations. Each time the EM algorithm is applied to a dataset, part of the observations are used for training (64\%), early stopping during model training (16\%) and early stopping of the EM algorithm (20\%) as laid out in Section~\ref{sec:method}. We opt to apply the same tuning parameter selection to each of the 100 datasets, since tuning each dataset separately would introduce variation in the model performance due to the chosen tuning values. Therefore, we simulate an additional dataset $\mathcal{D}_{\text{tune}}$ with $n=40\ 000$ observations to determine the tuning parameter selection, which is subsequently applied to each of the 100 datasets.

Table~\ref{tab:tuning1} and Table~\ref{tab:tuning2} in Appendix~\ref{app:figures} show the tuning grid as well as the selected model settings for the experiment with only linear effects and the experiment with both linear and non-linear effects, respectively. For the neural network, we fix the number of epochs (\texttt{n\_epoch}) at 50 while varying the batch size (\texttt{batch\_size}). This ensures relatively small parameter changes, allowing the EM algorithm to incrementally update the occurrence intensities and reporting probabilities. Additionally, we set the number of hidden layers $M=2$, to allow a fixed network architecture for the transfer of weights across the EM iterations as explained in Section~\ref{sec:method}. In all settings, we set the number of EM iterations $K=100$. The overall patience within the training process of the EM algorithm itself is set to 10, i.e.~training will stop early if there is no improvement in the observed log-likelihood for 10 consecutive EM iterations.

We simulate an additional dataset $\mathcal{D}_{\text{test}}$ consisting of $n=20\ 000$ observations to assess the out-of-sample performance of the trained (occurrence and reporting) models. For each of 100 experiment repetitions, the estimates for the occurrence intensities and reporting probabilities for the observations in the test set $\mathcal{D}_{\text{test}}$ are computed using the models trained on $\mathcal{D}_{1},\ldots,\mathcal{D}_{100}$. To summarise, 102 datasets are generated for each experimental setting: 100 datasets to learn the occurrence and reporting models, one dataset for tuning parameter selection and one dataset for out-of-sample evaluation. All generated datasets have the same structure, upon which we elaborate in the next section.

\subsection{Data generation} \label{sec:data}

As discussed in Section~\ref{sec:notation}, in our proposed framework, each observation $i$ corresponds to a unique combination of an entity $\text{ent}(i)$ and an occurrence period $\text{occ}(i)$. We characterise these entities and occurrence periods by simulating the covariate vectors $\boldsymbol{x}_{\text{ent}(i)}$ and $\boldsymbol{x}_{\text{per}(i)}$. Subsequently, given the covariate information $\boldsymbol{x}_i=(\boldsymbol{x}_{\text{ent}(i)},\boldsymbol{x}_{\text{per}(i)})$, we specify $\lambda(\boldsymbol{x}_i)$ and $p_j(\boldsymbol{x}_i)$ with $j=1,\ldots,d$. We opt for a maximum reporting delay of ten days, i.e.~$d=11$. Following Section~\ref{sec:notation}, the specified occurrence and reporting dynamics are then combined to simulate the full (unobserved) event information $\boldsymbol{N}_i=(N_{i,1},\ldots,N_{i,d})$. We use the superscript entity-specific (es) and period-specific (ps) to indicate whether a variable contains entity- or period-related information.

\subsubsection{Covariate information}

\begin{figure}[H]
    \centering

    \scalebox{0.95}{

\tikzset{every picture/.style={line width=0.75pt}} 

\begin{tikzpicture}[x=0.75pt,y=0.75pt,yscale=-1,xscale=1]

\draw  [color={rgb, 255:red, 0; green, 0; blue, 0 }  ,draw opacity=0.15 ][fill={rgb, 255:red, 0; green, 0; blue, 0 }  ,fill opacity=0.15 ] (3.7,26.78) -- (353.36,26.78) -- (353.36,16.9) -- (3.7,16.9) -- cycle ;
\draw    (3.7,21.9) -- (353.7,21.9) ;
\draw    (3.7,16.9) -- (3.7,26.78) ;
\draw    (353.7,16.9) -- (353.7,26.78) ;
\draw    (53.7,16.9) -- (53.7,26.78) ;
\draw    (103.7,16.9) -- (103.7,26.78) ;
\draw    (153.7,16.9) -- (153.7,26.78) ;
\draw    (203.7,16.9) -- (203.7,26.78) ;
\draw    (253.7,16.9) -- (253.7,26.78) ;
\draw    (303.7,16.9) -- (303.7,26.78) ;
\draw  [color={rgb, 255:red, 0; green, 0; blue, 0 }  ,draw opacity=0.15 ][fill={rgb, 255:red, 0; green, 0; blue, 0 }  ,fill opacity=0.15 ] (3.7,62.78) -- (353.36,62.78) -- (353.36,52.9) -- (3.7,52.9) -- cycle ;
\draw    (3.7,57.9) -- (353.7,57.9) ;
\draw    (3.7,52.9) -- (3.7,62.78) ;
\draw    (353.7,52.9) -- (353.7,62.78) ;
\draw    (53.7,52.9) -- (53.7,62.78) ;
\draw    (103.7,52.9) -- (103.7,62.78) ;
\draw    (153.7,52.9) -- (153.7,62.78) ;
\draw    (203.7,52.9) -- (203.7,62.78) ;
\draw    (253.7,52.9) -- (253.7,62.78) ;
\draw    (303.7,52.9) -- (303.7,62.78) ;
\draw  [color={rgb, 255:red, 0; green, 0; blue, 0 }  ,draw opacity=0.15 ][fill={rgb, 255:red, 0; green, 0; blue, 0 }  ,fill opacity=0.15 ] (3.7,98.78) -- (353.36,98.78) -- (353.36,88.9) -- (3.7,88.9) -- cycle ;
\draw    (3.7,93.9) -- (353.7,93.9) ;
\draw    (3.7,88.9) -- (3.7,98.78) ;
\draw    (353.7,88.9) -- (353.7,98.78) ;
\draw    (53.7,88.9) -- (53.7,98.78) ;
\draw    (103.7,88.9) -- (103.7,98.78) ;
\draw    (153.7,88.9) -- (153.7,98.78) ;
\draw    (203.7,88.9) -- (203.7,98.78) ;
\draw    (253.7,88.9) -- (253.7,98.78) ;
\draw    (303.7,88.9) -- (303.7,98.78) ;
\draw    (3.7,129.9) -- (353.7,129.9) ;
\draw    (3.7,124.9) -- (3.7,134.78) ;
\draw    (353.7,124.9) -- (353.7,134.78) ;
\draw    (53.7,124.9) -- (53.7,134.78) ;
\draw    (103.7,124.9) -- (103.7,134.78) ;
\draw    (153.7,124.9) -- (153.7,134.78) ;
\draw    (203.7,124.9) -- (203.7,134.78) ;
\draw    (253.7,124.9) -- (253.7,134.78) ;
\draw    (303.7,124.9) -- (303.7,134.78) ;
\draw    (3.7,165.15) -- (153.52,165.57) ;
\draw    (3.7,160.15) -- (3.7,170.03) ;
\draw    (53.7,160.15) -- (53.7,170.03) ;
\draw    (103.7,160.15) -- (103.7,170.03) ;
\draw  [fill={rgb, 255:red, 0; green, 0; blue, 0 }  ,fill opacity=1 ] (257.23,30.89) .. controls (257.23,30) and (257.95,29.28) .. (258.84,29.28) .. controls (259.73,29.28) and (260.44,30) .. (260.44,30.89) .. controls (260.44,31.77) and (259.73,32.49) .. (258.84,32.49) .. controls (257.95,32.49) and (257.23,31.77) .. (257.23,30.89) -- cycle ;
\draw  [fill={rgb, 255:red, 0; green, 0; blue, 0 }  ,fill opacity=1 ] (209.55,29.71) -- (211.27,32.49) -- (207.84,32.49) -- cycle ;
\draw  [fill={rgb, 255:red, 0; green, 0; blue, 0 }  ,fill opacity=1 ] (263.84,101.22) -- (267.11,101.22) -- (267.11,104.49) -- (263.84,104.49) -- cycle ;
\draw    (3.7,160.65) -- (3.7,170.53) ;
\draw    (53.7,160.65) -- (53.7,170.53) ;
\draw    (103.7,160.65) -- (103.7,170.53) ;
\draw    (153.7,160.65) -- (153.7,170.53) ;
\draw  [fill={rgb, 255:red, 0; green, 0; blue, 0 }  ,fill opacity=1 ] (307.23,30.89) .. controls (307.23,30) and (307.95,29.28) .. (308.84,29.28) .. controls (309.73,29.28) and (310.44,30) .. (310.44,30.89) .. controls (310.44,31.77) and (309.73,32.49) .. (308.84,32.49) .. controls (307.95,32.49) and (307.23,31.77) .. (307.23,30.89) -- cycle ;
\draw  [fill={rgb, 255:red, 0; green, 0; blue, 0 }  ,fill opacity=1 ] (257.23,66.89) .. controls (257.23,66) and (257.95,65.28) .. (258.84,65.28) .. controls (259.73,65.28) and (260.44,66) .. (260.44,66.89) .. controls (260.44,67.77) and (259.73,68.49) .. (258.84,68.49) .. controls (257.95,68.49) and (257.23,67.77) .. (257.23,66.89) -- cycle ;
\draw  [fill={rgb, 255:red, 0; green, 0; blue, 0 }  ,fill opacity=1 ] (307.23,66.89) .. controls (307.23,66) and (307.95,65.28) .. (308.84,65.28) .. controls (309.73,65.28) and (310.44,66) .. (310.44,66.89) .. controls (310.44,67.77) and (309.73,68.49) .. (308.84,68.49) .. controls (307.95,68.49) and (307.23,67.77) .. (307.23,66.89) -- cycle ;
\draw  [fill={rgb, 255:red, 0; green, 0; blue, 0 }  ,fill opacity=1 ] (257.23,102.89) .. controls (257.23,102) and (257.95,101.28) .. (258.84,101.28) .. controls (259.73,101.28) and (260.44,102) .. (260.44,102.89) .. controls (260.44,103.77) and (259.73,104.49) .. (258.84,104.49) .. controls (257.95,104.49) and (257.23,103.77) .. (257.23,102.89) -- cycle ;
\draw  [fill={rgb, 255:red, 0; green, 0; blue, 0 }  ,fill opacity=1 ] (307.23,102.89) .. controls (307.23,102) and (307.95,101.28) .. (308.84,101.28) .. controls (309.73,101.28) and (310.44,102) .. (310.44,102.89) .. controls (310.44,103.77) and (309.73,104.49) .. (308.84,104.49) .. controls (307.95,104.49) and (307.23,103.77) .. (307.23,102.89) -- cycle ;
\draw  [fill={rgb, 255:red, 0; green, 0; blue, 0 }  ,fill opacity=1 ] (257.23,138.89) .. controls (257.23,138) and (257.95,137.28) .. (258.84,137.28) .. controls (259.73,137.28) and (260.44,138) .. (260.44,138.89) .. controls (260.44,139.77) and (259.73,140.49) .. (258.84,140.49) .. controls (257.95,140.49) and (257.23,139.77) .. (257.23,138.89) -- cycle ;
\draw  [fill={rgb, 255:red, 0; green, 0; blue, 0 }  ,fill opacity=1 ] (307.23,138.89) .. controls (307.23,138) and (307.95,137.28) .. (308.84,137.28) .. controls (309.73,137.28) and (310.44,138) .. (310.44,138.89) .. controls (310.44,139.77) and (309.73,140.49) .. (308.84,140.49) .. controls (307.95,140.49) and (307.23,139.77) .. (307.23,138.89) -- cycle ;
\draw  [fill={rgb, 255:red, 0; green, 0; blue, 0 }  ,fill opacity=1 ] (313.84,101.22) -- (317.11,101.22) -- (317.11,104.49) -- (313.84,104.49) -- cycle ;
\draw  [fill={rgb, 255:red, 0; green, 0; blue, 0 }  ,fill opacity=1 ] (8.55,65.71) -- (10.27,68.49) -- (6.84,68.49) -- cycle ;
\draw  [fill={rgb, 255:red, 0; green, 0; blue, 0 }  ,fill opacity=1 ] (315.55,29.71) -- (317.27,32.49) -- (313.84,32.49) -- cycle ;
\draw  [fill={rgb, 255:red, 0; green, 0; blue, 0 }  ,fill opacity=1 ] (109.55,101.71) -- (111.27,104.49) -- (107.84,104.49) -- cycle ;

\draw (25,4.5) node [anchor=north west][inner sep=0.75pt]  [font=\scriptsize] [align=left] {1};
\draw (75,4.5) node [anchor=north west][inner sep=0.75pt]  [font=\scriptsize] [align=left] {2};
\draw (125,4.5) node [anchor=north west][inner sep=0.75pt]  [font=\scriptsize] [align=left] {3};
\draw (175,4.5) node [anchor=north west][inner sep=0.75pt]  [font=\scriptsize] [align=left] {4};
\draw (225,4.5) node [anchor=north west][inner sep=0.75pt]  [font=\scriptsize] [align=left] {5};
\draw (275,4.5) node [anchor=north west][inner sep=0.75pt]  [font=\scriptsize] [align=left] {6};
\draw (325,4.5) node [anchor=north west][inner sep=0.75pt]  [font=\scriptsize] [align=left] {7};
\draw (25,41.5) node [anchor=north west][inner sep=0.75pt]  [font=\scriptsize] [align=left] {8};
\draw (75,41.5) node [anchor=north west][inner sep=0.75pt]  [font=\scriptsize] [align=left] {9};
\draw (121.4,41.5) node [anchor=north west][inner sep=0.75pt]  [font=\scriptsize] [align=left] {10};
\draw (172.6,41.5) node [anchor=north west][inner sep=0.75pt]  [font=\scriptsize] [align=left] {11};
\draw (221,41.5) node [anchor=north west][inner sep=0.75pt]  [font=\scriptsize] [align=left] {12};
\draw (273.4,41.5) node [anchor=north west][inner sep=0.75pt]  [font=\scriptsize] [align=left] {13};
\draw (322.2,41.5) node [anchor=north west][inner sep=0.75pt]  [font=\scriptsize] [align=left] {14};
\draw (21,116) node [anchor=north west][inner sep=0.75pt]  [font=\scriptsize] [align=left] {22};
\draw (71,116) node [anchor=north west][inner sep=0.75pt]  [font=\scriptsize] [align=left] {23};
\draw (121,116) node [anchor=north west][inner sep=0.75pt]  [font=\scriptsize] [align=left] {24};
\draw (171,116) node [anchor=north west][inner sep=0.75pt]  [font=\scriptsize] [align=left] {25};
\draw (221,116) node [anchor=north west][inner sep=0.75pt]  [font=\scriptsize] [align=left] {26};
\draw (271,116) node [anchor=north west][inner sep=0.75pt]  [font=\scriptsize] [align=left] {27};
\draw (321,116) node [anchor=north west][inner sep=0.75pt]  [font=\scriptsize] [align=left] {28};
\draw (21,153) node [anchor=north west][inner sep=0.75pt]  [font=\scriptsize] [align=left] {29};
\draw (71,153) node [anchor=north west][inner sep=0.75pt]  [font=\scriptsize] [align=left] {30};
\draw (121,153) node [anchor=north west][inner sep=0.75pt]  [font=\scriptsize] [align=left] {31};
\draw (21,79) node [anchor=north west][inner sep=0.75pt]  [font=\scriptsize] [align=left] {15};
\draw (71,79) node [anchor=north west][inner sep=0.75pt]  [font=\scriptsize] [align=left] {16};
\draw (121,79) node [anchor=north west][inner sep=0.75pt]  [font=\scriptsize] [align=left] {17};
\draw (171,79) node [anchor=north west][inner sep=0.75pt]  [font=\scriptsize] [align=left] {18};
\draw (221,79) node [anchor=north west][inner sep=0.75pt]  [font=\scriptsize] [align=left] {19};
\draw (271,79) node [anchor=north west][inner sep=0.75pt]  [font=\scriptsize] [align=left] {20};
\draw (310.5,79) node [anchor=north west][inner sep=0.75pt]  [font=\scriptsize] [align=left] {$\tau=21$};

\end{tikzpicture}
    
    }

    \caption{Visualisation of simulated occurrence and reporting days. Days in which events can occur are highlighted in grey. Weekend days are indicated with a black circle, holidays with a black triangle and the first/last days of the calender month with a black square. }
    \label{fig:occurrenceperiods}
\end{figure}
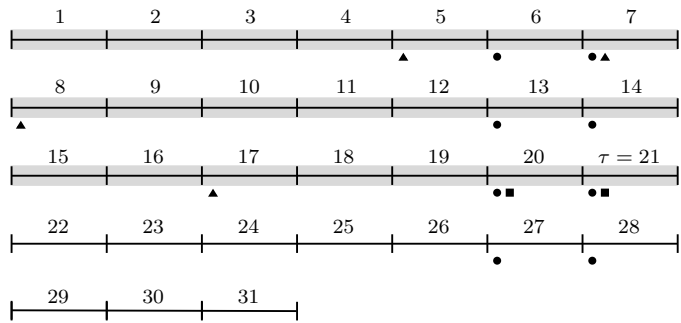

\paragraph{Occurrence- and reporting period-related information} We design a time window of 31 successive days $1,\ldots,31$, with the first period (day) being a Monday. The occurrence of events can happen in the first three weeks, i.e.~during the first 21 days. Since $d=11$, we also simulate days $22,\ldots,31$ because they are potential future reporting periods. For example, if an event occurs during day 21, it can be reported up to day $21+11-1=31$. The last occurrence period (day 21) is set to be the last observable period, i.e. $\tau=21$. For a given observation $i$, $\text{occ}(i)$ is the occurrence date and the set of possible reporting dates is given by $\text{occ}(i)+j-1$ with $j=1,\ldots,d$. In practical applications, the type of day (e.g.~weekend, holiday) impacts the occurrence and reporting of events as discussed in \cite{verbelen2022modeling}. Therefore, for each of the reporting dates (including the occurrence date), the covariate vector $\boldsymbol{x}_{\text{per}(i)}$ contains period-specific (ps) information: an indicator for a weekend day ($x_1^{(ps)}$), a variable indicating whether it is a holiday ($x_2^{(ps)}$) and an indicator for the first and last day of the calender month ($x_3^{(ps)}$). In Figure~\ref{fig:occurrenceperiods}, we visualise which of the 31 simulated days are weekend days, holidays or the first/last day of the month. For a given observation, we distinguish the covariates for each possible reporting date. Specifically, we use $x_1^{(ps,j)}$ to refer to $x_1^{(ps)}$ for day $\text{occ}(i)+j-1$ with $j=1,\ldots,11$, i.e.~$\textbf{x}_{\text{occ}(i)+j-1}=(x_1^{(ps,j)},x_2^{(ps,j)},x_3^{(ps,j)})$. Since we set $d=11$, the resulting covariate vector $\boldsymbol{x}_{\text{per}(i)}$ consists of 33 variables in total.

\paragraph{Entity-specific information} For a given observation $i$, the covariate vector $\boldsymbol{x}_{\text{ent}(i)}$ contains the entity-specific (es) information. In an epidemiological context, this can refer to regional characteristics such as the number of medical facilities or the number of cities. In insurance, this can be policyholder-specific information such as age or address. We simulate four entity-related variables: a categorical variable $x_1^{(es)}$ with two classes, a numerical variable $x_2^{(es)} \in \{18,\ldots,90\}$, a categorical variable $x_3^{(es)}$ with three classes and another categorical variable $x_4^{(es)}$ with three classes. For multi-class categorical variables, we define an indicator variable for each of the classes, e.g.~$x_{1,1}^{(es)}$ refers to the indicator variable for the first class of $x_1^{(es)}$. In an insurance application, for example, $x_1^{(es)},\ldots,x_4^{(es)}$ can refer to the gender, age, type of vehicle owned and bonus-malus scale of a policyholder, respectively.

\subsubsection{Response}
\paragraph{Generation event counts}
To simulate the full event information $\boldsymbol{N}_i=(N_{i,1},\ldots,N_{i,d})$, we first specify the occurrence intensities $\lambda(\boldsymbol{x}_i)= \exp(\boldsymbol{x}^{'}_{i}\boldsymbol{\beta}_{\lambda})$ and the reporting probabilities $p_j(\boldsymbol{x}_i)= \frac{\exp(\boldsymbol{x}^{'}_{i}\boldsymbol{\beta}_{p_j})}{\sum_{z=1}^{d} \exp(\boldsymbol{x}^{'}_{i}\boldsymbol{\beta}_{p_z})}$ with $j=1,\ldots,d$. The coefficient vectors are represented by $\boldsymbol{\beta}_{\lambda}$ and $\boldsymbol{\beta}_{p_j}$ for the occurrence and reporting model, respectively. We consider two different specifications for the coefficient vectors, which are given in Tables~\ref{tab:spec1} and~\ref{tab:spec2} in Appendix~\ref{app:figures}, respectively. The first specification includes only linear effects while the second one includes both linear and non-linear effects. Subsequently, using the specified occurrence intensity $\lambda(\boldsymbol{x}_i)$ and reporting probabilities $p_j(\boldsymbol{x}_i)$, we can simulate the full event information $\boldsymbol{N}_i$ in two steps. First, we use $N_i \sim \text{Poisson}(\lambda(\boldsymbol{x}_i))$ to simulate the number of occurred events for a given observation $i$. Subsequently, simulating $N_{i,j}|N_i \sim \text{Multinomial}(p_j(\boldsymbol{x}_i))$ with $j=1,..,d$ gives the number of occurred events that were reported for each of the possible reporting dates, resulting in the full event information $\boldsymbol{N}_i=(N_{i,1},\ldots,N_{i,d})$.  

\begin{figure}[H] 
\centering
\subfloat[Only linear effects]{

\begin{adjustbox}{width=0.45\textwidth}
    \includegraphics[width = \textwidth]{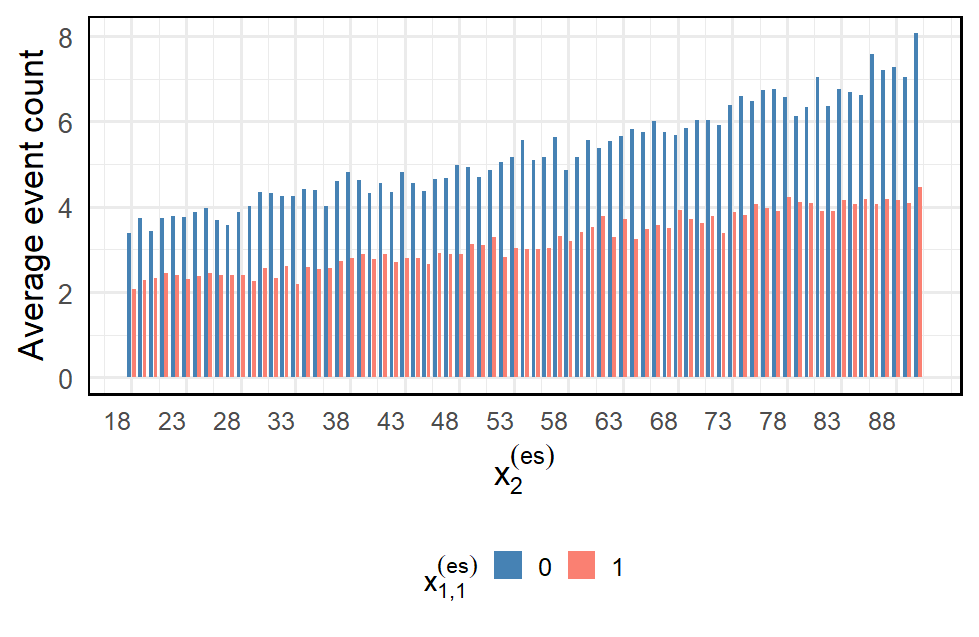}
\end{adjustbox}

}
\hspace{1.5em}
\subfloat[Linear and non-linear effects]{

\begin{adjustbox}{width=0.45\textwidth}
\includegraphics[width = \textwidth]{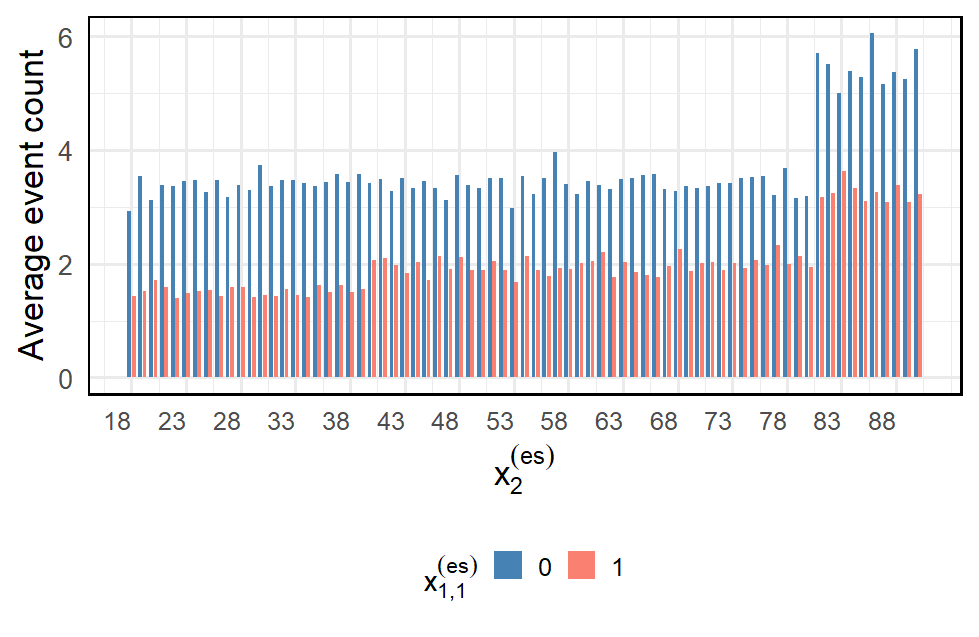}

\end{adjustbox}

}
\caption{Average of the event counts $N_i$ across different values of $x^{(es)}_{2}$ for $\mathcal{D}_{\text{tune}}$. The red bars show the average of the event occurrence counts $N_i$ for all observations with $x^{(es)}_{11}=0$ and the blue bars show the same for $x^{(es)}_{11}=1$. The left-hand panel shows the values for the setting with only linear effects. Conversely, the right-hand panel shows the values for the setting with both linear and non-linear effects.}   
\label{fig:exploratory}
\end{figure}

\begin{figure}[H] 
\centering
\subfloat[Occurrence structure]{

\begin{adjustbox}{width=0.45\textwidth}
    \includegraphics[width = \textwidth]{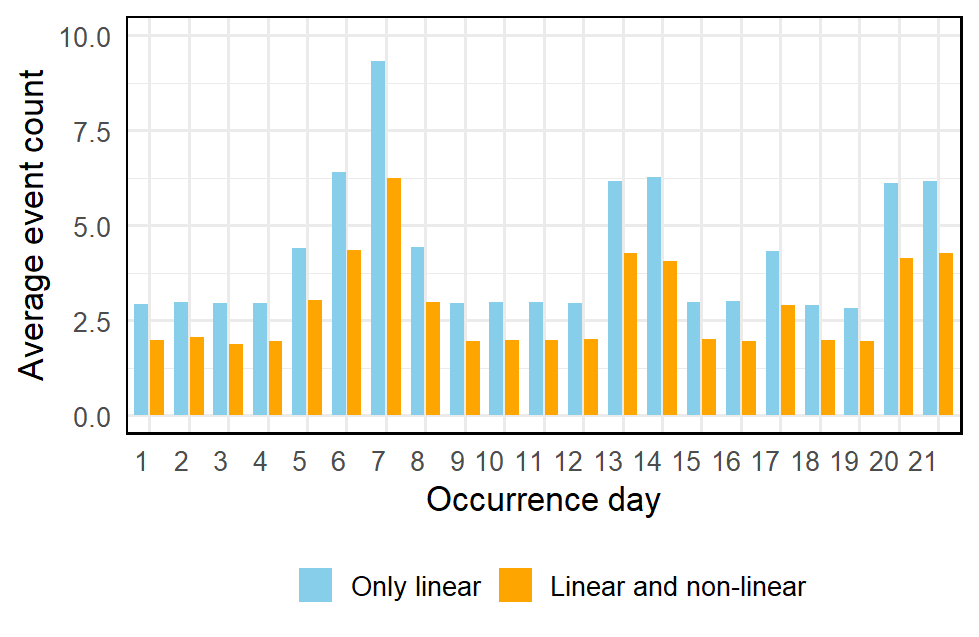}
\end{adjustbox}

}
\hspace{1.5em}
\subfloat[Reporting structure]{

\begin{adjustbox}{width=0.45\textwidth}
\includegraphics[width = \textwidth]{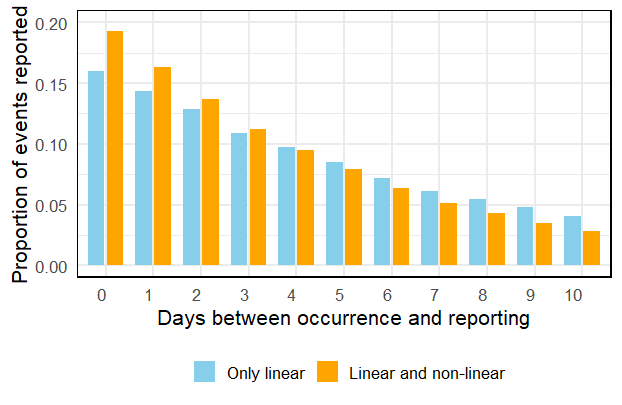}

\end{adjustbox}

}
\caption{ Occurrence and reporting pattern illustrated for both experiments on $\mathcal{D}_{\text{tune}}$. The left-hand panel shows the average number of event counts $N_i$ for each occurrence day $1,\ldots,21$. The right-hand panel illustrates the empirical proportions of events that are reported with a reporting delay of $0,\ldots,10$ days.}   
\label{fig:occandrepstructure}
\end{figure}

\begin{figure}[H] 
\centering
\subfloat[$x^{(es)}_{3,1}=1$ and $x^{(es)}_{4,3}=1$ ]{

\begin{adjustbox}{width=0.45\textwidth}
    \includegraphics[width = \textwidth]{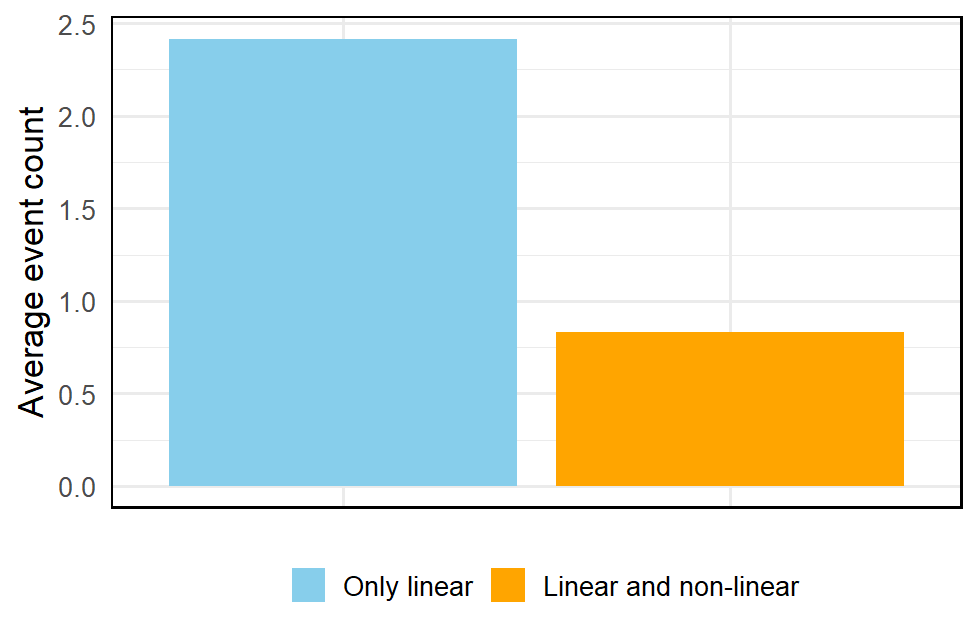}
\end{adjustbox}

}
\hspace{1.5em}
\subfloat[$x^{(es)}_{1,1}=1$ and $x^{(es)}_{4,3}=1$]{

\begin{adjustbox}{width=0.45\textwidth}
\includegraphics[width = \textwidth]{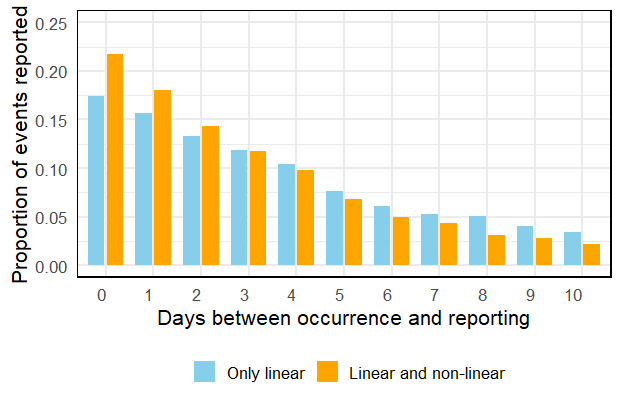}

\end{adjustbox}

}
\caption{Examples of specific occurrence and reporting effects within the simulated data, illustrated for both experiments on $\mathcal{D}_{\text{tune}}$. The left-hand panel shows the average number of event counts $N_i$ for observations with $x^{(es)}_{3,1}=1$ and $x^{(es)}_{4,3}=1$. The right-hand panel shows the empirical proportions of events that are reported with a reporting delay of $0,\ldots,10$ days for observations with $x^{(es)}_{1,1}=1$ and $x^{(es)}_{4,3}=1$. }   
\label{fig:occandrepstructurespecifics}
\end{figure}

\begin{figure}[H] 
\centering
\subfloat[$\text{occ}(i)=10$]{

\begin{adjustbox}{width=0.33\textwidth}
    \includegraphics[width = \textwidth]{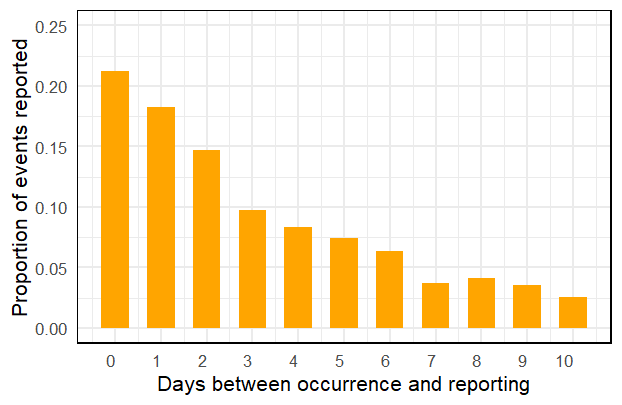}
\end{adjustbox}

}
\subfloat[$\text{occ}(i)=15$]{

\begin{adjustbox}{width=0.33\textwidth}
\includegraphics[width = \textwidth]{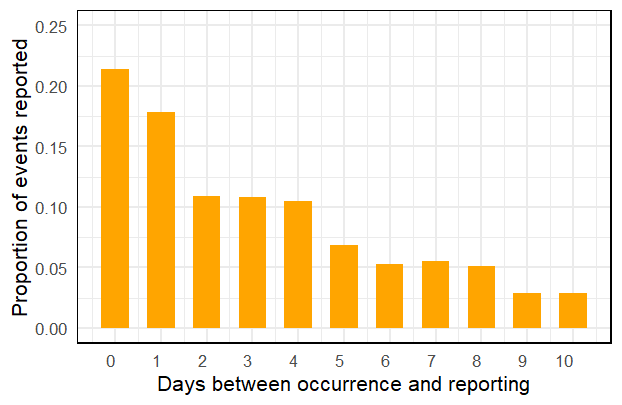}

\end{adjustbox}

} 
\subfloat[$\text{occ}(i)=21$]{

\begin{adjustbox}{width=0.33\textwidth}
\includegraphics[width = \textwidth]{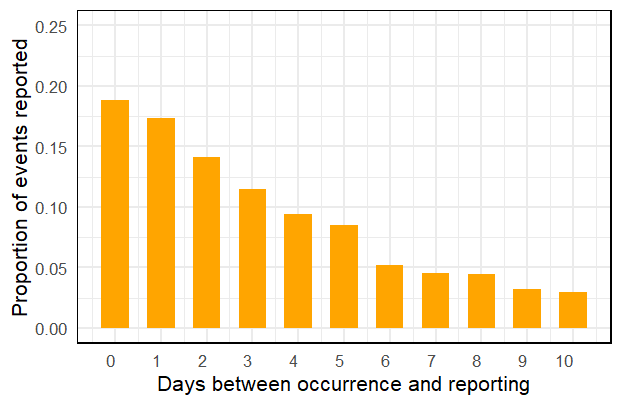}

\end{adjustbox}

}
\caption{Empirical proportions for events that are reported on day $\text{occ}(i)=10$ (left-hand side), day $\text{occ}(i)=15$ (middle) or day $\text{occ}(i)=21$ with a delay of $0,\ldots,10$ days. The proportions are calculated for the setting with linear and non-linear effects on $\mathcal{D}_{\text{tune}}$ for the setting including both linear and non-linear effects.} 
\label{fig:daystructure}
\end{figure}

\paragraph{Occurrence and reporting effects on response in simulated data}
 In Figure~\ref{fig:exploratory}, we illustrate an example of a simulated (non-)linear effect on $\mathcal{D}_{\text{tune}}$ for both experimental settings. We show the average of the event occurrence counts $N_i$ across different values of $x^{(es)}_{2}$, for the subset of observations with $x^{(es)}_{1,1}=0$ (blue bars) and $x^{(es)}_{1,1}=1$ (red bars), respectively. For both experiments, the average number of event occurrences is lower when $x^{(es)}_{1,1}=1$. In the setting with only linear effects (left-hand panel), we observe a linear upward trend in the average number of events for higher values of $x^{(es)}_{2}$. In the non-linear setting (right-hand panel), there is no such trend but the average number of events increases steeply for high values of $x^{(es)}_{2}$. Additionally, when $x^{(es)}_{1,1}=1$ and for low values of $x^{(es)}_{2}$, the values are lower relative to the observations with $x^{(es)}_{1,1}=0$. This is in alignment with the specifications given in Tables~\ref{tab:spec1} and \ref{tab:spec2} in Appendix~\ref{app:figures}. 
Figure~\ref{fig:occandrepstructure} shows the global occurrence (left-hand side) and reporting pattern (right-hand side) in our simulated data. With respect to the event occurrence, we observe that there is a higher occurrence on weekend days (such as day 13 and 14 in the scheme visualised in Figure~\ref{fig:daystructure}) and holidays (such as day 17). Overall, more events occur in the setting with only linear effects compared to the setting with both linear and non-linear effects. The reporting delay structure indicates that events are on average reported with shorter delay in the setting with both linear and non-linear effects.
Figure~\ref{fig:occandrepstructurespecifics} illustrates covariate effects for the occurrence and the reporting process that contribute to the overall pattern shown in Figure~\ref{fig:occandrepstructure}. The left-hand panel shows that the average number of event counts $N_i$ is significantly lower in the setting with both linear and non-linear effects as compared to the linear setting when $x^{(es)}_{3,1}=1$ and $x^{(es)}_{4,3}=1$. For observations with $x^{(es)}_{1,1}=1$ and $x^{(es)}_{4,3}=1$, reporting happens later in the linear setting. These empirical findings are consistent with the specifications for $\boldsymbol{\beta}_\lambda$ and $\boldsymbol{\beta}_{p_{1}},\ldots,\boldsymbol{\beta}_{p_{11}}$ in Tables~\ref{tab:spec1} and \ref{tab:spec2} in Appendix~\ref{app:figures}.

In Figure~\ref{fig:daystructure}, we visualise the simulated reporting structure at a more granular level for the setting including both linear and non-linear effects. We show the empirical reporting proportions for events that occurred on day 10, day 15 or day 21. For each occurrence day, the reporting probability steadily drops over time. We observe that the reporting probability is lower on a weekend day. For instance, when $\text{occ}(i)=10$ (left panel), reporting 3 or 4 days later corresponds to the weekend days 13 and 14 in the scheme of Figure~\ref{fig:daystructure}, respectively. At this delay, we observe an inflection point, evidencing the lower empirical reporting probability of reporting during weekend days. Other examples include a delay of 5 and 6 days in the middle panel and a delay of 6 and 7 days in the right-hand panel. Holidays also result in a lower probability of reporting. When $\text{occ}(i)=10$ (left panel), a delay of 7 days corresponds to day 17 which is a holiday, as shown in Figure~\ref{fig:daystructure}. At this delay, the graph shows a clear drop in the empirical proportion of events reported. Another example is a delay of 2 days for $\text{occ}(i)=15$ (middle panel), which also corresponds to day 17. Again, these empirical findings match the coefficient specifications $\boldsymbol{\beta}_\lambda,\boldsymbol{\beta}_{p_{1}},\ldots,\boldsymbol{\beta}_{p_{11}}$ that are given in Table~\ref{tab:spec2} in Appendix~\ref{app:figures}.

\subsection{Results}
\begin{figure}[H]
\centering

\textbf{(a) Only linear effects} \\[0.5em]

\begin{minipage}{0.48\textwidth}
\centering
\includegraphics[width=\textwidth]{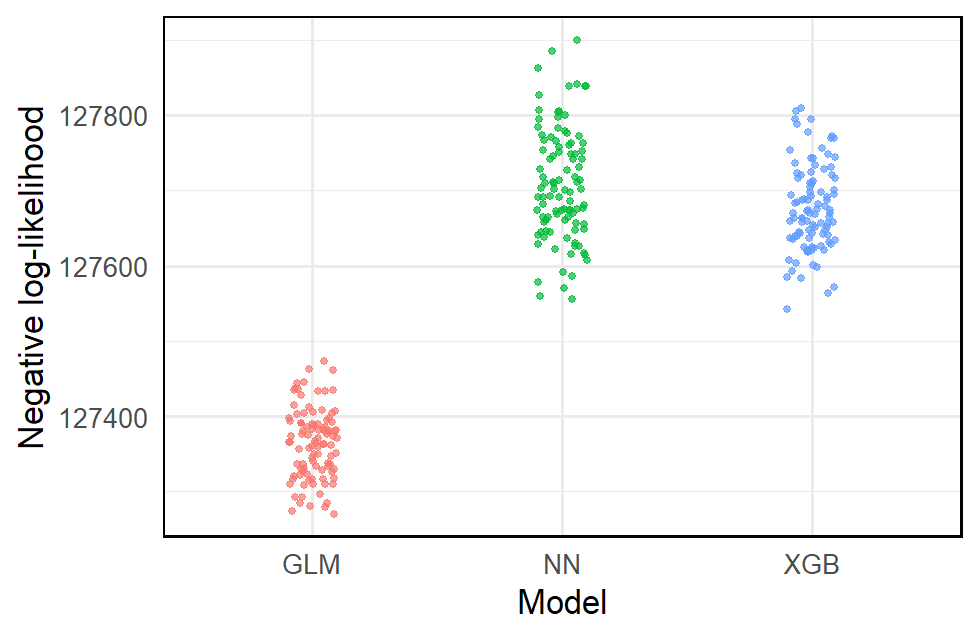}\\
{\footnotesize \textbf{(a.1)} $-LL(\widehat{\lambda}_{i}^{[l]}, \widehat{p}_{i,j}^{[l]};D_{\text{test}})$}
\end{minipage}
\hfill
\begin{minipage}{0.48\textwidth}
\centering
\includegraphics[width=\textwidth]{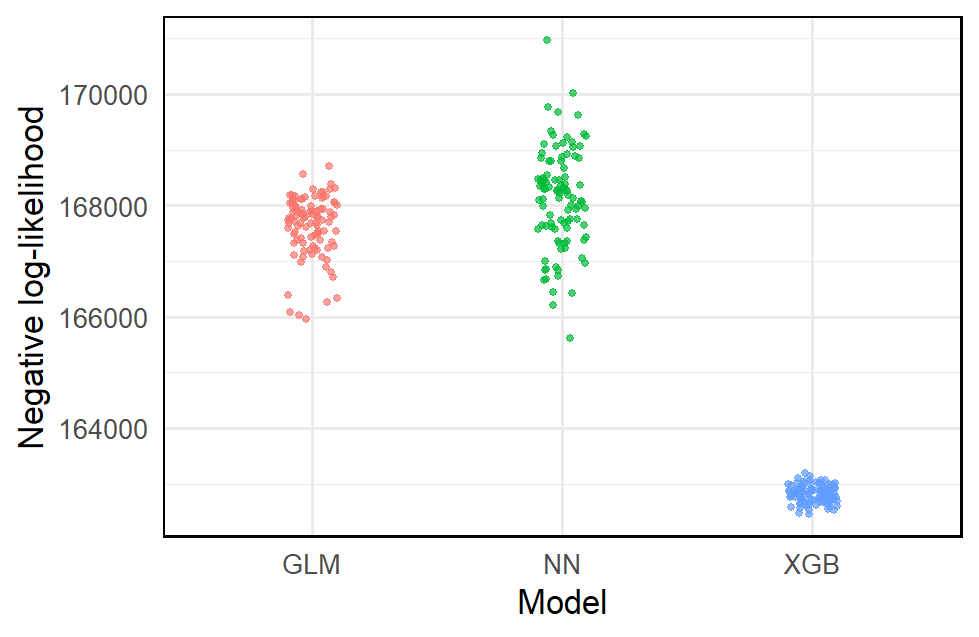}\\
{\footnotesize \textbf{(a.2)} $-LL_{c}(\widehat{\lambda}_{i}^{[l]}, \widehat{p}_{i,j}^{[l]};\mathcal{D}_{\text{test}})$}
\end{minipage}

\vspace{1em}

\textbf{(b) Linear and non-linear effects} \\[0.5em]

\begin{minipage}{0.48\textwidth}
\centering
\includegraphics[width=\textwidth]{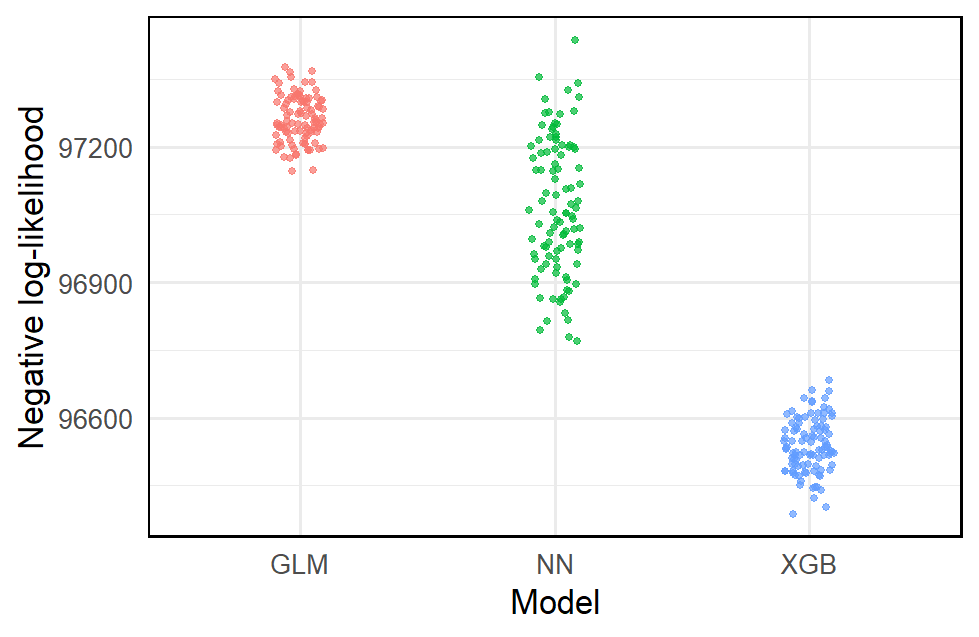}\\
{\footnotesize \textbf{(b.1)} $-LL(\widehat{\lambda}_{i}^{[l]}, \widehat{p}_{i,j}^{[l]};D_{\text{test}})$}
\end{minipage}
\hfill
\begin{minipage}{0.48\textwidth}
\centering
\includegraphics[width=\textwidth]{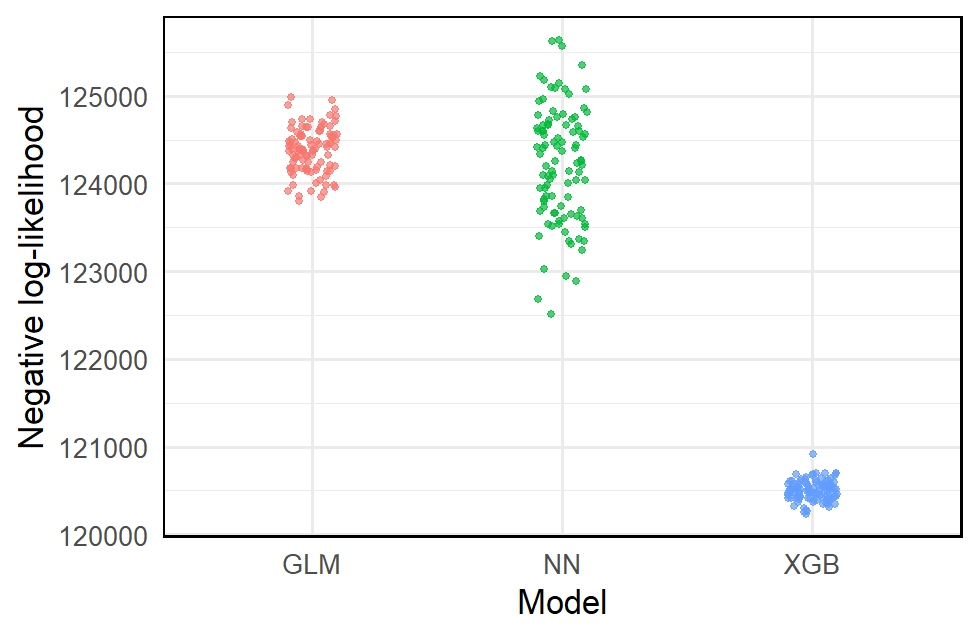}\\
{\footnotesize \textbf{(b.2)} $-LL_{c}(\widehat{\lambda}_{i}^{[l]}, \widehat{p}_{i,j}^{[l]};\mathcal{D}_{\text{test}})$}
\end{minipage}

\caption{Out-of-sample reported (left panels) and complete (right panels) negative log-likelihoods using the estimates of the occurrence intensities and reporting probabilities in $\mathcal{D}_{\text{test}}$ estimated using the occurrence and reporting model trained on $\mathcal{D}_{l}$ with $l=1\ldots,100$, for both experimental settings. Red, green, and blue dots correspond to the values obtained when using GLM, neural network, and XGBoost models in the maximization step, respectively.}
\label{fig:result_completeandreported}
\end{figure}

\paragraph{Out-of-sample performance}To each of the 100 simulated datasets $\mathcal{D}_{1},\ldots,\mathcal{D}_{100}$, for both experimental settings, we apply the GLM-based approach as well as the neural network- and XGBoost-based approaches detailed in Section~\ref{sec:method}. In the occurrence models, all entity-specific covariates as well as covariates related to the occurrence period are included. In the reporting models, all entity-specific and period-related covariates are included. Let $\widehat{\lambda}_{i}^{[l]}$ and $\widehat{p}_{i,j}^{[l]}$ with $l=1,\ldots,100$ denote the estimates for the occurrence intensities and reporting probabilities in the test set $\mathcal{D}_{\text{test}}$, respectively, estimated using the occurrence and reporting model trained on $\mathcal{D}_{l}$. In Figure~\ref{fig:result_completeandreported}, we visualise the overall out-of-sample performance for the trained GLMs, neural networks and XGBoost models. In the left panels, we show the out-of-sample observed negative log-likelihood $-LL(\widehat{\lambda}_{i}^{[l]}, \widehat{p}_{i,j}^{[l]};D_{\text{test}})$, which excludes event counts reported after $\tau=21$. The observed log-likelihood indicates a better performance when using GLMs in the presence of only linear effects. On the other hand, when both linear and non-linear effects are present, the use of XGBoost in the maximisation step is the most performant in terms of the observed likelihood. However, since the data are simulated, we know the complete data $\mathcal{D}_{\text{test}}$ and we can use this to compute the negative complete log-likelihood $-LL_{c}(\widehat{\lambda}_{i}^{[l]}, \widehat{p}_{i,j}^{[l]};\mathcal{D}_{\text{test}})$, as illustrated in the right-hand side panels in Figure~\ref{fig:result_completeandreported}. The resulting values indicate, in contrast to the observed log-likelihood, that XGBoost performs better in both settings, i.e.~in the presence of only linear effects as well as both linear and non-linear effects. This suggests that XGBoost generalises better to the unseen part of the data and more effectively captures the interdependencies between the occurrence and reporting model.

\begin{figure}[H]
\centering

\textbf{(a) Only linear effects} \\[0.5em]

\begin{minipage}{0.48\textwidth}
\centering
\includegraphics[width=\textwidth]{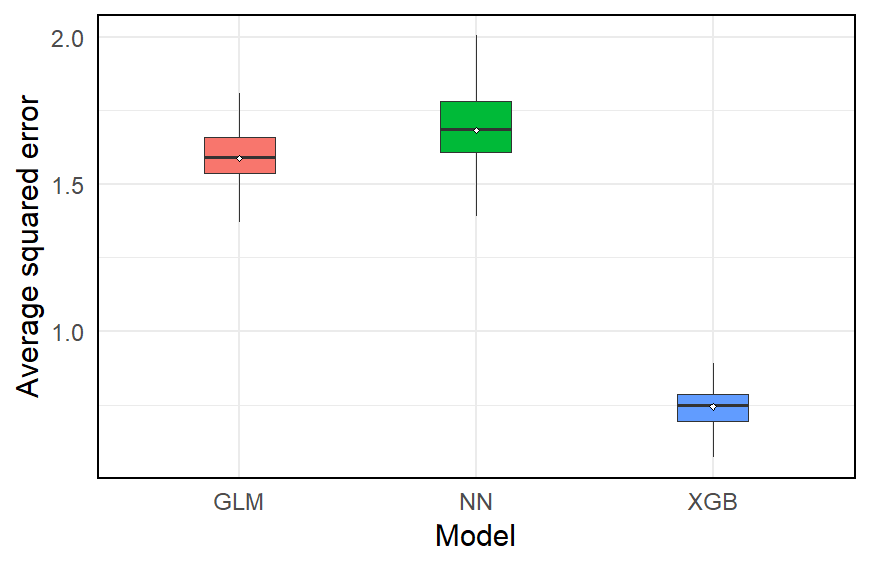}\\
{\footnotesize \textbf{(a.1)} $\text{ASE}(\widehat{\lambda}^{[l]};D_{\text{test}})$}
\end{minipage}
\hfill
\begin{minipage}{0.48\textwidth}
\centering
\includegraphics[width=\textwidth]{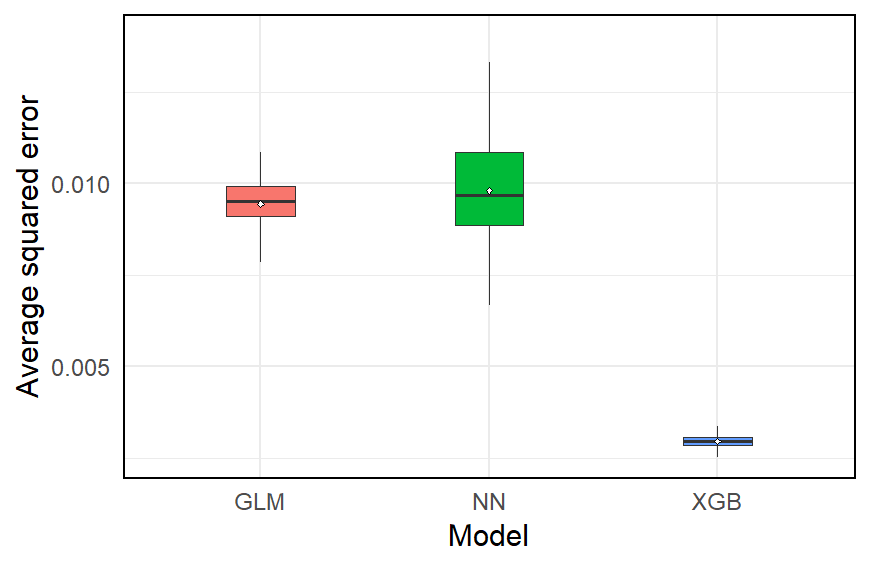}\\
{\footnotesize \textbf{(a.2)} $\text{ASE}(\widehat{p}^{[l]};D_{\text{test}})$}
\end{minipage}

\vspace{1em}

\textbf{(b) Linear and non-linear effects} \\[0.5em]

\begin{minipage}{0.48\textwidth}
\centering
\includegraphics[width=\textwidth]{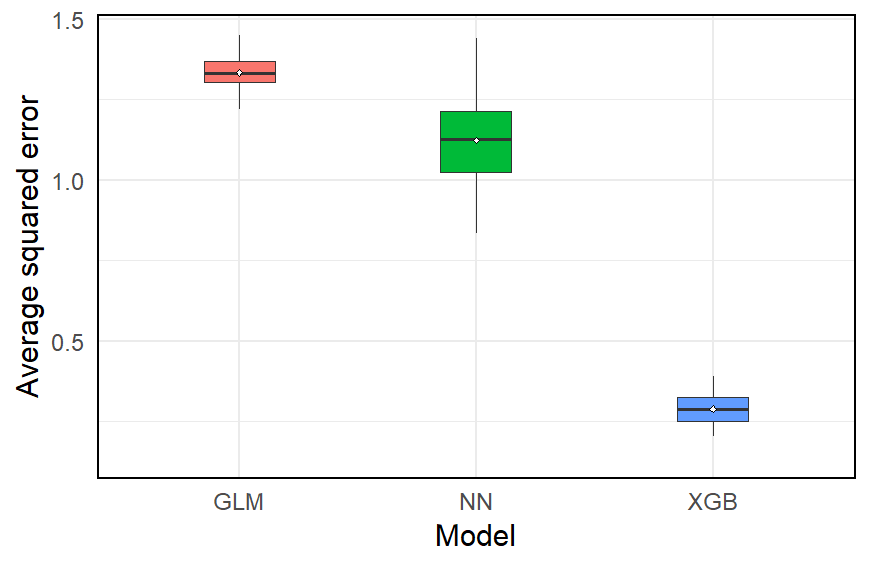}\\
{\footnotesize \textbf{(b.1)} $\text{ASE}(\widehat{\lambda}^{[l]};D_{\text{test}})$}
\end{minipage}
\hfill
\begin{minipage}{0.48\textwidth}
\centering
\includegraphics[width=\textwidth]{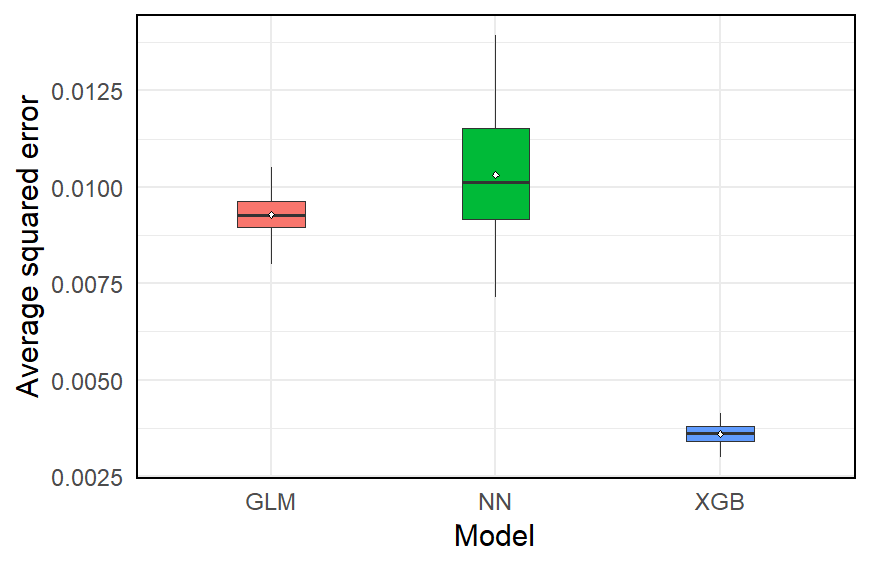}\\
{\footnotesize \textbf{(b.2)} $\text{ASE}(\widehat{p}^{[l]};D_{\text{test}})$}
\end{minipage}

\caption{Boxplots of the out-of-sample average squared errors $\text{ASE}(\widehat{\lambda}^{[l]};D_{\text{test}})$ (left panels) and $\text{ASE}(\widehat{p}^{[l]};D_{\text{test}})$  (right panels), for both experimental settings. The red, green, and blue boxplots correspond to GLM, neural network, and XGBoost models used in the maximization step, respectively.}
\label{fig:result_ASE}
\end{figure}

To better understand the underlying dynamics, we can separately examine the occurrence and reporting models to assess whether their individual performances differ from the overall performance sketched in Figure~\ref{fig:result_completeandreported}. Since we simulate data, we know the true underlying occurrence intensities $\lambda_i$ and reporting probabilities $p_{i,j}$. Given estimates $\widehat{\lambda}_i$ and $\widehat{p}_{i,j}$ as well as the true parameters $\lambda_i$ and $p_{i,j}$ for a dataset $D$ with $n$ observations, we define the average of the squared errors
\begin{align*}
    \text{ASE}(\widehat{\lambda};D)= \frac{1}{n}\sum_{i=1}^{n} (\lambda_i - \widehat{\lambda}_i)^{2} \:\:\: \text{and} \:\:\: \text{ASE}(\widehat{p};D)= \frac{1}{n}\sum_{i=1}^{n} \sum_{j=1}^{d} (p_{i,j} - \widehat{p}_{i,j})^{2},
\end{align*}
which determine how well we retrieve the true occurrence intensities and reporting probabilities, respectively. Figure~\ref{fig:result_ASE} shows boxplots of $\text{ASE}(\widehat{\lambda}^{[l]};D_{\text{test}})$ and $\text{ASE}(\widehat{p}^{[l]};D_{\text{test}})$ for the estimates of the occurrence intensities and reporting probabilities in $D_{\text{test}}$ for the models trained on $D_{l}$ with $l=1,\ldots,100$, for both experimental settings. These results corroborate the findings shown by the complete log-likelihood in Figure~\ref{fig:result_completeandreported}, i.e.~XGBoost results in an improved fit in both the linear-only and combined linear and non-linear scenarios. Both in the occurrence as well as the reporting model, XGBoost yields occurrence intensities and reporting probabilities that are closer to the true values while also having less variability within the results.

\begin{figure}[H]
\centering

\textbf{(a) Occurrence structure} \\[0.5em]

\begin{minipage}{0.32\textwidth}
\centering
\includegraphics[width=\textwidth]{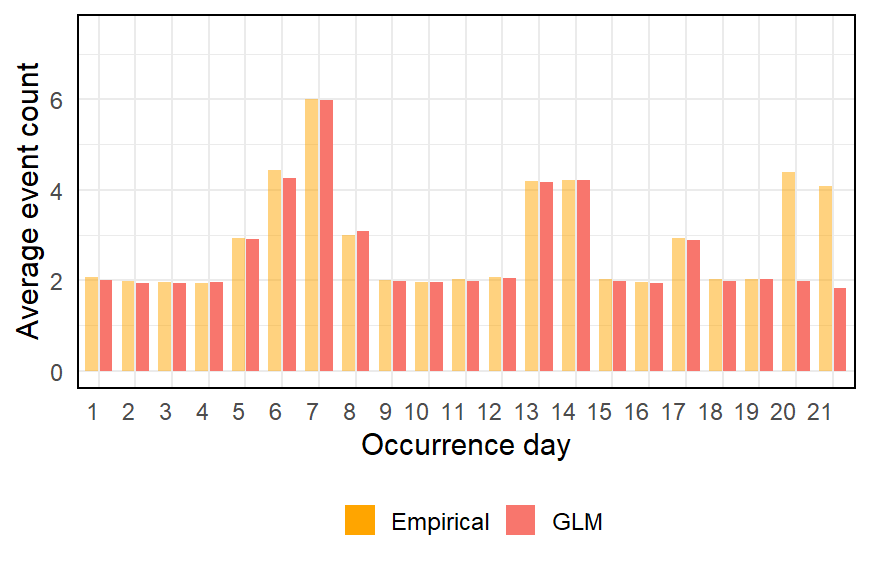}\\
{\footnotesize \textbf{(a.1)} GLM}
\end{minipage}
\hfill
\begin{minipage}{0.32\textwidth}
\centering
\includegraphics[width=\textwidth]{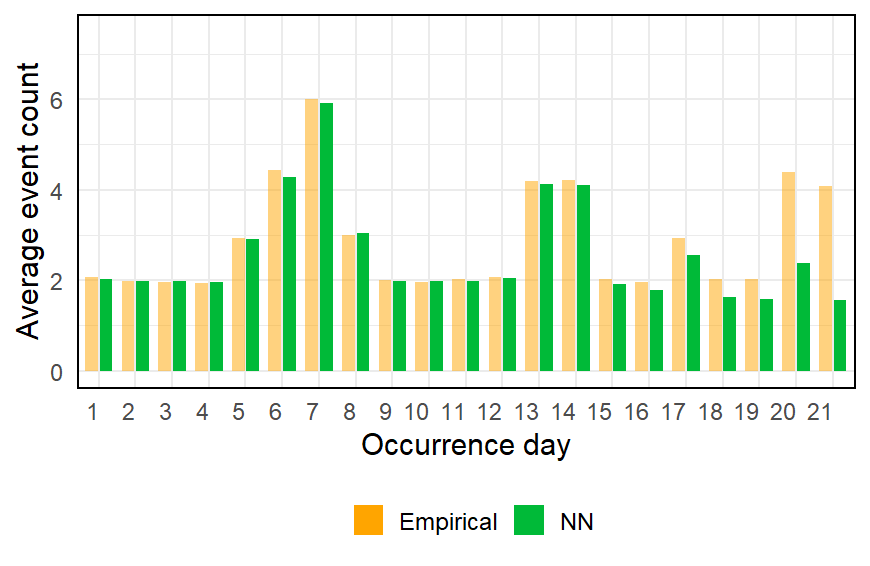}\\
{\footnotesize \textbf{(a.2)} Neural network}
\end{minipage}
\hfill
\begin{minipage}{0.32\textwidth}
\centering
\includegraphics[width=\textwidth]{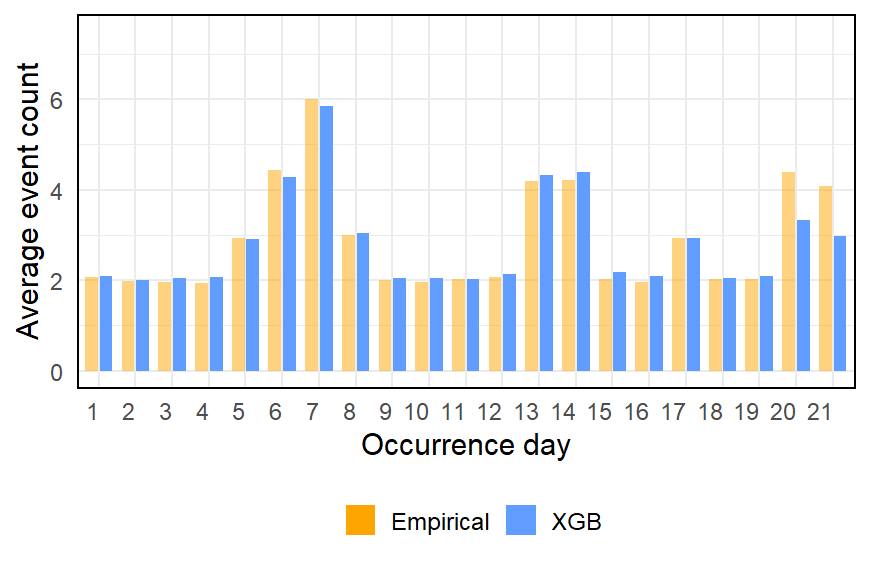}\\
{\footnotesize \textbf{(a.3)} XGBoost}
\end{minipage}

\vspace{1em}

\textbf{(b) Reporting structure} \\[0.5em]

\begin{minipage}{0.32\textwidth}
\centering
\includegraphics[width=\textwidth]{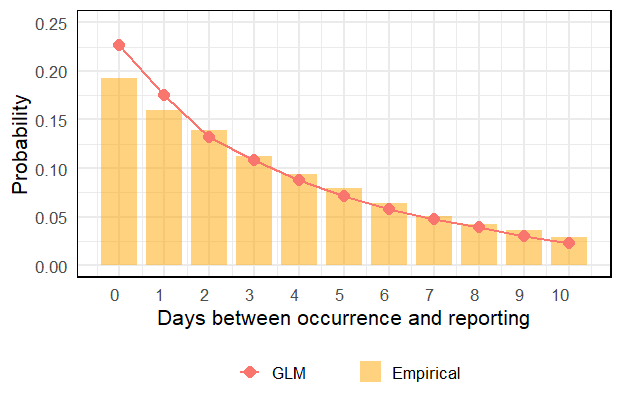}\\
{\footnotesize \textbf{(b.1)} GLM}
\end{minipage}
\hfill
\begin{minipage}{0.32\textwidth}
\centering
\includegraphics[width=\textwidth]{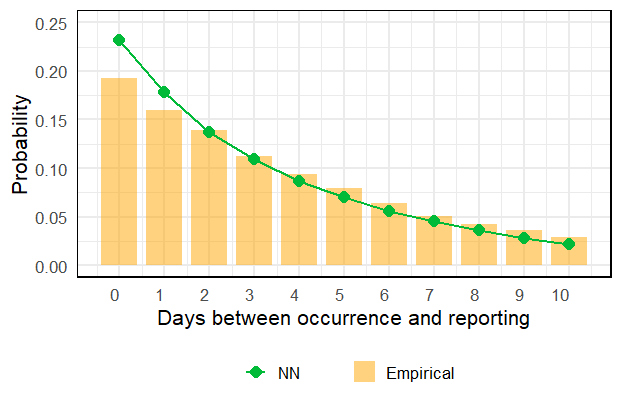}\\
{\footnotesize \textbf{(b.2)} Neural network}
\end{minipage}
\hfill
\begin{minipage}{0.32\textwidth}
\centering
\includegraphics[width=\textwidth]{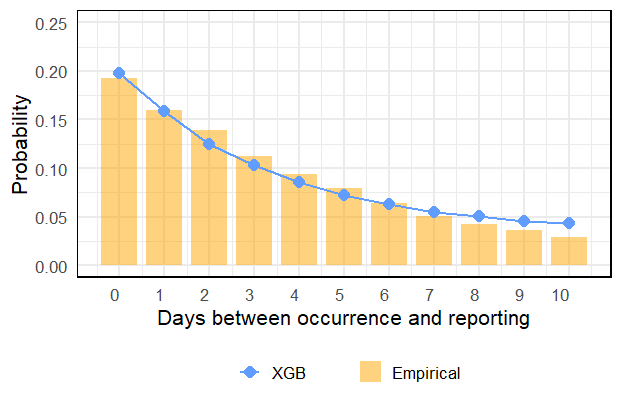}\\
{\footnotesize \textbf{(b.3)} XGBoost}
\end{minipage}

\caption{Visualisation of the overall out-of-sample occurrence and reporting fit for the setting with both linear and non-linear effects. The top three panels show the average occurred number of events per occurrence day in $\mathcal{D}_{\text{test}}$: the mean over the  estimates of the occurrence models trained on $\mathcal{D}_{1},\ldots,\mathcal{D}_{100}$ as well as the empirical quantities. The bottom three panels show the average probability of an event being reported $0,\ldots,10$ days later in $\mathcal{D}_{\text{test}}$: the mean over the estimates of the reporting models trained on $\mathcal{D}_{1},\ldots,\mathcal{D}_{100}$ as well as the empirical quantities. Orange is used to indicate the empirical quantities, while red, green and blue are used to denote the values for the GLMs, neural networks and XGBoost models, respectively.}
\label{fig:result_exp2_occandrepstructure}
\end{figure}

\begin{figure}[H]
\centering

\textbf{(a) Reporting structure for $\text{occ}(i)=10$} \\[0.5em]

\begin{minipage}{0.32\textwidth}
\centering
\includegraphics[width=\textwidth]{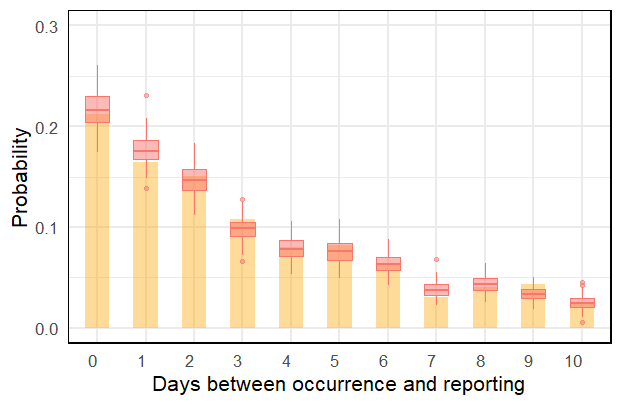}\\
{\footnotesize \textbf{(a.1)} GLM}
\end{minipage}
\hfill
\begin{minipage}{0.32\textwidth}
\centering
\includegraphics[width=\textwidth]{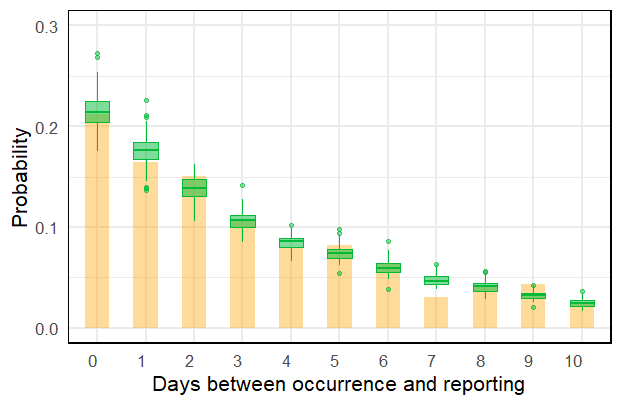}\\
{\footnotesize \textbf{(a.2)} Neural network}
\end{minipage}
\hfill
\begin{minipage}{0.32\textwidth}
\centering
\includegraphics[width=\textwidth]{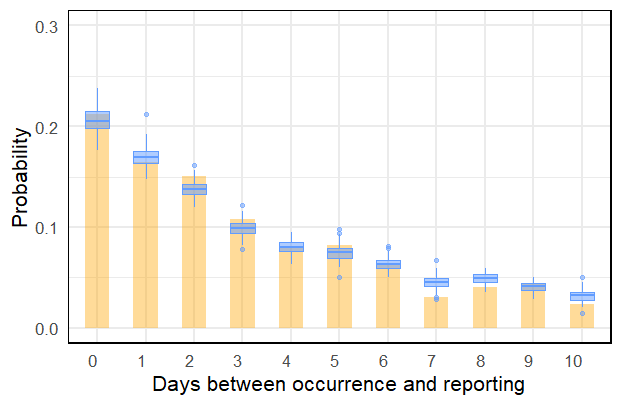}\\
{\footnotesize \textbf{(a.3)} XGBoost}
\end{minipage}

\vspace{1em}

\textbf{(b) Reporting structure for $\text{occ}(i)=17$} \\[0.5em]

\begin{minipage}{0.32\textwidth}
\centering
\includegraphics[width=\textwidth]{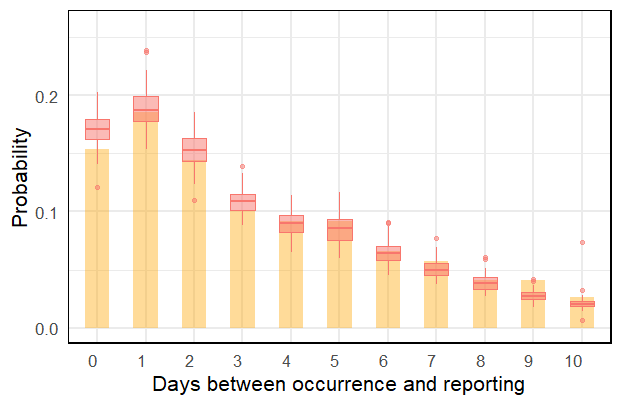}\\
{\footnotesize \textbf{(b.1)} GLM}
\end{minipage}
\hfill
\begin{minipage}{0.32\textwidth}
\centering
\includegraphics[width=\textwidth]{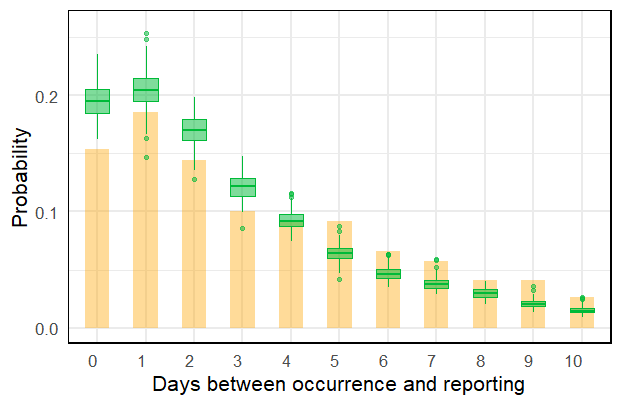}\\
{\footnotesize \textbf{(b.2)} Neural network}
\end{minipage}
\hfill
\begin{minipage}{0.32\textwidth}
\centering
\includegraphics[width=\textwidth]{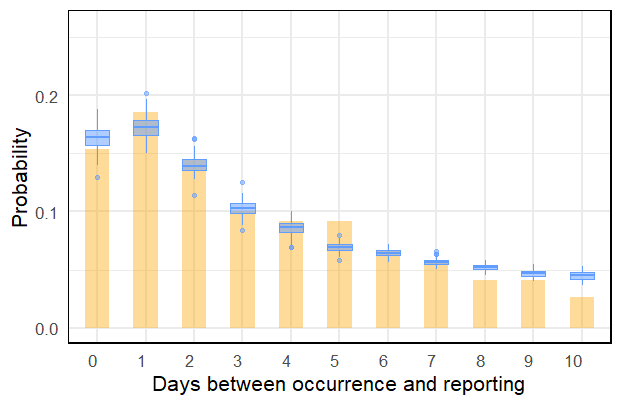}\\
{\footnotesize \textbf{(b.3)} XGBoost}
\end{minipage}

\vspace{1em}

\textbf{(c) Reporting structure for $\text{occ}(i)=21$} \\[0.5em]

\begin{minipage}{0.32\textwidth}
\centering
\includegraphics[width=\textwidth]{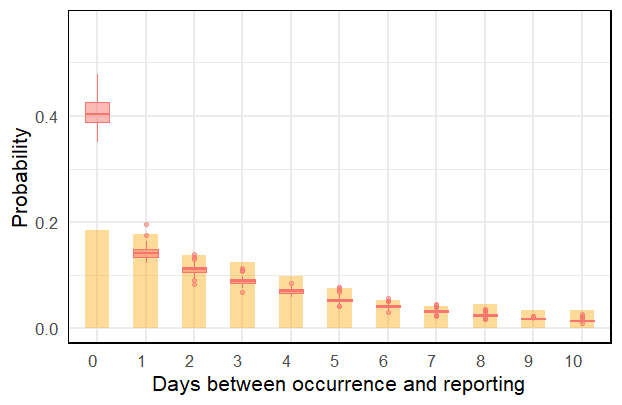}\\
{\footnotesize \textbf{(c.1)} GLM}
\end{minipage}
\hfill
\begin{minipage}{0.32\textwidth}
\centering
\includegraphics[width=\textwidth]{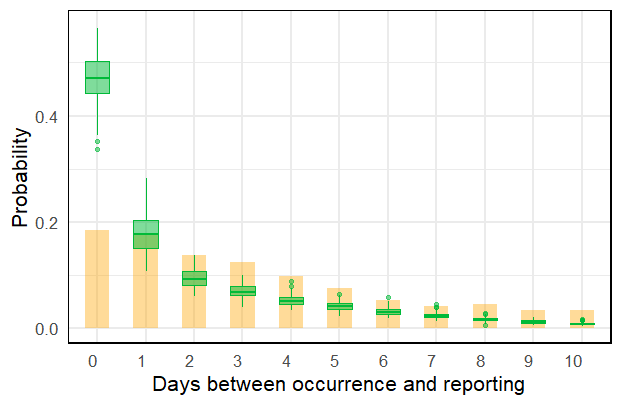}\\
{\footnotesize \textbf{(c.2)} Neural network}
\end{minipage}
\hfill
\begin{minipage}{0.32\textwidth}
\centering
\includegraphics[width=\textwidth]{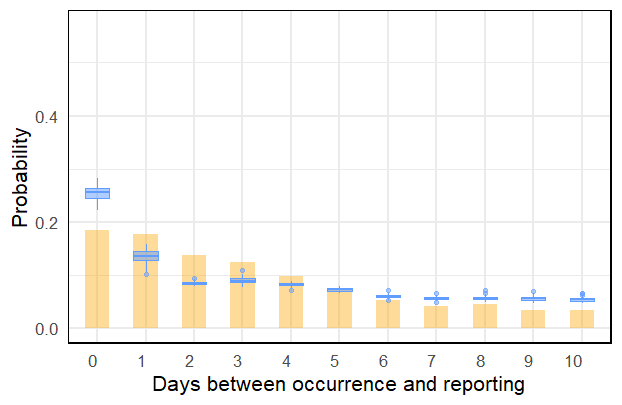}\\
{\footnotesize \textbf{(c.3)} XGBoost}
\end{minipage}

\caption{Out-of-sample reporting fit for $\text{occ}(i)=10$ (top panels), $\text{occ}(i)=17$ (middle panels) and $\text{occ}(i)=21$ (bottom panels) in the setting with both linear and non-linear effects. The orange line indicates empirical reporting probabilities in the test data, while the red, green, and blue boxplots correspond to estimated reporting probabilities in $\mathcal{D}_{\text{test}}$ from the GLMs, neural networks, and XGBoost models, averaged over the estimates of the reporting models trained on $\mathcal{D}_{1},\ldots,\mathcal{D}_{100}$, respectively.}
\label{fig:result_exp2_repstructure_day10and17}
\end{figure}

\paragraph{Occurrence and reporting structure}For the remainder of this section, we will focus on the setting with both linear and non-linear effects to explore differences between the GLMs, neural networks and XGBoost models. In Figure~\ref{fig:result_exp2_occandrepstructure}, we compare the empirical occurrence and reporting structure present in $\mathcal{D}_{\text{test}}$ with the predicted values (averaged over the 100 trained occurrence and reporting models computed on $\mathcal{D}_{l}$ with $l=1,\ldots,100$, respectively). We observe that the empirical trend for the event counts $N_i$ across different occurrence dates is generally well-captured by the predicted occurrence intensities. However, for $\text{occ}(i)=20$ and $\text{occ}(i)=21$, the results are worse with only the XGBoost model partially capturing the trend. The initial estimates of the event counts $N^{(0)}_{i,j}$ are an underestimation, since the data are only partially observed. The closer to the edge of the observation window, i.e.~$\tau=21$, the more the event counts will initially be underestimated. Across iterations, the EM algorithm adjusts the dataset to account for this underestimation as the trend within the data is learned. In our generated time window, shown in Figure~\ref{fig:daystructure}, we have $x_{3}^{(ps)}=1$ for the last two observable days, i.e.~day 20 and 21. The models mistakenly explain the lower event counts for those two days as resulting from an effect of $x_{3}^{(ps)}$, while this variable is redundant for explaining the occurrence intensities as shown in Table~\ref{tab:spec2} in Appendix~\ref{app:figures}. As a result, the EM algorithm will not properly adjust the data in the expectation step. However, the XGBoost approach still partially captures the effect, which is not the case for the GLM and neural network. Our results show that of the three considered models, the XGBoost approach best captures the occurrence structure. For the reporting structure, our results indicate that the XGBoost approach best captures the overall reporting probabilities, specifically for short delays, while providing a slightly worse fit for the tail probabilities compared to the other models. Both the GLMs and neural networks predict too high reporting probabilities when the delay is small.
In Figure~\ref{fig:result_exp2_repstructure_day10and17}, we zoom in on the (predicted) reporting structure for three occurrence dates increasingly close to the edge of the observation window, i.e.~$\text{occ}(i)=10$, $\text{occ}(i)=17$ and $\text{occ}(i)=21$. We show that the reporting structure is captured at the level of these specific occurrence days. Overall, the estimated reporting probabilities are more precise for the XGBoost models. As time moves closer to the edge of the observation window $\tau=21$, the accuracy of the estimated probabilities slightly decreases. Specifically, for days close to the present time $\tau$, when most event counts are nowcasted, the models tend to overestimate the probabilities associated with a short delay in reporting more and the tail fit deteriorates with an overestimation for XGBoost and an underestimation for GLM and neural net. The former effect is more noticeable for the GLMs and the neural networks as time moves compared to the XGBoost, as indicated by Figure~\ref{fig:result_exp2_occandrepstructure}.
\paragraph{Non-linear effects} A tool we can leverage to inspect feature effects are SHapley Additive exPlanations (SHAP) \citep{lundberg2017unified}. SHAP values quantify the marginal contribution of a variable to the prediction by averaging over all possible subsets of variables, at the observational level. In Figure~\ref{fig:result_shapdependency}, we use SHAP values to illustrate the marginal contribution of $x_{2}^{(es)}$ in the occurrence model, for different values of $x_{1,1}^{(es)}$. Each dot represents the SHAP value corresponding to a single observation. Since SHAP values are computationally expensive, we show the SHAP values calculated on $D_{\text{tune}}$ in the setting with both linear and non-linear effects. When comparing with the right-hand panel in Figure~\ref{fig:exploratory}, we observe that the GLM fails to capture the dependency effect without explicitly knowing the functional form, in contrast to the neural network and XGBoost. The neural network results in a smoother effect compared to XGBoost, reflecting the hard splits made in regression trees. The neural network fails to capture the dependency effect for the lower values of $x_{2}^{(es)}$, whereas XGBoost picks up the effect at both the higher and lower values. 

\begin{figure}[H] 
\centering
\subfloat[GLM]{

\begin{adjustbox}{width=0.33\textwidth}
    \includegraphics[width = \textwidth]{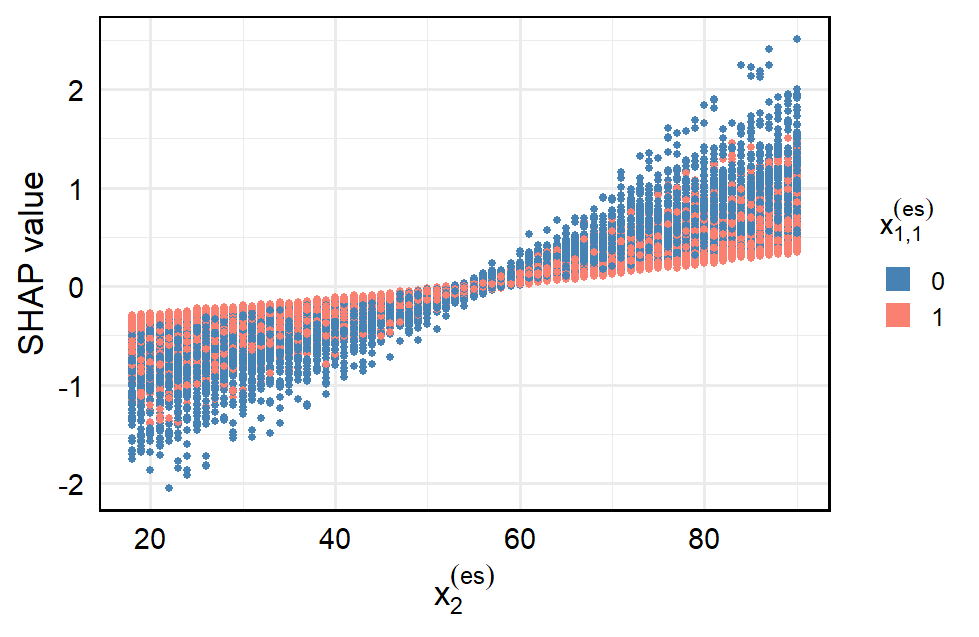}
\end{adjustbox}

}
\subfloat[Neural network]{

\begin{adjustbox}{width=0.33\textwidth}
\includegraphics[width = \textwidth]{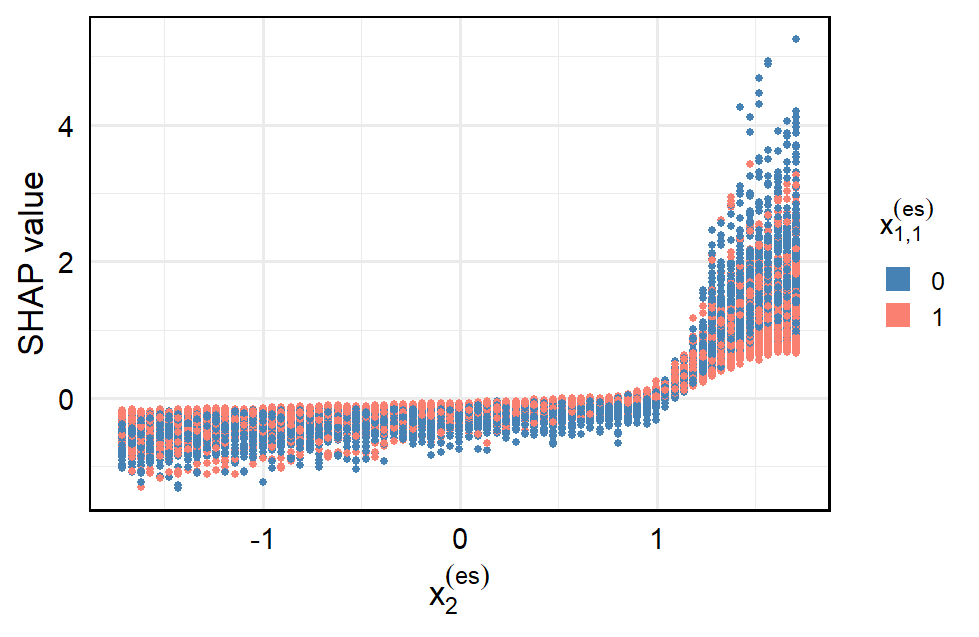}

\end{adjustbox}

} 
\subfloat[XGBoost]{

\begin{adjustbox}{width=0.33\textwidth}
\includegraphics[width = \textwidth]{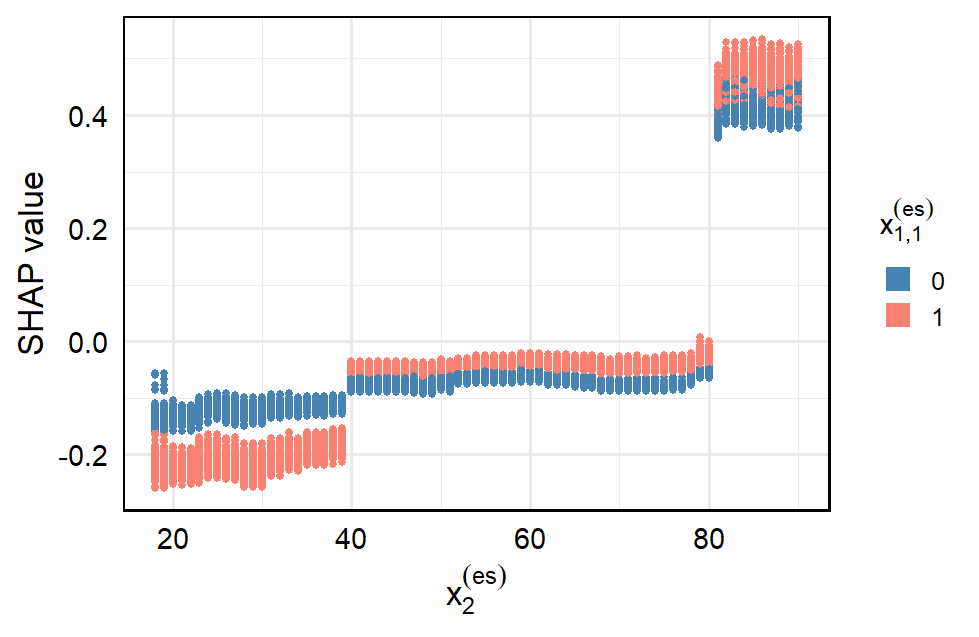}

\end{adjustbox}

}
\caption{Example of SHAP dependency plots of $x_{1,1}^{(es)}$ and $x_{2}^{(es)}$ computed on $D_{\text{tune}}$ for the occurrence model in the setting with both linear and non-linear effects. The three panels show results using GLM (left), neural network
(middle) and XGboost (right), respectively. Blue and red dots denote the SHAP values for observations with $x_{1,1}^{(es)}=0$ and $x_{1,1}^{(es)}=1$, respectively.}
\label{fig:result_shapdependency}
\end{figure}
\paragraph{Comparison with vanilla approach}

\begin{figure}[H] 
\centering
\subfloat[Occurrence model]{

\begin{adjustbox}{width=0.45\textwidth}
    \includegraphics[width = \textwidth]{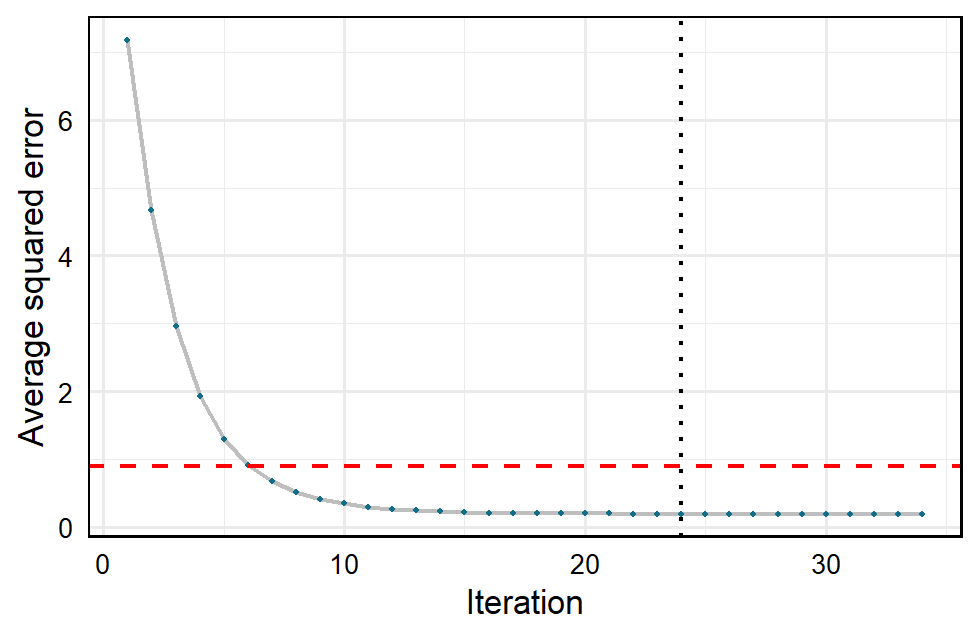}
\end{adjustbox}

}
\hspace{1.5em}
\subfloat[Reporting model]{

\begin{adjustbox}{width=0.45\textwidth}
\includegraphics[width = \textwidth]{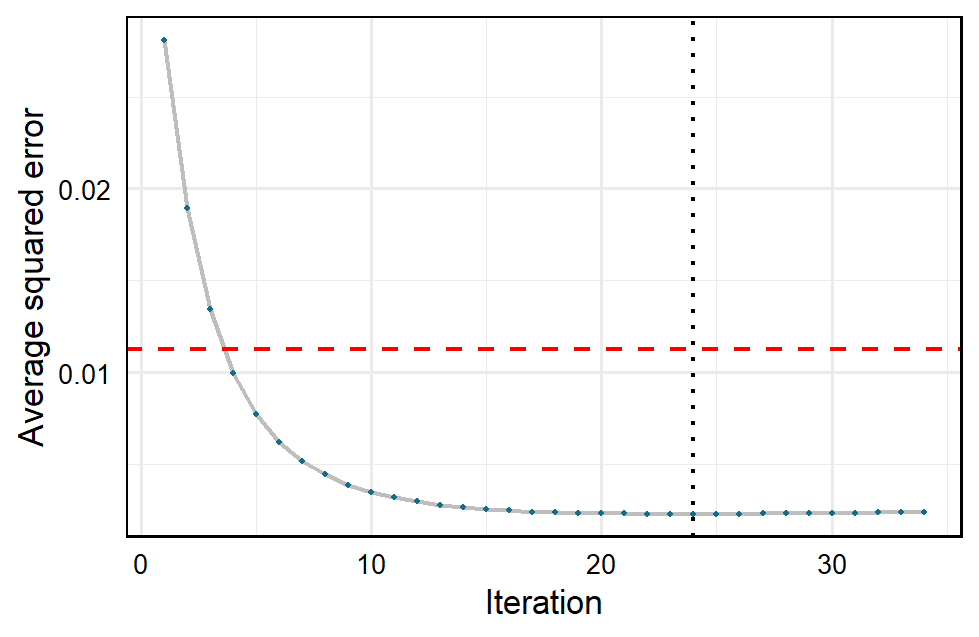}

\end{adjustbox}

}
\caption{Example of the EM algorithm's training process for the experiment including both linear as well as non-linear effects with XGBoost. The left- and the right-hand side show $\text{ASE}(\widehat{\lambda}; D_{\text{tune}})$ and $\text{ASE}(\widehat{p}_{i,j};D_{\text{tune}})$, respectively. The black vertical dotted line indicates the optimal iteration (with \texttt{EM\_patience}=10) for the EM scheme with an additive XGBoost approach. The red horizontal line shows the average squared error reached for the single-iteration vanilla approach.} 
\label{fig:result_xgb}
\end{figure}

We have shown that the EM scheme with an additive XGBoost approach is performant compared to alternative approaches. However, we also want to assess if it is an improvement over a vanilla approach where we fit a single (large) occurrence and reporting model. The additive approach incrementally adapts the model to the changing input data, while the vanilla approach uses the initial expectation of the complete data $\mathcal{D}^{(1)}$ as (static) input data and implements a single maximisation step. In Figure~\ref{fig:result_xgb}, we illustrate this on $D_{\text{tune}}$ for the experiment including both linear and non-linear effects. Specifically, for the additive approach, we learn the occurrence and reporting model as detailed in Section~\ref{sec:method}. Next, we apply the vanilla approach, with the same number of regression trees for the occurrence and reporting model as in the additive case. As a result, in both cases, the final occurrence and reporting model consist of the same number of regression trees. We observe that after only a few iterations, the EM algorithm outperforms the vanilla case, as evidenced by lower values for $\text{ASE}(\widehat{\lambda};D_{\text{tune}})$ and $\text{ASE}(\widehat{p};D_{\text{tune}})$.

%

\section{Reporting of Argentinian Covid-19 cases} \label{sec:covid}

We apply our methodology to a dataset containing real-time case information about the Covid-19 pandemic in Argentina with symptom onset between 1 May 2020 and 15 January 2022. The dataset was compiled using data from \href{https://global.health/}{\texttt{global.health}}, which was originally collected using the methodology of \cite{xu2020Epidemiological}. For each case, both the date of symptom onset and the date of official reporting are recorded. Additionally, we also know if diagnosis happened through a laboratory test or a clinical diagnosis. We only consider cases with a maximum delay of 21 days ($d=22$), which cover approximately 98\% of the total number of cases. The data contains demographic information (age, gender, location) at the individual level. Additionally, we augment the data with day-specific information (for example whether a day is a national holiday in Argentina) for all days within the 21-day reporting window (including the day of symptom onset). We split the dataset into two parts. The first part, which we from now on refer to as $\mathcal{D}$, contains 133\ 157 cases with symptom onset between 1 May 2020 and 1 October 2021 ($\tau$) and is used for model training as well as model validation, following Section~\ref{sec:method}. The second part, which we from now on refer to as $\mathcal{D}_{\text{test}}$, contains 171\ 384 cases with symptom onset between 2 October 2021 and 15 January 2022 and is used for out-of-time evaluation. 

\begin{figure}[H] 
\centering
\subfloat[Occurrence structure]{

\begin{adjustbox}{width=0.45\textwidth}
    \includegraphics[width = \textwidth]{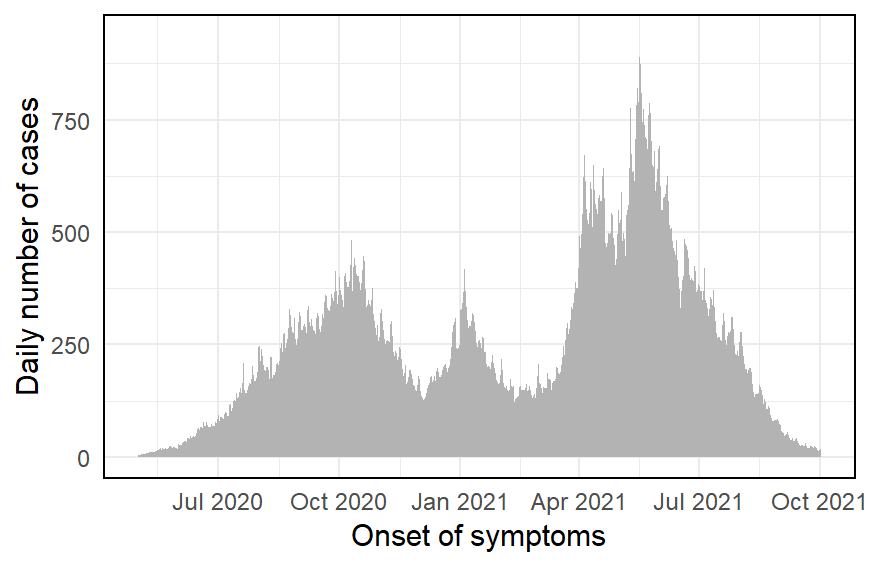}
\end{adjustbox}

}
\hspace{1.5em}
\subfloat[Reporting structure]{

\begin{adjustbox}{width=0.45\textwidth}
\includegraphics[width = \textwidth]{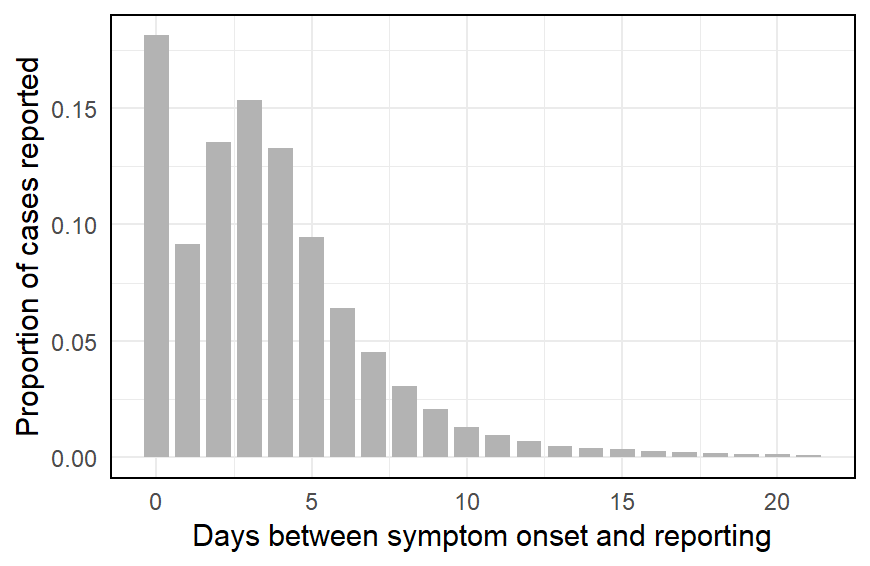}

\end{adjustbox}

}
\caption{Occurrence and reporting pattern for the (training and validation) dataset containing real-time case information about the Covid-19 pandemic in Argentina between 1 May 2020 and 1 October 2021. The left-hand panel shows the daily number of cases $N_i$ for whom there was symptom onset on that day. The right-hand panel illustrates the empirical proportions of the number of cases that are reported with a reporting delay of $0,\ldots,21$ days.}   
\label{fig:covidempir}
\end{figure}

Figure~\ref{fig:covidempir} shows the overall occurrence- and reporting structure for the dataset between 1 May 2020 and 1 October 2021. The distinct epidemiological waves are clearly observable over time. The reporting structure shows two modes, i.e. at a delay of zero and three days. The first is due to reporting being very fast when an individual is clinically diagnosed while reporting is slower when diagnosis happens via a laboratory test. Unfortunately, the dataset does not contain information about non-positive cases, hindering occurrence analysis at the individual level. Additionally, capturing future Covid-19 waves is not possible as one would require data beyond the scope of the available variables such as compliance to e.g.~regulations, air traffic,\ldots. Therefore, our analysis in this section will focus on the reporting and nowcasting of Covid-19 cases at the individual and regional level, leveraging the demographic as well as day-specific information.

We apply our methodology at a granular level, i.e. the individual case level with daily reporting, and examine the previously discussed settings (GLM, neural network and XGBoost). Analogous to Section~\ref{sec:simul}, we perform a random grid search to tune the model-specific parameters for both the neural network- and XGBoost-based approaches. Table~\ref{tab:covidtuning} in Appendix~\ref{app:covid} shows the selected model settings for both approaches. Table~\ref{tab:covidlikelihood} shows the out-of-sample negative complete log-likelihood for each approach, computed on $\mathcal{D}_{\text{test}}$ (case information between 2 October 2021 and 15 January 2022) using the estimates from the occurrence and reporting model trained on $\mathcal{D}$ (case information between 1 May 2020 and 1 October 2021). To learn the occurrence model, demographic information (such as age of the infected person) is used as well as day-specific information (such as an indicator for weekend days) related to the day of symptom onset. Note that all estimated occurrence intensities are approximately one since we only have information related to positive cases. For the reporting model, demographic information is used in addition to day-specific information for each day within the 21-day reporting period given the date of symptom onset. For the latter, this corresponds to 1149 variables after one-hot encoding. Our (out-of-sample) results indicate that the XGBoost-based approach outperforms the GLM-based and neural network-based approaches.
\begin{table}[H]
\centering
\begin{tabular}{m{9em}m{8.5em}}
  \hline\\[-1.1em]
 &$-LL_{c}(\widehat{\lambda}_i, \widehat{p}_{i,j};\mathcal{D}_{\text{test}})$  \\
 \hline\\[-1.1em]
 \textbf{Model}&\\
 \quad GLM&\qquad 566\ 421.4\\
 \quad Neural network&\qquad 585\ 407.0 \\
 \quad XGBoost&\qquad \textbf{524\ 862.9}\\

\hline 
 
\end{tabular}
\caption{Out-of-sample complete negative log-likelihood $-LL_{c}(\widehat{\lambda}_i, \widehat{p}_{i,j};\mathcal{D}_{\text{test}})$ for the GLM-, neural network- and XGBoost-based EM approaches for the application to Argentinian Covid-19 data.}
\label{tab:covidlikelihood}
\end{table}

\begin{figure}[H]
\centering

\textbf{(a) In-sample} \\[0.5em]

\begin{minipage}{0.32\textwidth}
\centering
\includegraphics[width=\textwidth]{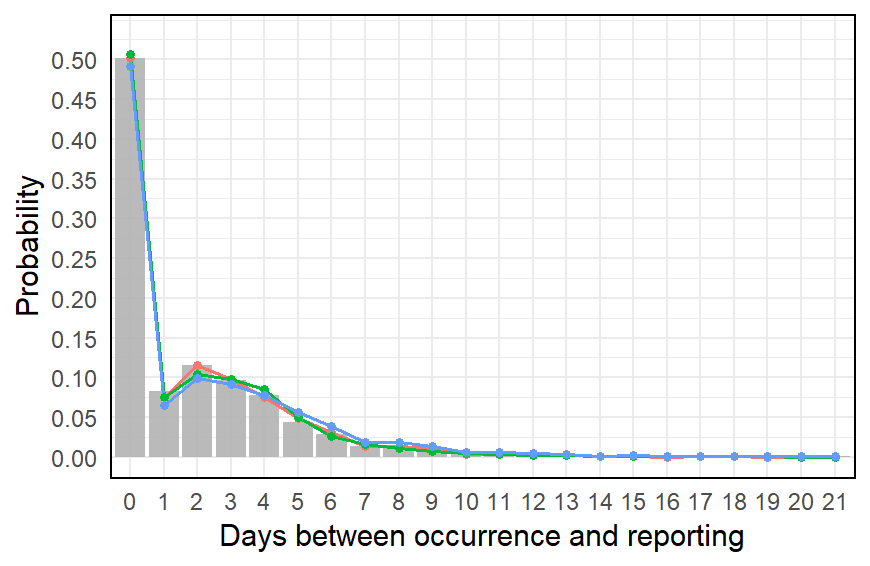}\\
{\footnotesize \textbf{(a.1)} Age 1-5}
\end{minipage}
\hfill
\begin{minipage}{0.32\textwidth}
\centering
\includegraphics[width=\textwidth]{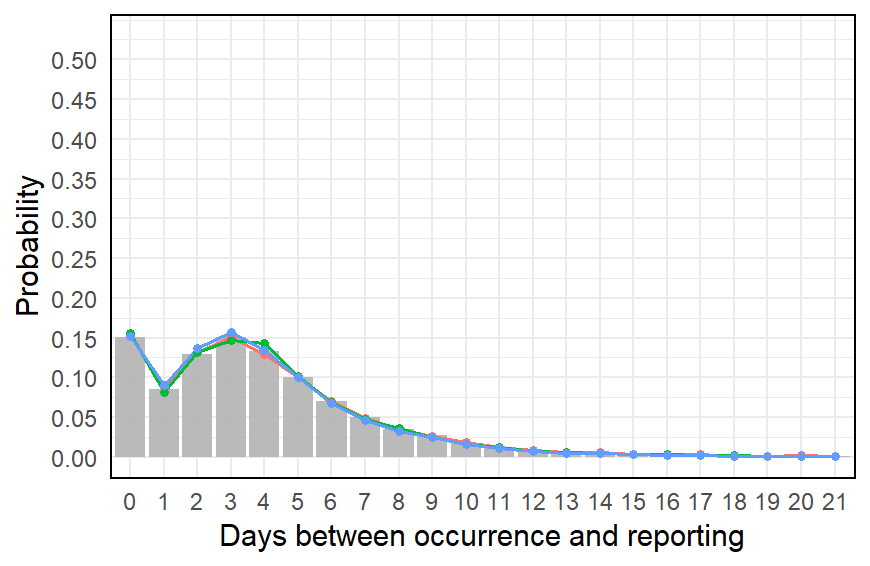}\\
{\footnotesize \textbf{(a.2)} Age 51-55}
\end{minipage}
\hfill
\begin{minipage}{0.32\textwidth}
\centering
\includegraphics[width=\textwidth]{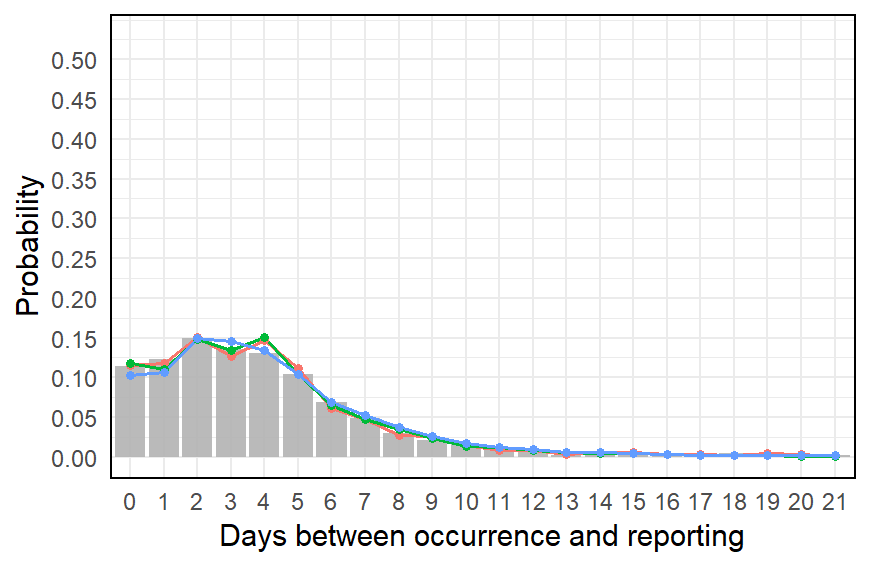}\\
{\footnotesize \textbf{(a.3)} Age 86-90}
\end{minipage}

\vspace{1em}

\textbf{(b) Out-of-sample} \\[0.5em]

\begin{minipage}{0.32\textwidth}
\centering
\includegraphics[width=\textwidth]{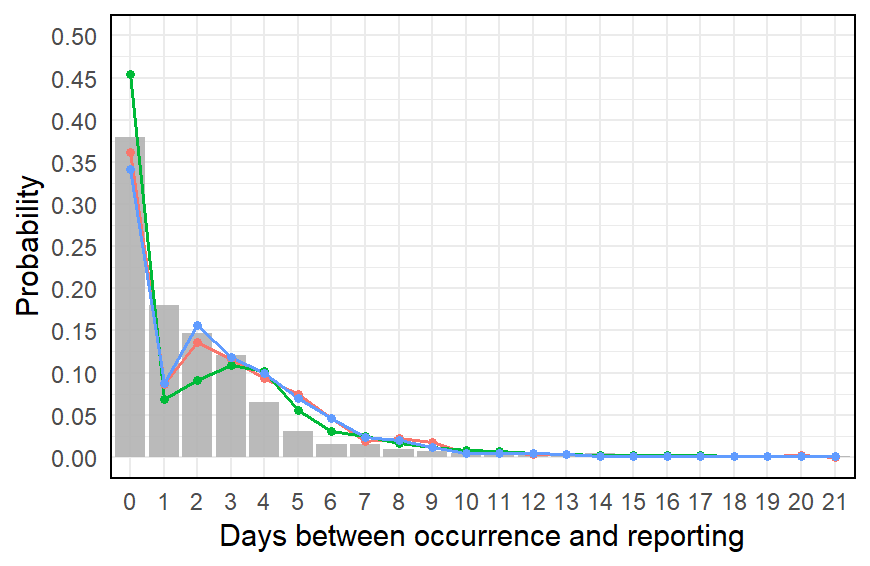}\\
{\footnotesize \textbf{(b.1)} Age 1-5}
\end{minipage}
\hfill
\begin{minipage}{0.32\textwidth}
\centering
\includegraphics[width=\textwidth]{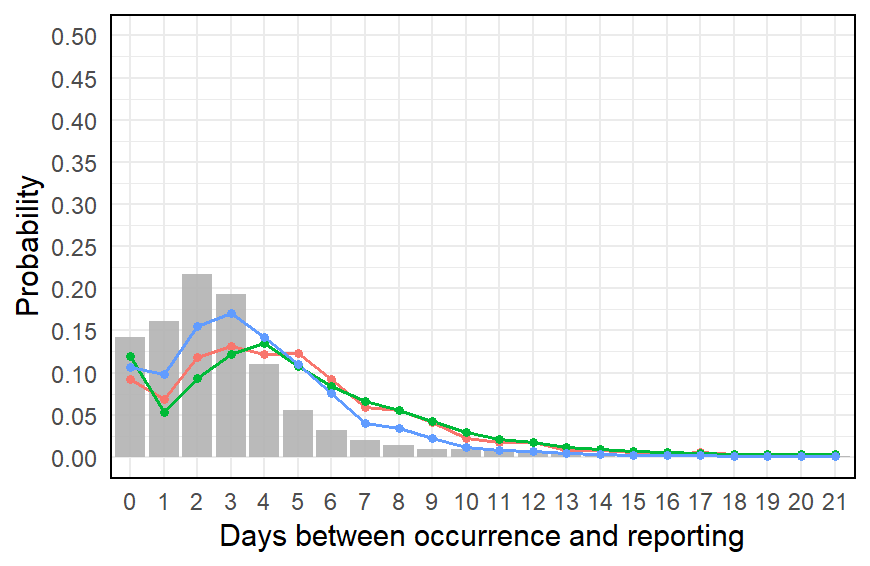}\\
{\footnotesize \textbf{(b.2)} Age 51-55}
\end{minipage}
\hfill
\begin{minipage}{0.32\textwidth}
\centering
\includegraphics[width=\textwidth]{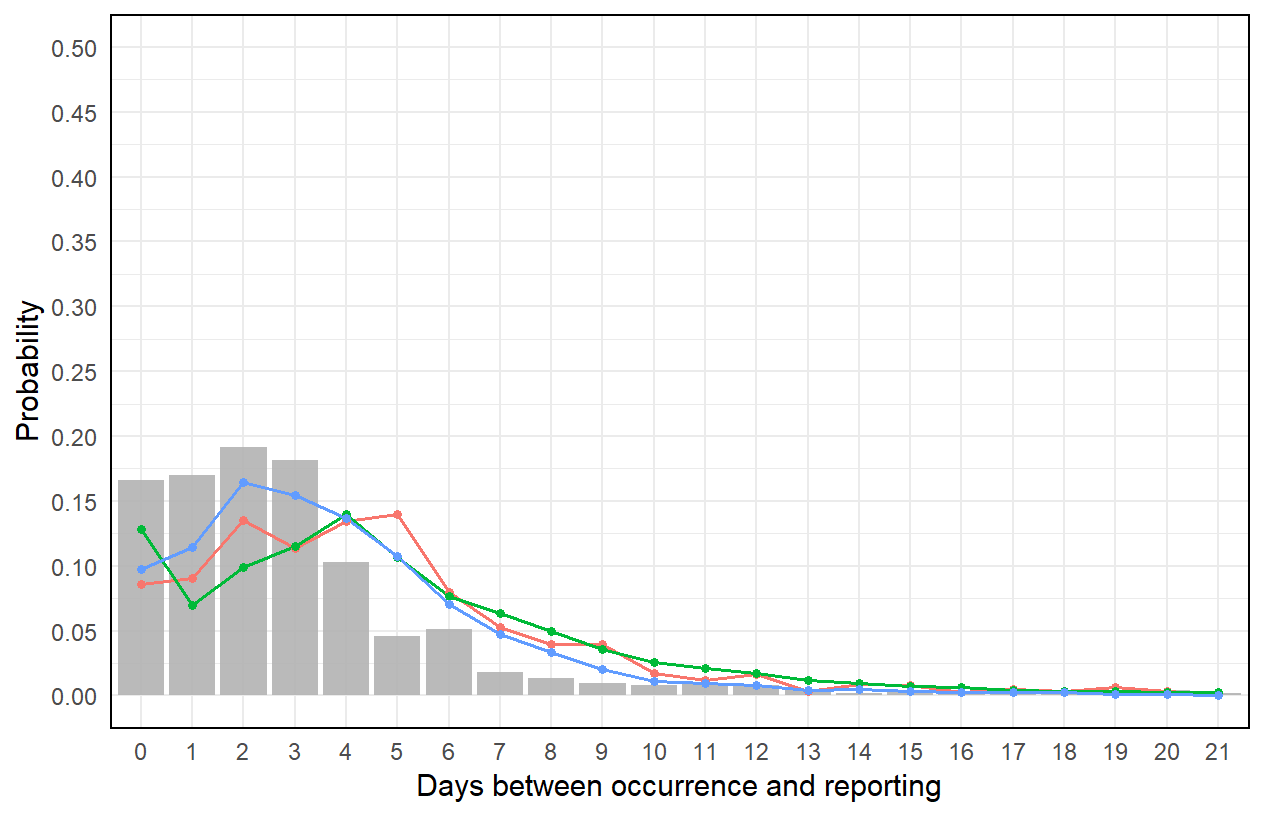}\\
{\footnotesize \textbf{(b.3)} Age 86-90}
\end{minipage}

\caption{In-sample (top panels) and out-of-sample (bottom panels) reporting probabilities for a young infant aged 1-5 (left panels), a middle aged person aged 51-55(middle panels) and an old person aged 86-90 (right panels). The grey bars show the empirical probabilities in the training and test set, respectively. The red, green and blue lines correspond to the estimated reporting probabilities from the GLMs, neural networks and XGBoost models, respectively.}
\label{fig:covidage}
\end{figure}

In Figure~\ref{fig:covidage}, we show the in-sample and out-of-sample estimated reporting probabilities for a young infant (aged 1-5), a middle-aged person (aged 51-55) and an old person (aged 86-90), an (out-of-sample) overview for all ages is provided in Figure~\ref{fig:covidage_full} in Appendix~\ref{app:covid}. We observe that reporting is significantly earlier for young infants compared to middle-aged and (and to a lesser extent) old people. This is to be expected since parents are more likely to see a doctor and get a clinical diagnosis for their child, compared to older people who can request a laboratory test themselves. Overall, all three approaches capture the patterns well, with XGBoost having a better fit, especially for the middle and old ages. Across all ages, we (empirically) observe earlier reporting in the out-of-sample period (2 October 2021 to 15 January 2022) compared to the time window used for training (1 May 2020 to 1 October 2020). This is most likely the result of latent time dynamics, such as increased familiarity with testing and the streamlining of the surveillance workflow as the pandemic progressed, which are difficult to capture with the available variables.

\begin{figure}[H] 
\centering
\subfloat[Monday 6 December 2021]{

\begin{adjustbox}{width=0.45\textwidth}
    \includegraphics[width = \textwidth]{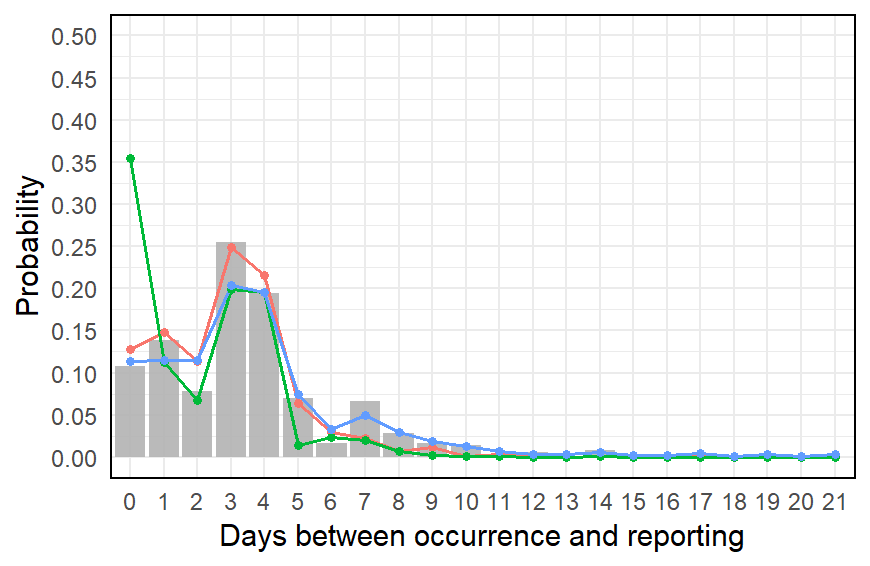}
\end{adjustbox}

}
\hspace{1.5em}
\subfloat[Monday 13 December 2021]{

\begin{adjustbox}{width=0.45\textwidth}
\includegraphics[width = \textwidth]{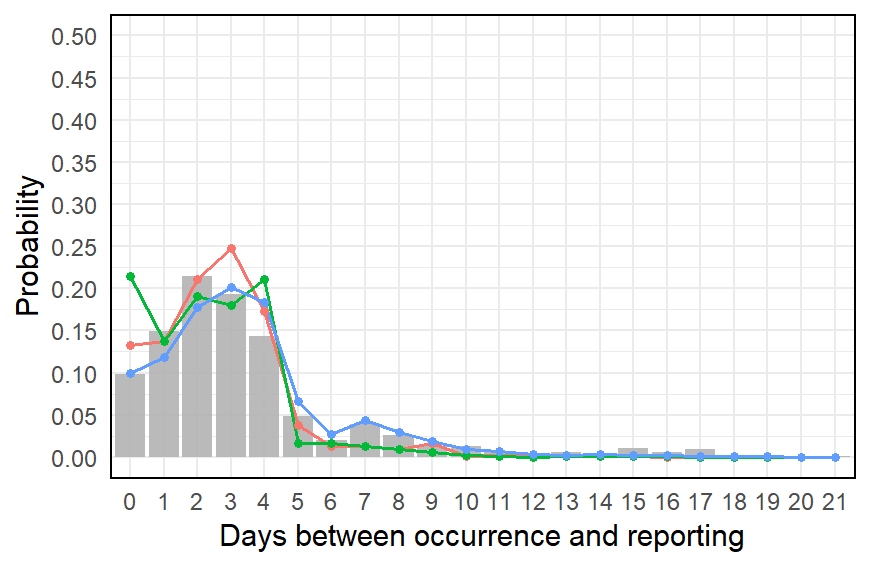}

\end{adjustbox}

}
\caption{Out-of-sample reporting probabilities for cases with symptom onset on 6 December 2021 (left panel) and 13 December 2021 (right panel). The grey bars show the empirical probabilities in the test set. The red, green and blue lines correspond to the estimated reporting probabilities from the GLMs, neural networks and XGBoost models, respectively.}   
\label{fig:covidholiday}
\end{figure}

Figure~\ref{fig:covidholiday} illustrates how the models capture lower empirical reporting on holidays. We look at all cases with symptom onset on Monday 6 December 2021 (left panel) and Monday 13 December 2021 (right panel). For the former, reporting with a two delay corresponds to 8 December 2021, which is a public holiday in Argentina. For the latter, there are no holidays during the possible reporting window. We observe a clear decrease in the estimated probabilities when reporting on a holiday (delay of two days in the left panel), which is not present otherwise. Additionally, we observe a drop in the reporting probabilities corresponding to weekend days (delay of 5 or 6 days in both panels of Figure~\ref{fig:covidholiday}). We illustrate this underreporting on weekend days more clearly in Figure~\ref{fig:coviddayofweek} in Appendix~\ref{app:covid}.

\begin{figure}[H]
\centering

\textbf{(a) In-sample} \\[0.5em]

\begin{minipage}{0.32\textwidth}
\centering
\includegraphics[width=\textwidth]{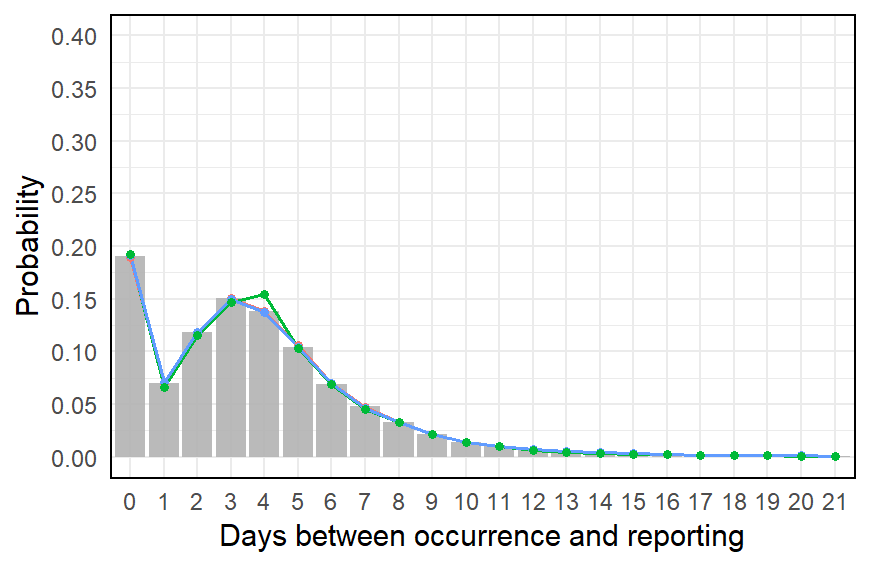}\\
{\footnotesize \textbf{(a.1)} Buenos Aires}
\end{minipage}
\hfill
\begin{minipage}{0.32\textwidth}
\centering
\includegraphics[width=\textwidth]{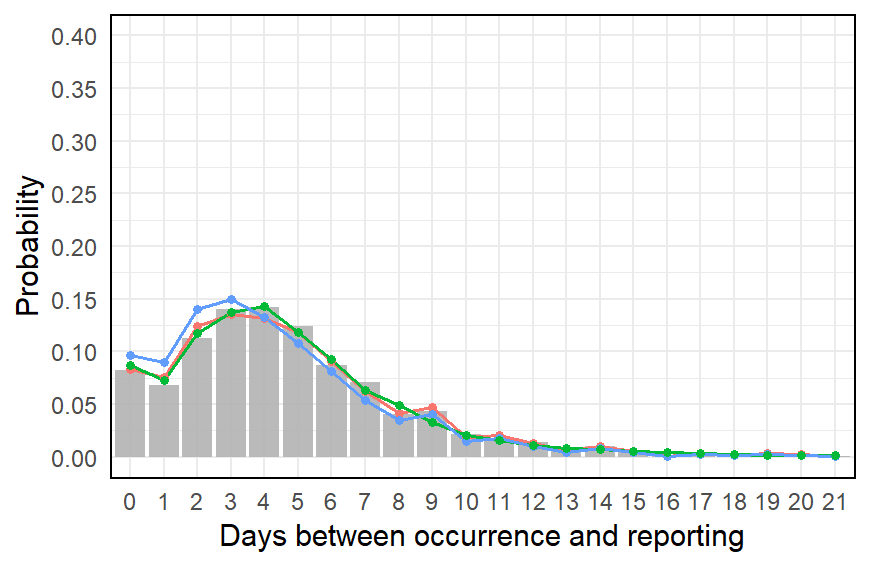}\\
{\footnotesize \textbf{(a.2)} Salta}
\end{minipage}
\hfill
\begin{minipage}{0.32\textwidth}
\centering
\includegraphics[width=\textwidth]{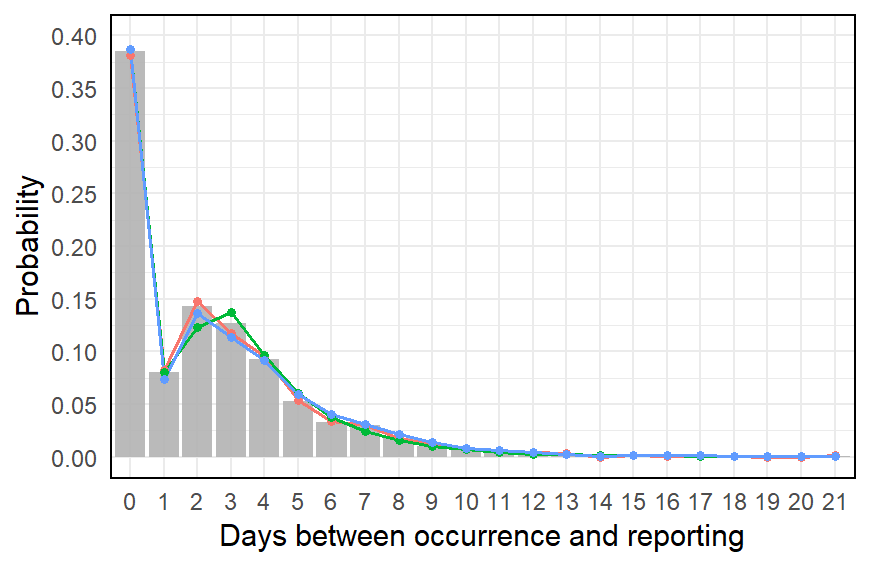}\\
{\footnotesize \textbf{(a.3)} Chubut}
\end{minipage}

\vspace{1em}

\textbf{(b) Out-of-sample} \\[0.5em]

\begin{minipage}{0.32\textwidth}
\centering
\includegraphics[width=\textwidth]{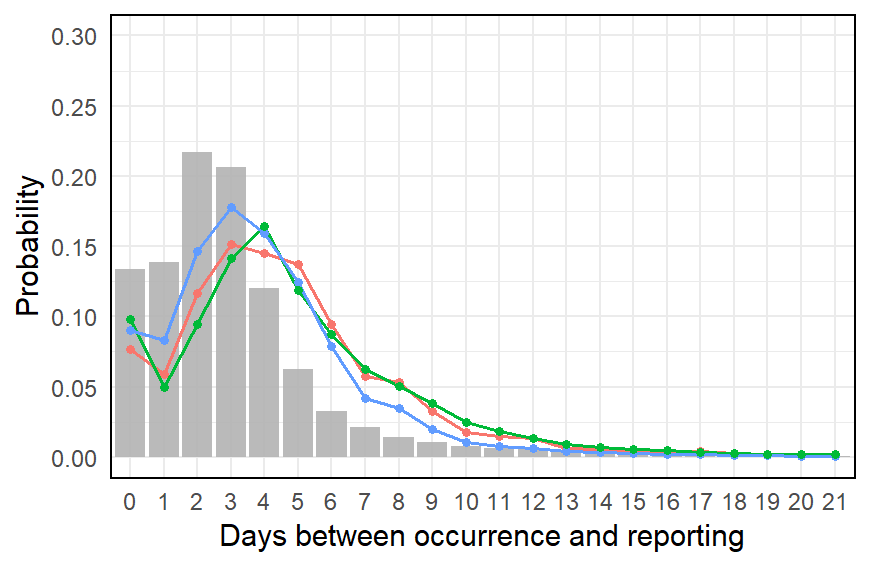}\\
{\footnotesize \textbf{(b.1)} Buenos Aires}
\end{minipage}
\hfill
\begin{minipage}{0.32\textwidth}
\centering
\includegraphics[width=\textwidth]{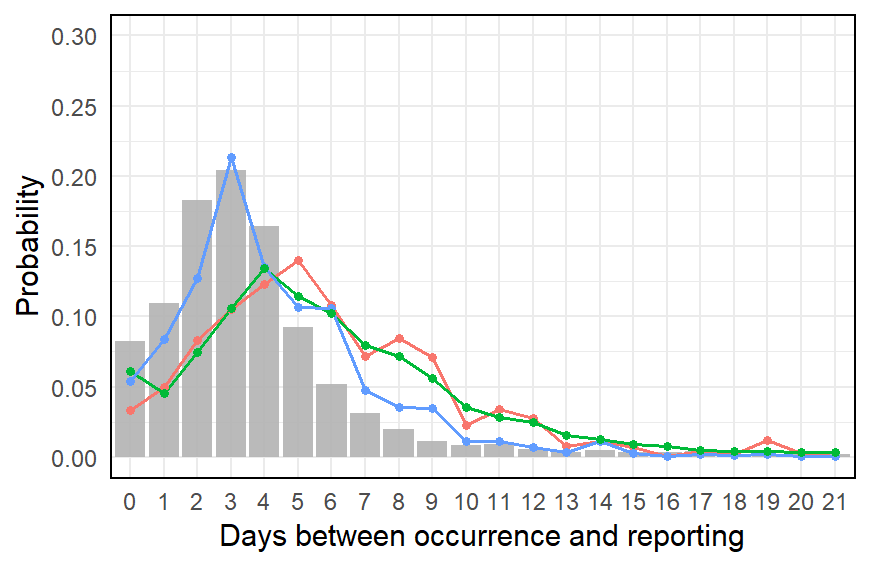}\\
{\footnotesize \textbf{(b.2)} Salta}
\end{minipage}
\hfill
\begin{minipage}{0.32\textwidth}
\centering
\includegraphics[width=\textwidth]{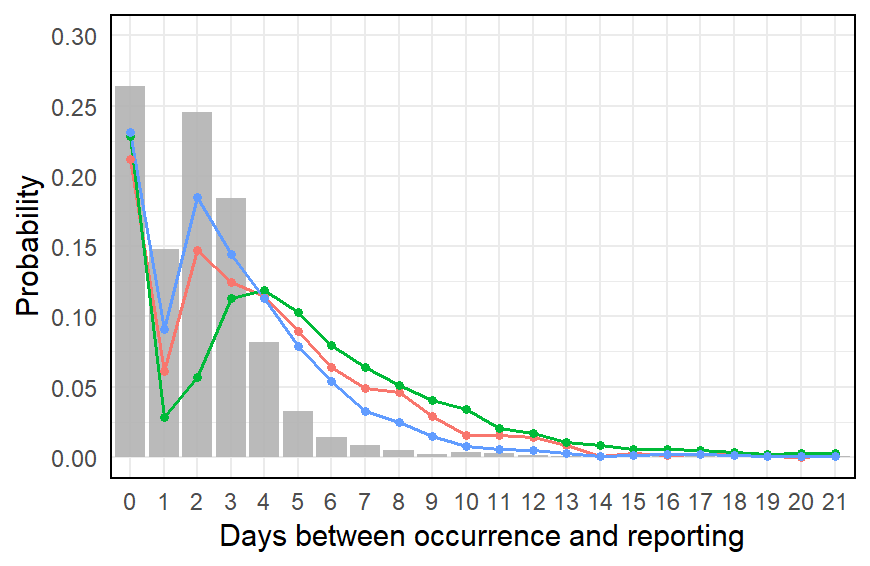}\\
{\footnotesize \textbf{(b.3)} Chubut}
\end{minipage}

\caption{In-sample (top panels) and out-of-sample (bottom panels) reporting probabilities for cases in Buenos Aires (left panels), in Salta (middle panels) and in Chubut (right panels). The grey bars show the empirical probabilities in the training and test set, respectively. The red, green and blue lines correspond to the estimated reporting probabilities from the GLMs, neural networks and XGBoost models, respectively.}
\label{fig:covidregionspecific}
\end{figure}

Reporting can significantly differ between regions. In Figure~\ref{fig:covidregionspecific}, we illustrate this for three regions in Argentina, see Figure~\ref{fig:covidlocation_full} in Appendix~\ref{app:covid} for all regions. We again observe the shift towards earlier reporting in the test period compared to the training window, as evidenced by the underestimation of the early reporting probabilities and overestimation of the tail. Overall, the XGBoost approach best captures the region-specific out-of-sample reporting probabilities. 

\begin{figure}[H]
\centering

\textbf{(a) 29 September 2021 ($\tau-2$)} \\[0.5em]

\begin{minipage}{0.32\textwidth}
\centering
\includegraphics[width=\textwidth]{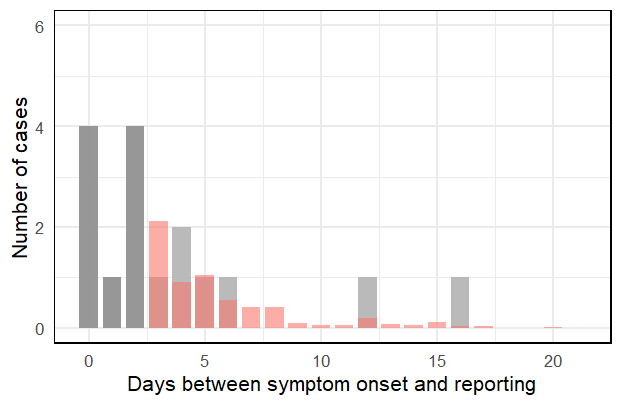}\\
{\footnotesize \textbf{(a.1)} GLM}
\end{minipage}
\hfill
\begin{minipage}{0.32\textwidth}
\centering
\includegraphics[width=\textwidth]{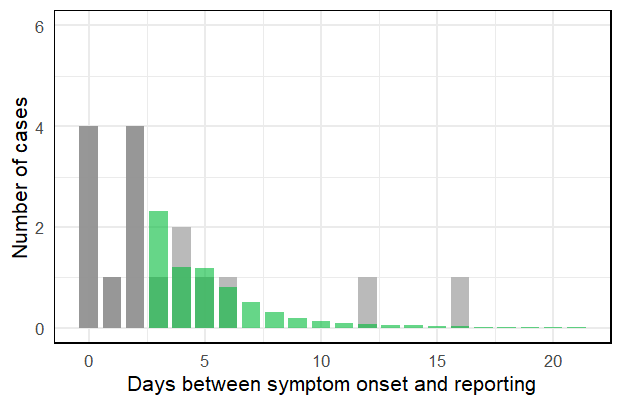}\\
{\footnotesize \textbf{(a.2)} Neural network}
\end{minipage}
\hfill
\begin{minipage}{0.32\textwidth}
\centering
\includegraphics[width=\textwidth]{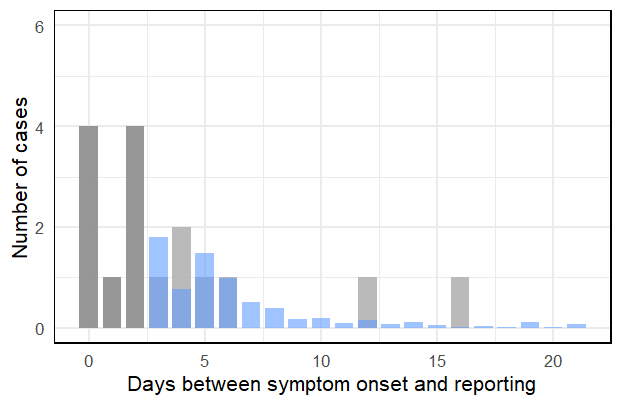}\\
{\footnotesize \textbf{(a.3)} XGBoost}
\end{minipage}

\vspace{1em}

\textbf{(b) 30 September 2021 ($\tau-1$)} \\[0.5em]

\begin{minipage}{0.32\textwidth}
\centering
\includegraphics[width=\textwidth]{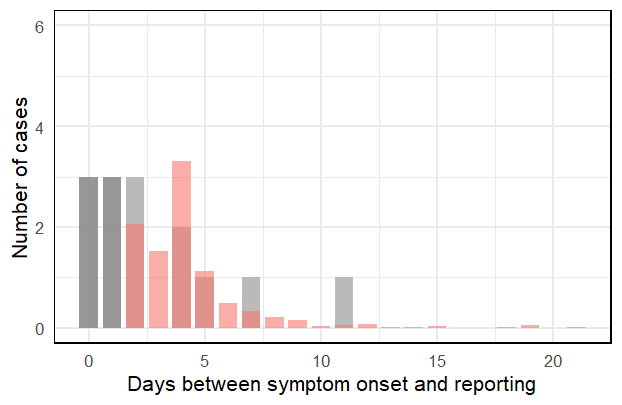}\\
{\footnotesize \textbf{(b.1)} GLM}
\end{minipage}
\hfill
\begin{minipage}{0.32\textwidth}
\centering
\includegraphics[width=\textwidth]{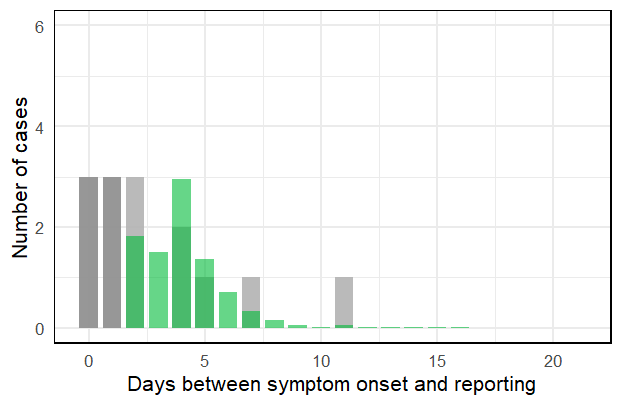}\\
{\footnotesize \textbf{(b.2)} Neural network}
\end{minipage}
\hfill
\begin{minipage}{0.32\textwidth}
\centering
\includegraphics[width=\textwidth]{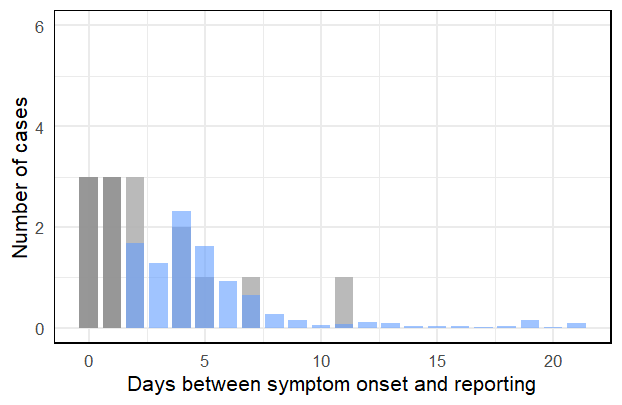}\\
{\footnotesize \textbf{(b.3)} XGBoost}
\end{minipage}

\vspace{1em}

\textbf{(c) 1 October 2021 ($\tau$)} \\[0.5em]

\begin{minipage}{0.32\textwidth}
\centering
\includegraphics[width=\textwidth]{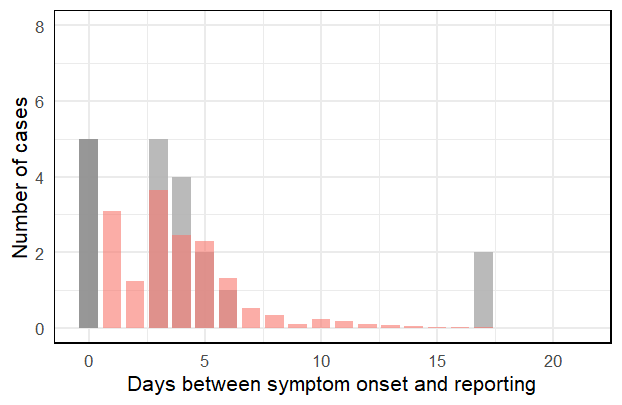}\\
{\footnotesize \textbf{(c.1)} GLM}
\end{minipage}
\hfill
\begin{minipage}{0.32\textwidth}
\centering
\includegraphics[width=\textwidth]{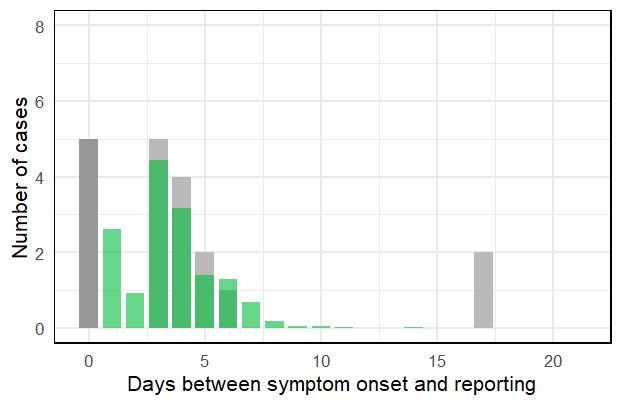}\\
{\footnotesize \textbf{(c.2)} Neural network}
\end{minipage}
\hfill
\begin{minipage}{0.32\textwidth}
\centering
\includegraphics[width=\textwidth]{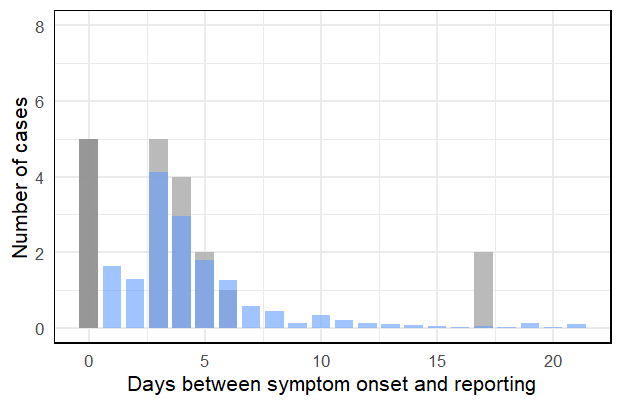}\\
{\footnotesize \textbf{(c.3)} XGBoost}
\end{minipage}

\caption{Empirically observed and predicted number of cases for 29 September 2021 (top panels), 30 September 2021 (middle panels) and 1 October 2021 (bottom panels). The dark grey bars show the events that are already reported at present time, while the light grey bars show the events that are not yet observable. The red, green and blue bars show the model predictions by the GLM, neural network and XGboost model, respectively.}
\label{fig:covid_events}
\end{figure}

In Figure~\ref{fig:covid_events}, for the last three days in our training window, we show the number of cases that are already reported (dark grey bars) and not yet reported (light grey bars) at time $\tau$. For cases not yet reported, we show the predicted count from the GLM (red), neural network (green) and XGBoost (blue) models. Despite the limited number of cases with symptom onset during this period, the models capture the ultimately realised scenario quite well. For example, no cases with symptom onset on Friday 1 October 2021 were reported with a delay of 1 or 2 days since these correspond to weekend days. The models reflect this by predicting a low number of reported cases at these delays, with XGBoost best capturing this effect. 
%

\section{Discussion} \label{sec:conclus}
In this paper, we propose an expectation-maximisation framework for nowcasting, in which we leverage machine learning techniques to estimate the parameters governing the occurrence and reporting processes of events with delayed reporting. The framework generalises existing GLM-based approaches to not require a functional form for the relation between the covariates and the occurrence intensities and the reporting probabilities. We adapt the EM algorithm such that information is effectively passed on between iterations when an XGBoost model or a neural network is used in the maximisation step. We then simulate data and compare the effectiveness of machine learning models with the GLM-based framework, when only linear effects are present as well as when there are both linear and non-linear effects. We find that using an additive XGBoost approach results in the most accurate parameter estimates both for the occurrence as well as the reporting process. By applying our framework to the reporting of Covid-19 cases in Argentina, we are able to gain insight in the reporting structure both at the individual and regional level, with XGBoost again achieving the best out-of-sample fit. Future research can investigate a hybrid approach, i.e.~by using a different type of model (GLM, neural network, XGBoost,$\ldots$) for the occurrence and the reporting process, respectively. A second direction for future research is to explore the use of diffusion models \citep{luo2022understanding} to model events with reported delay, as it has been successfully applied to other missing data problems, see for example \cite{zheng2022diffusion}.

\biblist

\pagebreak
\appendix

\section{Algorithmic details} \label{append:model}
Following Section~\ref{sec:method} of the main document, we split the dataset $D$ in three parts. A dataset $D_{\text{train}}$ to train the occurrence and reporting model, a first validation dataset $D_{\text{val},1}$ for early stopping within training of the occurrence and reporting model and a second validation dataset $D_{\text{val},2}$ for early stopping across EM iterations. The latter is also used to determine the optimal values for the tuning parameters (for the neural network- and XGBoost approach). After the $k$th expectation step, the expectation of the corresponding complete datasets $\mathcal{D}_{\text{train}}$, $\mathcal{D}_{\text{val},1}$ and $\mathcal{D}_{\text{val},2}$ are referred to as $\mathcal{D}^{(k)}_{\text{train}}$, $\mathcal{D}^{(k)}_{\text{val},1}$ and $\mathcal{D}^{(k)}_{\text{val},2}$, respectively.\\
\begin{adjustbox}{width=0.82\textwidth}
   \begin{algorithm}[H] \label{algo:EM}
    \caption{\textbf{Pseudocode model-agnostic EM algorithm}}

Initialise $\lambda^{(0)}_{i}$ and $p_{i,j}^{(0)}$ for $i=1,\ldots,n$, and $j=1,\ldots,d$.

\For{$k=1,\ldots,K$}{
\vspace{0.2cm}

\textbf{Expectation step}

Construct $\mathcal{D}^{(k)}=(\boldsymbol{N}_{i}^{(k)},\boldsymbol{x}_i)_{i=1}^{n}$ with $\boldsymbol{N}_{i}^{(k)}=(N_{i,1}^{(k)},\ldots,N_{i,d}^{(k)})$ by defining
\begin{align*}
    N_{i,j}^{(k)}= \begin{cases}N_{i,j} & j \leq \tau_i \\ \widehat{\lambda}^{(k-1)}(\boldsymbol{x}_i) \cdot \widehat{p}_{j}^{(k-1)}(\boldsymbol{x_i}) & \tau_i<j \leq d\end{cases},
\end{align*}
where $\widehat{\lambda}^{(k-1)}(\boldsymbol{x}_i) = \exp{(\widehat{f}^{(k-1)}_{\text{occ}}(\boldsymbol{x}_i))}$ and $\widehat{p}_{j}^{(k-1)}(\boldsymbol{x}_i)= \frac{\exp(\widehat{f}_{\text{rep}}^{(k-1)[j]}(\boldsymbol{x}_{i}))}{\sum_{z=1}^{d} \exp(\widehat{f}_{\text{rep}}^{(k-1)[z]}(\boldsymbol{x}_{i}))}$.

\textbf{Maximisation step} \\
\vspace{0.2cm}
\begin{adjustwidth}{0.5cm}{0cm}
\textbf{Occurrence intensities} \\
Learn a predictive function $f^{(k)}_{\text{occ}}(.)$, see details in Section~\ref{append:modelocc}, by maximising the expectation of the unobserved log-likelihood
\begin{align*}
    Q^{\text{occ}}(\lambda_{i}^{(k)};\mathcal{D}_{\text{train}}^{(k)}) &= \mathbb{E}\left(LL_{c}^{\text{occ}}(\lambda_{i}^{(k)};\mathcal{D}) | \mathcal{D}_{\text{train}}^{(k)}\right)\\
    &= \sum_{i=1}^n\left[ -\lambda_{i}^{(k)}+N^{(k)}_i \cdot \log \left(\lambda_{i}^{(k)}\right) \right].
\end{align*}
and set the predicted occurrence intensities
\begin{align*}
    \widehat{\lambda}^{(k)}(\boldsymbol{x}_i) = \exp{(\widehat{f}^{(k)}_{\text{occ}}(\boldsymbol{x}_i))}.
\end{align*}
\end{adjustwidth}
\vspace{0.2cm}
\begin{adjustwidth}{0.5cm}{0cm}
\textbf{Reporting probabilities} \\
    Learn a predictive function $f^{(k)}_{\text{rep}}(.)=(f^{(k)[1]}_{\text{rep}}(.),\ldots,f^{(k)[d]}_{\text{rep}}(.))$, see details in Section~\ref{append:modelrep},\\ by maximising the expectation of the unobserved log-likelihood
\begin{align*}
    Q^{\text{rep}}(p_{i,j}^{(k)};\mathcal{D}_{\text{train}}^{(k)})&= \mathbb{E}\left(LL_{c}^{\text{rep}}(p_{i,j}^{(k)};\mathcal{D}) | \mathcal{D}_{\text{train}}^{(k)}\right)\\ &= \sum_{i=1}^n \sum_{j=1}^d N^{(k)}_{i, j} \cdot \log \left(p_{i,j}^{(k)}\right), \:\:\: \text{subject to} \:\:\: \sum_{j=1}^d p_{i,j}^{(k)} = 1 \:\:\: \text{for} \:\:\: i=1,\ldots,n,
\end{align*}
and set the estimated reporting probabilities
\begin{align*}
    \widehat{p}_{j}^{(k)}(\boldsymbol{x}_i) = \frac{\exp(\widehat{f}_{\text{rep}}^{(k)[j]}(\boldsymbol{x}_{i}))}{\sum_{z=1}^{d} \exp(\widehat{f}_{\text{rep}}^{(k)[z]}(\boldsymbol{x}_{i}))}.
\end{align*}
\end{adjustwidth}

\If{$k>\texttt{EM\_patience}$}{

Stop EM algorithm if the following condition is satisfied:

\begin{align*}
    LL(\widehat{\lambda}^{(k)}_i, \widehat{p}^{(k)}_{i,j};D_{\text{val},2}) = \displaystyle\max_{z = k-\texttt{EM\_patience}, \dots, k} LL(\widehat{\lambda}^{(z)}_i, \widehat{p}^{(z)}_{i,j};D_{\text{val},2}).
\end{align*}

}

}

\end{algorithm} 
\end{adjustbox}

\subsection{Occurrence model} \label{append:modelocc}

In the $k$th iteration of the EM algorithm as described in Algorithm~\ref{algo:EM} in Section~\ref{sec:method}, we consider 

\begin{align*}
    -Q^{\text{occ}}(\lambda_{i}^{(k)};\mathcal{D}^{(k)}) = \sum_{i=1}^n\left[ \lambda_{i}^{(k)}-N^{(k)}_i \cdot \log \left(\lambda_{i}^{(k)}\right) \right]
\end{align*}
as the loss function for the occurrence model, we set $\lambda_{i}^{(k)} = \exp(f_{\text{occ}}^{(k)}(\boldsymbol{x}_{i}))$ to ensure the positivity of the predicted occurrence intensities.

\paragraph{XGBoost} Algorithm~\ref{algo:XGBocc} details the training algorithm of the occurrence model to learn the occurrence intensities $\lambda(\boldsymbol{x}_i)$ for $i=1,\ldots,n$ when relying on a XGBoost model. For notational purposes, we set $f_{\text{occ}}^{(k)}(\boldsymbol{x}_{i}) \overset{\text{not.}}{=} f_{\text{occ}(T^{(k)}_{\text{occ}})}^{(k)}(\boldsymbol{x}_{i})$ with $T^{(k)}_{\text{occ}}$ denoting the number of iterations used for the XGBoost algorithm in the $k$th iteration of the EM algorithm. After training, the resulting predicted occurrence intensities are expressed as
\begin{align*}
    \widehat{\lambda}^{(k)}(\boldsymbol{x}_i) = \exp{(\widehat{f}_{\text{occ}}^{(k)}(\boldsymbol{x}_{i}))}.
\end{align*}

\begin{algorithm}[H] \label{algo:XGBocc}
    \caption{\textbf{Pseudocode occurrence XGBoost model}}

Set $\widehat{f}_{\text{occ}(0)}^{(k)}(\boldsymbol{x}_{i})=\widehat{f}_{\text{occ}}^{(k-1)}(\boldsymbol{x}_{i})$ for $i=1,\ldots,n$. 

\For{$t=1,\ldots,T^{(k)}_{\text{occ}}$}{
\vspace{0.2cm}
Calculate the gradient $g_{t}(\boldsymbol{x}_{i})$ and the hessian $h_{t}(\boldsymbol{x}_{i})$ for $i=1,\ldots,n$:
\begin{align*}
    g_{t}(\boldsymbol{x}_{i}) &= \left[-\frac{\partial Q^{\text{occ}}(\lambda_{i}^{(k)}|\mathcal{D}_{\text{train}}^{(k)})}{\partial f_{\text{occ}}^{(k)}(\boldsymbol{x}_{i})}\right]_{f_{\text{occ}}^{(k)}(\boldsymbol{x}_{i})=\widehat{f}_{\text{occ}(t-1)}^{(k)}(\boldsymbol{x}_{i})} = \exp(\widehat{f}_{\text{occ}(t-1)}^{(k)}(\boldsymbol{x}_{i}))-N^{(k)}_i,\\
   h_{t}(\boldsymbol{x}_{i}) &= \left[-\frac{\partial^2 Q^{\text{occ}}(\lambda_{i}^{(k)}|\mathcal{D}_{\text{train}}^{(k)})}{\partial f_{\text{occ}}^{(k)}(\boldsymbol{x}_{i})^2}\right]_{f_{\text{occ}}^{(k)}(\boldsymbol{x}_{i})=\widehat{f}_{\text{occ}(t-1)}^{(k)}(\boldsymbol{x}_{i})}= \exp(\widehat{f}_{\text{occ}(t-1)}^{(k)}(\boldsymbol{x}_{i})) .
\end{align*}

Learn a regression tree $\delta_{t}^{(k)}$ of a fixed size \texttt{tree\_depth\_occ}
\begin{align*}
    \widehat{\delta}_{t}^{(k)}=\underset{\delta_{t}^{(k)}}{\operatorname{argmin}} \frac{1}{n} \sum_{i=1}^n \frac{1}{2} h_{t}(\boldsymbol{x}_{i})\left( \delta_{t}^{(k)}(\boldsymbol{x}_i) - \frac{g_{t}(\boldsymbol{x}_{i})}{h_{t}(\boldsymbol{x}_{i})} \right)^2.
\end{align*}

Update the predictions:
\begin{align*}
    \widehat{f}_{\text{occ}(t)}^{(k)}(\boldsymbol{x}_{i})=\widehat{f}_{\text{occ}(t-1)}^{(k)}(\boldsymbol{x}_{i})+\texttt{eta\_occ} \cdot \widehat{\delta}_{t}^{(k)}(\boldsymbol{x}_{i}),
\end{align*}
where \texttt{eta\_occ} denotes the learning rate.\\
\If{$t>\texttt{xgb\_patience}$}{

Stop training if the following condition is satisfied:

\begin{align*}
    Q^{\text{occ}}(\exp{(\widehat{f}_{\text{occ}(t)}^{(k)}(\boldsymbol{x}_{i}))}|\mathcal{D}_{\text{val},1}^{(k)}) = \displaystyle\max_{z = t-\texttt{xgb\_patience}, \dots, t} Q^{\text{occ}}(\exp{(\widehat{f}_{\text{occ}(z)}^{(k)}(\boldsymbol{x}_{i}))}|\mathcal{D}_{\text{val},1}^{(k)}).
\end{align*}

}
}

\end{algorithm}
\paragraph{Neural network}

\begin{figure}[H]
    \centering

    \scalebox{0.9}{

\tikzset{every picture/.style={line width=0.75pt}} 

\begin{tikzpicture}[x=0.75pt,y=0.75pt,yscale=-1,xscale=1]

\draw  [fill={rgb, 255:red, 155; green, 155; blue, 155 }  ,fill opacity=0.56 ] (46,36.81) .. controls (46,23) and (57.19,11.81) .. (71,11.81) .. controls (84.81,11.81) and (96,23) .. (96,36.81) .. controls (96,50.61) and (84.81,61.81) .. (71,61.81) .. controls (57.19,61.81) and (46,50.61) .. (46,36.81) -- cycle ;
\draw  [fill={rgb, 255:red, 155; green, 155; blue, 155 }  ,fill opacity=0.56 ] (46,135.81) .. controls (46,122) and (57.19,110.81) .. (71,110.81) .. controls (84.81,110.81) and (96,122) .. (96,135.81) .. controls (96,149.61) and (84.81,160.81) .. (71,160.81) .. controls (57.19,160.81) and (46,149.61) .. (46,135.81) -- cycle ;
\draw [line width=1.5]  [dash pattern={on 1.69pt off 2.76pt}]  (71,73.81) -- (71.04,102.44) ;
\draw  [fill={rgb, 255:red, 155; green, 155; blue, 155 }  ,fill opacity=0.56 ] (46,225.81) .. controls (46,212) and (57.19,200.81) .. (71,200.81) .. controls (84.81,200.81) and (96,212) .. (96,225.81) .. controls (96,239.61) and (84.81,250.81) .. (71,250.81) .. controls (57.19,250.81) and (46,239.61) .. (46,225.81) -- cycle ;
\draw  [fill={rgb, 255:red, 155; green, 155; blue, 155 }  ,fill opacity=0.56 ] (46,324.81) .. controls (46,311) and (57.19,299.81) .. (71,299.81) .. controls (84.81,299.81) and (96,311) .. (96,324.81) .. controls (96,338.61) and (84.81,349.81) .. (71,349.81) .. controls (57.19,349.81) and (46,338.61) .. (46,324.81) -- cycle ;
\draw [line width=1.5]  [dash pattern={on 1.69pt off 2.76pt}]  (71,262.81) -- (71.04,291.44) ;
\draw  [fill={rgb, 255:red, 248; green, 231; blue, 28 }  ,fill opacity=0.39 ] (165,87.81) .. controls (165,74) and (176.19,62.81) .. (190,62.81) .. controls (203.81,62.81) and (215,74) .. (215,87.81) .. controls (215,101.61) and (203.81,112.81) .. (190,112.81) .. controls (176.19,112.81) and (165,101.61) .. (165,87.81) -- cycle ;
\draw  [fill={rgb, 255:red, 248; green, 231; blue, 28 }  ,fill opacity=0.39 ] (165,277.81) .. controls (165,264) and (176.19,252.81) .. (190,252.81) .. controls (203.81,252.81) and (215,264) .. (215,277.81) .. controls (215,291.61) and (203.81,302.81) .. (190,302.81) .. controls (176.19,302.81) and (165,291.61) .. (165,277.81) -- cycle ;
\draw    (107.4,39.59) -- (156.77,77.25) ;
\draw    (157.57,102.05) -- (107.54,138.84) ;
\draw    (107.4,225.79) -- (156.77,263.45) ;
\draw    (157.57,288.25) -- (107.54,325.04) ;
\draw [line width=1.5]  [dash pattern={on 1.69pt off 2.76pt}]  (190.37,170.59) -- (190.41,199.22) ;
\draw    (107.17,206.74) -- (157.57,117.94) ;
\draw    (157.17,242.34) -- (107.57,155.94) ;
\draw    (107.4,57.07) -- (156.77,216.34) ;
\draw    (107.4,301.07) -- (157.57,143.38) ;
\draw  [fill={rgb, 255:red, 248; green, 231; blue, 28 }  ,fill opacity=0.39 ] (339,87.54) .. controls (339,73.74) and (350.19,62.54) .. (364,62.54) .. controls (377.81,62.54) and (389,73.74) .. (389,87.54) .. controls (389,101.35) and (377.81,112.54) .. (364,112.54) .. controls (350.19,112.54) and (339,101.35) .. (339,87.54) -- cycle ;
\draw  [fill={rgb, 255:red, 248; green, 231; blue, 28 }  ,fill opacity=0.39 ] (339,277.54) .. controls (339,263.74) and (350.19,252.54) .. (364,252.54) .. controls (377.81,252.54) and (389,263.74) .. (389,277.54) .. controls (389,291.35) and (377.81,302.54) .. (364,302.54) .. controls (350.19,302.54) and (339,291.35) .. (339,277.54) -- cycle ;
\draw [line width=1.5]  [dash pattern={on 1.69pt off 2.76pt}]  (365.37,170.59) -- (365.41,199.22) ;
\draw    (227.4,87.82) -- (256.77,87.88) ;
\draw    (227.4,277.82) -- (256.77,277.88) ;
\draw    (227.4,107.42) -- (257.57,246.28) ;
\draw    (256.77,107.22) -- (226.6,246.08) ;
\draw   (46.52,384.34) .. controls (46.54,389.01) and (48.88,391.33) .. (53.55,391.31) -- (59.67,391.29) .. controls (66.34,391.27) and (69.68,393.59) .. (69.7,398.26) .. controls (69.68,393.59) and (73,391.25) .. (79.67,391.23)(76.67,391.24) -- (86.74,391.2) .. controls (91.41,391.19) and (93.73,388.85) .. (93.72,384.18) ;
\draw   (162.28,383.94) .. controls (162.27,388.61) and (164.6,390.94) .. (169.27,390.95) -- (259.49,391.08) .. controls (266.16,391.09) and (269.49,393.42) .. (269.48,398.09) .. controls (269.49,393.42) and (272.82,391.1) .. (279.49,391.11)(276.49,391.1) -- (383.6,391.25) .. controls (388.27,391.26) and (390.6,388.93) .. (390.61,384.26) ;
\draw  [dash pattern={on 4.5pt off 4.5pt}] (41,8.65) -- (100.77,8.65) -- (100.77,164.58) -- (41,164.58) -- cycle ;
\draw  [dash pattern={on 4.5pt off 4.5pt}] (41,197.58) -- (100.77,197.58) -- (100.77,353.51) -- (41,353.51) -- cycle ;
\draw    (295.4,87.82) -- (324.77,87.88) ;
\draw    (295.4,277.82) -- (324.77,277.88) ;
\draw    (295.4,107.42) -- (325.57,246.28) ;
\draw    (324.77,107.22) -- (294.6,246.08) ;
\draw [line width=1.5]  [dash pattern={on 1.69pt off 2.76pt}]  (283.17,87.99) -- (267.58,88.05) ;
\draw [line width=1.5]  [dash pattern={on 1.69pt off 2.76pt}]  (283.17,277.99) -- (267.58,278.05) ;
\draw  [fill={rgb, 255:red, 184; green, 233; blue, 134 }  ,fill opacity=0.55 ] (441,183.54) .. controls (441,169.74) and (452.19,158.54) .. (466,158.54) .. controls (479.81,158.54) and (491,169.74) .. (491,183.54) .. controls (491,197.35) and (479.81,208.54) .. (466,208.54) .. controls (452.19,208.54) and (441,197.35) .. (441,183.54) -- cycle ;
\draw    (396.39,91) -- (436.26,157.65) ;
\draw    (396.39,276.3) -- (436.26,209.65) ;
\draw   (442.86,383.84) .. controls (442.87,388.51) and (445.21,390.83) .. (449.88,390.81) -- (456.01,390.79) .. controls (462.68,390.77) and (466.02,393.09) .. (466.03,397.76) .. controls (466.02,393.09) and (469.34,390.75) .. (476.01,390.73)(473.01,390.74) -- (483.08,390.7) .. controls (487.75,390.69) and (490.07,388.35) .. (490.05,383.68) ;

\draw (33,400) node [anchor=north west][inner sep=0.75pt]  [font=\normalsize] [align=left] {Input layer};
\draw (225,400) node [anchor=north west][inner sep=0.75pt]  [font=\normalsize] [align=left] {Hidden layers};
\draw (3,268.13) node [anchor=north west][inner sep=0.75pt] [font=\Large]  [align=left] {$\boldsymbol{x}_{\text{per}}$};
\draw (3,77.53) node [anchor=north west][inner sep=0.75pt] [font=\Large] [align=left]{$\boldsymbol{x}_{\text{ent}}$};
\draw (423,400) node [anchor=north west][inner sep=0.75pt]  [font=\normalsize] [align=left] {Output layer};
\draw (176,75) node [anchor=north west][inner sep=0.75pt] [font=\large] [align=left] {$\theta_{1,1}^{(k)}$};
\draw (348,75) node [anchor=north west][inner sep=0.75pt] [font=\large] [align=left] {$\theta_{M,1}^{(k)}$};
\draw (344,265) node [anchor=north west][inner sep=0.75pt] [font=\large] [align=left] {$\theta_{M,q_{M}}^{(k)}$};
\draw (174,265) node [anchor=north west][inner sep=0.75pt] [font=\large] [align=left] {$\theta_{1,q_{1}}^{(k)}$};
\draw (444,171) node [anchor=north west][inner sep=0.75pt] [font=\normalsize] [align=left] {$\theta_{M+1,1}^{(k)}$};

\end{tikzpicture}
}
    
    \caption{Network architecture for the occurrence model in the $k$th iteration of the EM algorithm. The grey nodes represent the input neurons, the yellow nodes represent the neurons $\theta_{m,u}^{(k)}(\boldsymbol{z})$ for $m=1,\ldots,M$ and $u=1,\ldots,q_m$ in the $M$ hidden layers. The output layer consists of a single neuron $\theta_{M+1,1}^{(k)}$, represented by the green node. }
    \label{fig:NNocc}
\end{figure}
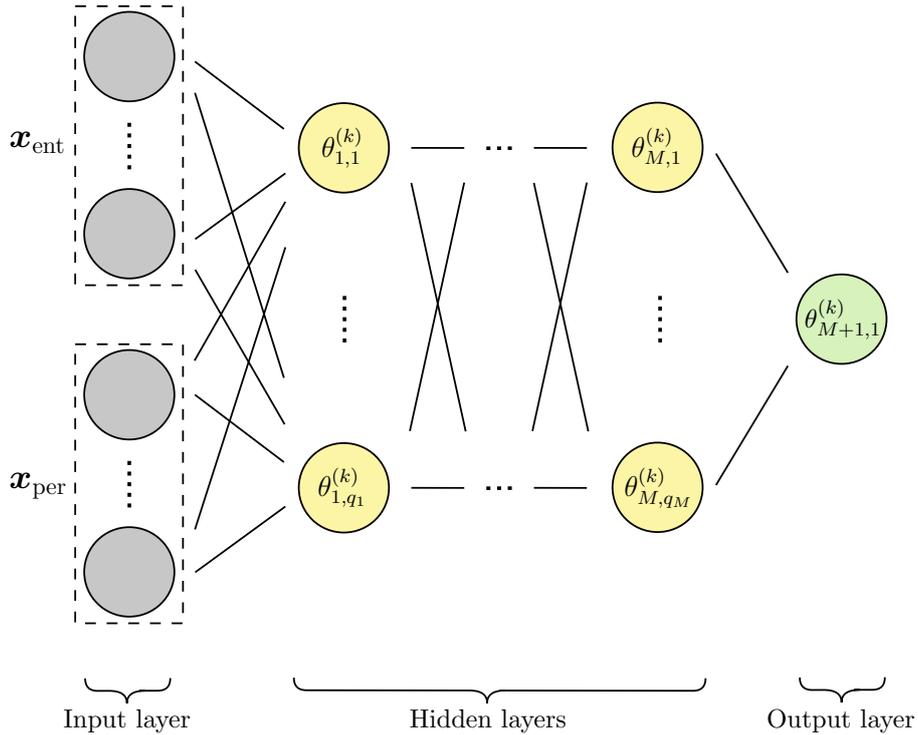

 The network parameters are learned using the \texttt{adam} algorithm \citep{kingma2014adam}. We denote the number of training epochs, the batch size and the early stopping patience as \texttt{n\_epoch}, \texttt{batch\_size} and \texttt{nn\_patience}, respectively. After training, the resulting predicted occurrence intensities are expressed as 
\begin{align*}
    \widehat{\lambda}^{(k)}(\boldsymbol{x}_i) =\widehat{\theta}^{(k)}_{M+1,1}(\boldsymbol{x}_i)= \exp{(\widehat{f}_{\text{occ}}^{(k)}(\boldsymbol{x}_{i}))} = \exp{(\widehat{b}^{(k)}_{M+1,1}+\left\langle\widehat{\boldsymbol{w}}^{(k)}_{M+1,1},\left(\widehat{\theta}^M \circ \cdots \circ \widehat{\theta}^1\right)(\boldsymbol{x}_i)\right\rangle)}. 
\end{align*}
In Figure~\ref{fig:NNocc}, we visualise the network architecture to learn the occurrence intensities as described in Section~\ref{sec:method}.

\subsection{Reporting model} \label{append:modelrep}

The \texttt{xgboost} and \texttt{keras} implementation in R, which we use to learn the occurrence and reporting models, require a separate observation for each event that has occurred for a specific entity-occurrence date combination. Therefore, in the $k$th iteration of the EM algorithm, we transform the dataset $\mathcal{D}^{(k)}$ such that the likelihood remains the same. Specifically, we define the dataset $\overline{\mathcal{D}}^{(k)}=(\overline{y}_{i}^{(k)},\overline{\boldsymbol{x}}^{(k)}_i,w_i^{(k)})_{i=1}^{\overline{n}^{(k)}}$ with $\overline{n}^{(k)}= \sum_{i=1}^{n} \sum_{j=1}^{d} \lceil N_{i,j}^{(k)} \rceil$ where $\lceil N_{i,j}^{(k)} \rceil$ is the smallest integer greater or equal to $N_{i,j}^{(k)}$ and with $\overline{y}_{i}^{(k)} \in \{1,\ldots,d\}$ denoting the reporting delay for the corresponding event. The covariate vector $\overline{\boldsymbol{x}}^{(k)}_i$ corresponds to the covariate vector of the entity-occurrence date combination for which the event occurred. We use $w_i$ to denote the weight related to an observation, since the event counts $N^{(k)}_{i,j}$ can be decimal values. For example, if $N^{(k)}_{i^{*},2}=2.5$, we define three event observations $(2,\boldsymbol{x}_{i^{*}},1)$, $(2,\boldsymbol{x}_{i^{*}},1)$ and $(2,\boldsymbol{x}_{i^{*}},0.5)$, respectively. By construction, we then have that
\begin{align*}
    \sum_{i=1}^{\overline{n}^{(k)}} w_{i}^{(k)} \sum_{j=1}^d   y^{(k)}_{i, j} \cdot \log \left(p_{j}^{(k)}(\overline{\boldsymbol{x}}^{(k)}_i)\right)=\sum_{i=1}^n \sum_{j=1}^d N^{(k)}_{i, j} \cdot \log \left(p_{j}^{(k)}(\boldsymbol{x}_i)\right) \:\: \text{with} \:\: y^{(k)}_{i, j} =
\begin{cases}
1, & \text{if }\:\: \overline{y}^{(k)}_i = j \\
0, & \text{else}.
\end{cases}
\end{align*}
In the $k$th iteration of the EM algorithm as described in Section~\ref{sec:method}, we consider the negative log-likelihood
\begin{align*}
    -Q^{\text{rep}}(p_{j}^{(k)}(\boldsymbol{x}_i);\mathcal{D}^{(k)})&= \sum_{i=1}^n \sum_{j=1}^d -N^{(k)}_{i, j} \cdot \log \left(p_{j}^{(k)}(\boldsymbol{x}_i)\right)=-\sum_{i=1}^{\overline{n}^{(k)}} w_{i}^{(k)} \sum_{j=1}^d   y^{(k)}_{i, j} \cdot \log \left(p_{j}^{(k)}(\overline{\boldsymbol{x}}^{(k)}_i)\right), 
\end{align*}
as the loss function for the reporting model with 
\begin{align*}
    (p_{1}^{(k)}(\boldsymbol{x}_i),\ldots,p_{d}^{(k)}(\boldsymbol{x}_i)) = \left(\frac{\exp(f_{\text{rep}}^{(k)[1]}(\boldsymbol{x}_{i}))}{\sum_{z=1}^{d} \exp(f_{\text{rep}}^{(k)[z]}(\boldsymbol{x}_{i}))},\ldots,\frac{\exp(f_{\text{rep}}^{(k)[d]}(\boldsymbol{x}_{i}))}{\sum_{z=1}^{d} \exp(f_{\text{rep}}^{(k)[z]}(\boldsymbol{x}_{i}))}  \right)
\end{align*}
to satisfy the sum-to-one constraint for the estimated reporting probabilities.

\paragraph{XGBoost} Algorithm~\ref{algo:XGBrep} details the training algorithm of the reporting model to learn the reporting probabilities $p_j(\boldsymbol{x}_i)$ for $i=1,\ldots,n$ and $j=1,\ldots,d$ when relying on a XGBoost model. For notational purposes, we set $f_{\text{rep}}^{(k)}(\boldsymbol{x}_{i}) \overset{\text{not.}}{=} f_{\text{rep}(T^{(k)}_{\text{rep}})}^{(k)}(\boldsymbol{x}_{i})$ with $T^{(k)}_{\text{rep}}$ denoting the number of iterations used for the XGBoost algorithm in the $k$th iteration of the EM algorithm. After training, the resulting estimated reporting probabilities are expressed as
\begin{align*}
    \widehat{p}_{j}^{(k)}(\boldsymbol{x}_i) = \frac{\exp(\widehat{f}_{\text{rep}}^{(k)[j]}(\boldsymbol{x}_{i}))}{\sum_{z=1}^{d} \exp(\widehat{f}_{\text{rep}}^{(k)[z]}(\boldsymbol{x}_{i}))}.
\end{align*}

\begin{algorithm}[H] \label{algo:XGBrep}
\caption{\textbf{Pseudocode reporting XGBoost model}}

Set $\widehat{f}_{\text{rep}(0)}^{(k)[j]}(\overline{\boldsymbol{x}}^{(k)}_i)=\widehat{f}_{\text{rep}}^{(k-1)[j]}(\overline{\boldsymbol{x}}^{(k)}_i)$ for $i=1,\ldots,\overline{n}$ and $j=1,\ldots,d$. 

\For{$t=1,\ldots,T^{(k)}_{\text{rep}}$}{
  \vspace{0.2cm}
  Calculate the gradient $g_{t,j}(\overline{\boldsymbol{x}}^{(k)}_i)$ and the hessian $h_{t,j}(\overline{\boldsymbol{x}}^{(k)}_i)$ for $i=1,\ldots,\overline{n}$ and $j=1,\ldots,d$:
    \begin{align*}
  g_{t,j}(\overline{\boldsymbol{x}}^{(k)}_i) &= \left[-\frac{\partial Q^{\text{rep}}(p_{i,j}^{(k)}|\mathcal{D}_{\text{train}}^{(k)})}{\partial f_{\text{rep}}^{(k)[j]}(\overline{\boldsymbol{x}}^{(k)}_i)}\right]_{f_{\text{rep}}^{(k)[j]}(\overline{\boldsymbol{x}}^{(k)}_i)=\widehat{f}_{\text{rep}(t-1)}^{(k)[j]}(\overline{\boldsymbol{x}}^{(k)}_i)}\\ &= \frac{\exp(\widehat{f}_{\text{rep}(t-1)}^{(k)[j]}(\overline{\boldsymbol{x}}^{(k)}_i))}{\sum_{z=1}^{d} \exp(\widehat{f}_{\text{rep}(t-1)}^{(k)[z]}(\overline{\boldsymbol{x}}^{(k)}_i))}-\overline{y}_{i,j},\\
  h_{t,j}(\overline{\boldsymbol{x}}^{(k)}_i) &= \left[-\frac{\partial^2 Q^{\text{rep}}(p_{i,j}^{(k)}|\mathcal{D}_{\text{train}}^{(k)})}{\partial f_{\text{rep}}^{(k)[j]}(\overline{\boldsymbol{x}}^{(k)}_i)^2}\right]_{f_{\text{rep}}^{(k)[j]}(\overline{\boldsymbol{x}}^{(k)}_i)=\widehat{f}_{\text{rep}(t-1)}^{(k)[j]}(\overline{\boldsymbol{x}}^{(k)}_i)}\\&= \frac{\exp(\widehat{f}_{\text{rep}(t-1)}^{(k)[j]}(\overline{\boldsymbol{x}}^{(k)}_i))}{\sum_{z=1}^{d} \exp(\widehat{f}_{\text{rep}(t-1)}^{(k)[z]}(\overline{\boldsymbol{x}}^{(k)}_i))}\left(1- \frac{\exp(\widehat{f}_{\text{rep}(t-1)}^{(k)[j]}(\overline{\boldsymbol{x}}^{(k)}_i))}{\sum_{z=1}^{d} \exp(\widehat{f}_{\text{rep}(t-1)}^{(k)[z]}(\overline{\boldsymbol{x}}^{(k)}_i))} \right) .
  \end{align*}

  \For{$j=1,\ldots,d$}{

Learn a regression tree $\delta_{t,j}^{(k)}$ of a fixed size \texttt{tree\_depth\_rep}
  \begin{align*}
  \widehat{\delta}_{t,j}^{(k)}=\underset{\delta_{t,j}^{(k)}}{\operatorname{argmin}} \frac{1}{\overline{n}^{(k)}} \sum_{i=1}^{\overline{n}^{(k)}} \frac{1}{2} h_{t,j}(\overline{\boldsymbol{x}}^{(k)}_i)\left( \delta_{t,j}^{(k)}(\overline{\boldsymbol{x}}^{(k)}_i) - \frac{g_{t,j}(\overline{\boldsymbol{x}}^{(k)}_i)}{h_{t,j}(\overline{\boldsymbol{x}}^{(k)}_i)} \right)^2.
  \end{align*}
  
  Update the predictions:
    \begin{align*}
  \widehat{f}_{\text{rep}(t)}^{(k)[j]}(\boldsymbol{x}_{i})=\widehat{f}_{\text{rep}(t-1)}^{(k)[j]}(\boldsymbol{x}_{i})+\texttt{eta\_rep} \cdot \widehat{\delta}_{t,j}^{(k)}(\boldsymbol{x}_{i}),
  \end{align*}
  where \texttt{eta\_rep} denotes the learning rate.
  
  }

  \If{$t>\texttt{xgb\_patience}$}{

Stop training if the following condition is satisfied:

\begin{align*}
    Q^{\text{rep}}(\widehat{p}_{j}^{(k,t)}(\boldsymbol{x}_{i})|\mathcal{D}_{\text{val},1}^{(k)}) = \displaystyle\max_{z = t-\texttt{xgb\_patience}, \dots, t} Q^{\text{rep}}(\widehat{p}_{j}^{(k,z)}(\boldsymbol{x}_{i})|\mathcal{D}_{\text{val},1}^{(k)}),
\end{align*}
where
\begin{align*}
    \widehat{p}_{j}^{(k,t)}(\boldsymbol{x}_{i})=\frac{\exp(\widehat{f}_{\text{rep}(t)}^{(k)[j]}(\boldsymbol{x}_{i}))}{\sum_{z=1}^{d} \exp(\widehat{f}_{\text{rep}(t)}^{(k)[z]}(\boldsymbol{x}_{i}))}.
\end{align*}
}
}
\end{algorithm}

\begin{figure}[H]
    \centering

    \scalebox{0.9}{

\tikzset{every picture/.style={line width=0.75pt}} 

\begin{tikzpicture}[x=0.75pt,y=0.75pt,yscale=-1,xscale=1]

\draw  [fill={rgb, 255:red, 155; green, 155; blue, 155 }  ,fill opacity=0.56 ] (46,38.81) .. controls (46,25) and (57.19,13.81) .. (71,13.81) .. controls (84.81,13.81) and (96,25) .. (96,38.81) .. controls (96,52.61) and (84.81,63.81) .. (71,63.81) .. controls (57.19,63.81) and (46,52.61) .. (46,38.81) -- cycle ;
\draw  [fill={rgb, 255:red, 155; green, 155; blue, 155 }  ,fill opacity=0.56 ] (46,137.81) .. controls (46,124) and (57.19,112.81) .. (71,112.81) .. controls (84.81,112.81) and (96,124) .. (96,137.81) .. controls (96,151.61) and (84.81,162.81) .. (71,162.81) .. controls (57.19,162.81) and (46,151.61) .. (46,137.81) -- cycle ;
\draw [line width=1.5]  [dash pattern={on 1.69pt off 2.76pt}]  (71,75.81) -- (71.04,104.44) ;
\draw  [fill={rgb, 255:red, 155; green, 155; blue, 155 }  ,fill opacity=0.56 ] (46,227.81) .. controls (46,214) and (57.19,202.81) .. (71,202.81) .. controls (84.81,202.81) and (96,214) .. (96,227.81) .. controls (96,241.61) and (84.81,252.81) .. (71,252.81) .. controls (57.19,252.81) and (46,241.61) .. (46,227.81) -- cycle ;
\draw  [fill={rgb, 255:red, 155; green, 155; blue, 155 }  ,fill opacity=0.56 ] (46,326.81) .. controls (46,313) and (57.19,301.81) .. (71,301.81) .. controls (84.81,301.81) and (96,313) .. (96,326.81) .. controls (96,340.61) and (84.81,351.81) .. (71,351.81) .. controls (57.19,351.81) and (46,340.61) .. (46,326.81) -- cycle ;
\draw [line width=1.5]  [dash pattern={on 1.69pt off 2.76pt}]  (71,264.81) -- (71.04,293.44) ;
\draw  [fill={rgb, 255:red, 248; green, 231; blue, 28 }  ,fill opacity=0.39 ] (165,89.81) .. controls (165,76) and (176.19,64.81) .. (190,64.81) .. controls (203.81,64.81) and (215,76) .. (215,89.81) .. controls (215,103.61) and (203.81,114.81) .. (190,114.81) .. controls (176.19,114.81) and (165,103.61) .. (165,89.81) -- cycle ;
\draw  [fill={rgb, 255:red, 248; green, 231; blue, 28 }  ,fill opacity=0.39 ] (165,279.81) .. controls (165,266) and (176.19,254.81) .. (190,254.81) .. controls (203.81,254.81) and (215,266) .. (215,279.81) .. controls (215,293.61) and (203.81,304.81) .. (190,304.81) .. controls (176.19,304.81) and (165,293.61) .. (165,279.81) -- cycle ;
\draw    (107.4,41.59) -- (156.77,79.25) ;
\draw    (157.57,104.05) -- (107.54,140.84) ;
\draw    (107.4,227.79) -- (156.77,265.45) ;
\draw    (157.57,290.25) -- (107.54,327.04) ;
\draw [line width=1.5]  [dash pattern={on 1.69pt off 2.76pt}]  (190.37,172.59) -- (190.41,201.22) ;
\draw    (107.17,208.74) -- (157.57,119.94) ;
\draw    (157.17,244.34) -- (107.57,157.94) ;
\draw    (107.4,59.07) -- (156.77,218.34) ;
\draw    (107.4,303.07) -- (157.57,145.38) ;
\draw  [fill={rgb, 255:red, 248; green, 231; blue, 28 }  ,fill opacity=0.39 ] (339,89.54) .. controls (339,75.74) and (350.19,64.54) .. (364,64.54) .. controls (377.81,64.54) and (389,75.74) .. (389,89.54) .. controls (389,103.35) and (377.81,114.54) .. (364,114.54) .. controls (350.19,114.54) and (339,103.35) .. (339,89.54) -- cycle ;
\draw  [fill={rgb, 255:red, 248; green, 231; blue, 28 }  ,fill opacity=0.39 ] (339,279.54) .. controls (339,265.74) and (350.19,254.54) .. (364,254.54) .. controls (377.81,254.54) and (389,265.74) .. (389,279.54) .. controls (389,293.35) and (377.81,304.54) .. (364,304.54) .. controls (350.19,304.54) and (339,293.35) .. (339,279.54) -- cycle ;
\draw [line width=1.5]  [dash pattern={on 1.69pt off 2.76pt}]  (365.37,172.59) -- (365.41,201.22) ;
\draw    (227.4,89.82) -- (256.77,89.88) ;
\draw    (227.4,279.82) -- (256.77,279.88) ;
\draw    (227.4,109.42) -- (257.57,248.28) ;
\draw    (256.77,109.22) -- (226.6,248.08) ;
\draw   (46.52,386.34) .. controls (46.54,391.01) and (48.88,393.33) .. (53.55,393.31) -- (59.67,393.29) .. controls (66.34,393.27) and (69.68,395.59) .. (69.7,400.26) .. controls (69.68,395.59) and (73,393.25) .. (79.67,393.23)(76.67,393.24) -- (86.74,393.2) .. controls (91.41,393.19) and (93.73,390.85) .. (93.72,386.18) ;
\draw   (162.28,385.94) .. controls (162.27,390.61) and (164.6,392.94) .. (169.27,392.95) -- (259.49,393.08) .. controls (266.16,393.09) and (269.49,395.42) .. (269.48,400.09) .. controls (269.49,395.42) and (272.82,393.1) .. (279.49,393.11)(276.49,393.1) -- (383.6,393.25) .. controls (388.27,393.26) and (390.6,390.93) .. (390.61,386.26) ;
\draw  [dash pattern={on 4.5pt off 4.5pt}] (41,10.65) -- (100.77,10.65) -- (100.77,166.58) -- (41,166.58) -- cycle ;
\draw  [dash pattern={on 4.5pt off 4.5pt}] (41,199.58) -- (100.77,199.58) -- (100.77,355.51) -- (41,355.51) -- cycle ;
\draw  [fill={rgb, 255:red, 184; green, 233; blue, 134 }  ,fill opacity=0.55 ] (444,89.54) .. controls (444,75.74) and (455.19,64.54) .. (469,64.54) .. controls (482.81,64.54) and (494,75.74) .. (494,89.54) .. controls (494,103.35) and (482.81,114.54) .. (469,114.54) .. controls (455.19,114.54) and (444,103.35) .. (444,89.54) -- cycle ;
\draw  [fill={rgb, 255:red, 184; green, 233; blue, 134 }  ,fill opacity=0.55 ] (444,279.54) .. controls (444,265.74) and (455.19,254.54) .. (469,254.54) .. controls (482.81,254.54) and (494,265.74) .. (494,279.54) .. controls (494,293.35) and (482.81,304.54) .. (469,304.54) .. controls (455.19,304.54) and (444,293.35) .. (444,279.54) -- cycle ;
\draw [line width=1.5]  [dash pattern={on 1.69pt off 2.76pt}]  (470.37,172.59) -- (470.41,201.22) ;
\draw    (402.4,89.82) -- (431.77,89.88) ;
\draw    (402.4,279.82) -- (431.77,279.88) ;
\draw    (402.4,109.42) -- (432.57,248.28) ;
\draw    (431.77,109.22) -- (401.6,248.08) ;
\draw   (446.86,386.34) .. controls (446.87,391.01) and (449.21,393.33) .. (453.88,393.31) -- (460.01,393.29) .. controls (466.68,393.27) and (470.02,395.59) .. (470.03,400.26) .. controls (470.02,395.59) and (473.34,393.25) .. (480.01,393.23)(477.01,393.24) -- (487.08,393.2) .. controls (491.75,393.19) and (494.07,390.85) .. (494.05,386.18) ;
\draw    (295.4,89.82) -- (324.77,89.88) ;
\draw    (295.4,279.82) -- (324.77,279.88) ;
\draw    (295.4,109.42) -- (325.57,248.28) ;
\draw    (324.77,109.22) -- (294.6,248.08) ;
\draw [line width=1.5]  [dash pattern={on 1.69pt off 2.76pt}]  (283.17,89.99) -- (267.58,90.05) ;
\draw [line width=1.5]  [dash pattern={on 1.69pt off 2.76pt}]  (283.17,279.99) -- (267.58,280.05) ;

\draw (33,400) node [anchor=north west][inner sep=0.75pt]  [font=\normalsize] [align=left] {Input layer};
\draw (225,400) node [anchor=north west][inner sep=0.75pt]  [font=\normalsize] [align=left] {Hidden layers};
\draw (3,268.13) node [anchor=north west][inner sep=0.75pt] [font=\Large]  [align=left] {$\overline{\boldsymbol{x}}^{(k)}_{\text{per}}$};
\draw (3,77.53) node [anchor=north west][inner sep=0.75pt] [font=\Large] [align=left]{$\overline{\boldsymbol{x}}^{(k)}_{\text{ent}}$};
\draw (423,400) node [anchor=north west][inner sep=0.75pt]  [font=\normalsize] [align=left] {Output layer};
\draw (176,77) node [anchor=north west][inner sep=0.75pt] [font=\large] [align=left] {$\theta_{1,1}^{(k)}$};
\draw (348,77) node [anchor=north west][inner sep=0.75pt] [font=\large] [align=left] {$\theta_{M,1}^{(k)}$};
\draw (344,267) node [anchor=north west][inner sep=0.75pt] [font=\large] [align=left] {$\theta_{M,q_{M}}^{(k)}$};
\draw (173,267) node [anchor=north west][inner sep=0.75pt] [font=\large] [align=left] {$\theta_{1,q_{1}}^{(k)}$};
\draw (446,77) node [anchor=north west][inner sep=0.75pt] [font=\normalsize] [align=left] {$\theta_{M+1,1}^{(k)}$};
\draw (446,267) node [anchor=north west][inner sep=0.75pt] [font=\normalsize] [align=left] {$\theta_{M+1,d}^{(k)}$};

\end{tikzpicture}
    
}
    
    \caption{Network architecture for the reporting model in the $k$th iteration of the EM algorithm. The grey nodes represent the input neurons, the yellow nodes represent the neurons $\theta_{m,u}^{(k)}(\boldsymbol{z})$ for $m=1,\ldots,M$ and $u=1,\ldots,q_m$ in the $M$ hidden layers. The output layer consists of $d$ neurons $\theta_{M+1,1}^{(k)},\ldots,\theta_{M+1,d}^{(k)}$, represented by the green nodes. }
    \label{fig:NNrep}
\end{figure}
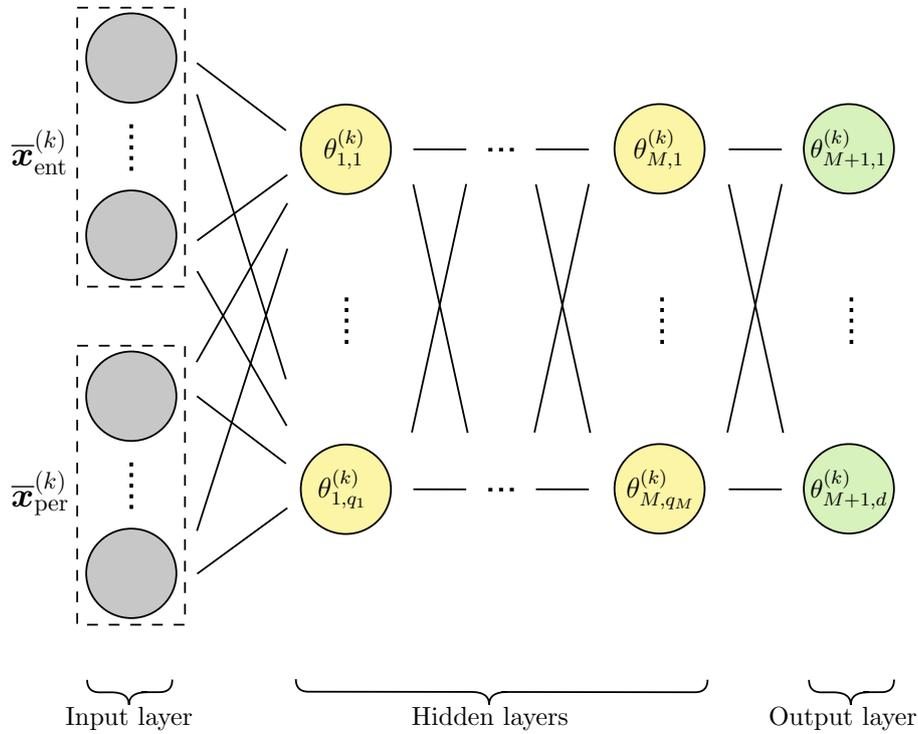

\paragraph{Neural network} 
Figure~\ref{fig:NNrep} shows the network architecture used when learning the reporting probabilities $p_j(\boldsymbol{x}_i)$ for $i=1,\ldots,n$ and $j=1,\ldots,d$. We use the same architecture, notation and training algorithm as for the occurrence model, with the difference being that the output layer becomes $d$-dimensional instead of one dimensional. After training, the resulting estimated reporting probabilities are expressed as
\begin{align*}
    \widehat{p}_{j}^{(k)}(\boldsymbol{x}_i) =\widehat{\theta}^{(k)}_{M+1,j}(\boldsymbol{x}_i)&= \frac{\exp(\widehat{f}_{\text{rep}}^{(k)[j]}(\boldsymbol{x}_{i}))}{\sum_{z=1}^{d} \exp(\widehat{f}_{\text{rep}}^{(k)[z]}(\boldsymbol{x}_{i}))}\\&=\frac{\exp\left(\widehat{b}^{(k)}_{M+1,j}+\left\langle\widehat{\boldsymbol{w}}^{(k)}_{M+1,j},\left(\widehat{\theta}^M \circ \cdots \circ \widehat{\theta}^1\right)(\boldsymbol{x}_i)\right\rangle\right)}{\sum_{z=1}^{d} \exp\left(\widehat{b}^{(k)}_{M+1,z}+\left\langle\widehat{\boldsymbol{w}}^{(k)}_{M+1,z},\left(\widehat{\theta}^M \circ \cdots \circ \widehat{\theta}^1\right)(\boldsymbol{x}_i)\right\rangle\right)}.
\end{align*}
In Figure~\ref{fig:NNrep}, we visualise the network architecture to learn the occurrence intensities as described in Section~\ref{sec:method}.

\pagebreak

\section{Simulation experiments } \label{app:figures} 
\subsection{Data generation}
To simulate the occurrence- and reporting period-related information, we chose the period between 11 April 2022 and 11 May 2022 to simulate day $1,\ldots,31$, with 1 May 2022 between the last observable day, i.e. day 21. Consequently, the weekend day indicator $x^{\text{(ps)}}_{1}$ and the indicator for the first and last day of the calender month $x^{\text{(ps)}}_{3}$ are simulated according to that period. The holiday indicator $x^{\text{(ps)}}_{2}$ matches with the official Dutch holidays in the same period. We simulate the entity-specific information, i.e. the variables $x^{\text{(es)}}_{1}$, $x^{\text{(es)}}_{2}$ and $x^{\text{(es)}}_{3}$, from (separate) uniform distributions.

Table~\ref{tab:spec1} presents the coefficient vectors $\boldsymbol{\beta}_{\lambda}$ and  $\boldsymbol{\beta}_{p_1},\ldots,\boldsymbol{\beta}_{p_{11}}$ used to simulate the occurrence intensities $\lambda(\boldsymbol{x}_i)$ and reporting probabilities $p_1(\boldsymbol{x}_i),\ldots,p_{11}(\boldsymbol{x}_i)$ in the experiment including only linear effects. Variables not explicitly shown in Table~\ref{tab:spec1} are simulated to have no effect. For example, for $x_1^{(ps,j)}$ with $j=2,\ldots,11$, all corresponding values in the coefficient vector $\boldsymbol{\beta}_{p_1}$ are set to zero.

\begin{table}[H]
\hspace{-3em}

\begin{adjustbox}{width=\textwidth}
    \begin{tabular}{ccccccccccccc}
  \hline
 &$\boldsymbol{\beta}_{\lambda}$&$\boldsymbol{\beta}_{p_1}$&$\boldsymbol{\beta}_{p_2}$&$\boldsymbol{\beta}_{p_3}$&$\boldsymbol{\beta}_{p_4}$&$\boldsymbol{\beta}_{p_5}$&$\boldsymbol{\beta}_{p_6}$&$\boldsymbol{\beta}_{p_7}$&$\boldsymbol{\beta}_{p_8}$&$\boldsymbol{\beta}_{p_9}$&$\boldsymbol{\beta}_{p_{10}}$&$\boldsymbol{\beta}_{p_{11}}$\\ [0.1em]
 \hline
 1&0&2&1.8&1.6&1.4&1.2&1&0.8&0.6&0.4&0.2&0\\
 $x_{1,1}^{(es)}$&0&0&0&0&0&0&0&0&0&0&0&0\\  
 \vspace{0.3em}
 $x_{1,2}^{(es)}$&0.5&0&0&0&0&0&0&0&0&0&0&0\\  
 \vspace{0.3em}
 $x_2^{(es)}$&0.01&0.001&0.002&0.003&0.004&0.005&0.006&0.007&0.008&0.009&0.010&0.011\\   
 \vspace{0.3em}
 $x_{3,1}^{(es)}$&0&0&0&0&0&0&0&0&0&0&0&0\\  
 \vspace{0.3em}
 $x_{3,2}^{(es)}$&0.2&-0.02&-0.025&-0.03&-0.035&-0.04&-0.045&-0.05&-0.055&-0.06&-0.065&-0.07\\   
 \vspace{0.3em}
 $x_{3,3}^{(es)}$&0.4&-0.05&-0.055&-0.06&-0.065&-0.07&-0.075&-0.08&-0.085&-0.09&-0.095&-0.1\\  
 \vspace{0.3em}
 $x_{4,1}^{(es)}$&0.3&-0.1&-0.1&-0.1&-0.1&-0.1&0&0&0&0&0&0\\  
 \vspace{0.3em}
 $x_{4,2}^{(es)}$&0&0&0&0&0&0&0&0&0&0&0&0\\   
 \vspace{0.3em}  
 $x_{4,3}^{(es)}$&-0.3&0.2&0.2&0.2&0.2&0.2&0&0&0&0&0&0\\ 
 \vspace{0.3em}
$x_1^{(ps,j)}$&0.75&-0.2&-0.2&-0.2&-0.2&-0.2&-0.2&-0.2&-0.2&-0.2&-0.2&-0.2\\  
\vspace{0.3em}
$x_2^{(ps,j)}$&0.4&-0.3&-0.3&-0.3&-0.3&-0.3&-0.3&-0.3&-0.3&-0.3&-0.3&-0.3\\   
\vspace{0.3em} 
$x_3^{(ps,j)}$&0&0.05&0.05&0.05&0.05&0.05&0.05&0.05&0.05&0.05&0.05&0.05\\  
\hline
&$j=1$&$j=1$&$j=2$&$j=3$&$j=4$&$j=5$&$j=6$&$j=7$&$j=8$&$j=9$&$j=10$&$j=11$\\
\hline 
 
\end{tabular}
\end{adjustbox}

\caption{Coefficient specification for the simulation experiment with only linear effects. The first column shows the covariates for which a coefficient is specified. The second column gives the parameter values for the occurrence intensities $\lambda(\boldsymbol{x}_i)$, i.e. $\boldsymbol{\beta}_{\lambda}$. The second to twelfth column indicate the coefficient vectors $\boldsymbol{\beta}_{p_1},\ldots,\boldsymbol{\beta}_{p_{11}}$ for the reporting probabilities $p_1(\boldsymbol{x}_i),\ldots,p_{11}(\boldsymbol{x}_i)$, respectively. }
\label{tab:spec1}
\end{table}

Table~\ref{tab:spec2} presents the coefficient vectors $\beta_{\lambda}$ and  $\beta_{p_1},\ldots,\beta_{p_{11}}$ used to simulate the occurrence intensities $\lambda(\boldsymbol{x}_i)$ and reporting probabilities $p_1(\boldsymbol{x}_i),\ldots,p_{11}(\boldsymbol{x}_i)$ in the experiment including both linear and non-linear effects. Similar to the first experiment, variables not explicitly shown in Table~\ref{tab:spec2} are simulated to have no effect.
\begin{table}[H]
\hspace{-5.75em}

\begin{adjustbox}{width=\textwidth}
    \begin{tabular}{ccccccccccccc}
  \hline
 &$\boldsymbol{\beta}_{\lambda}$&$\boldsymbol{\beta}_{p_1}$&$\boldsymbol{\beta}_{p_2}$&$\boldsymbol{\beta}_{p_3}$&$\boldsymbol{\beta}_{p_4}$&$\boldsymbol{\beta}_{p_5}$&$\boldsymbol{\beta}_{p_6}$&$\boldsymbol{\beta}_{p_7}$&$\boldsymbol{\beta}_{p_8}$&$\boldsymbol{\beta}_{p_9}$&$\boldsymbol{\beta}_{p_{10}}$&$\boldsymbol{\beta}_{p_{11}}$\\ [0.1em]
 \hline
 1&0&2&1.8&1.6&1.4&1.2&1&0.8&0.6&0.4&0.2&0\\  
 $x_{1,1}^{(es)}$&0&0&0&0&0&0&0&0&0&0&0&0\\  
 \vspace{0.3em}
 $x_{1,2}^{(es)}$&0.5&0&0&0&0&0&0&0&0&0&0&0\\  
 \vspace{0.3em}
 $\log{x_2^{(es)}+1}$&0.01&0.001&0.002&0.003&0.004&0.005&0.006&0.007&0.008&0.009&0.010&0.011\\   
 \vspace{0.3em}
 $x_{3,1}^{(es)}$&0&0&0&0&0&0&0&0&0&0&0&0\\  
 \vspace{0.3em}
 $x_{3,2}^{(es)}$&0.2&-0.02&-0.025&-0.03&-0.035&-0.04&-0.045&-0.05&-0.055&-0.06&-0.065&-0.07\\   
 \vspace{0.3em}
 $x_{3,3}^{(es)}$&0.4&-0.05&-0.055&-0.06&-0.065&-0.07&-0.075&-0.08&-0.085&-0.09&-0.095&-0.1\\  
 \vspace{0.3em}
 $x_{4,1}^{(es)}$&0.3&-0.1&-0.1&-0.1&-0.1&-0.1&0&0&0&0&0&0\\  
 \vspace{0.3em}
 $x_{4,2}^{(es)}$&0&0&0&0&0&0&0&0&0&0&0&0\\   
 \vspace{0.3em}  
 $x_{4,3}^{(es)}$&-0.3&0.2&0.2&0.2&0.2&0.2&0&0&0&0&0&0\\ 
 \vspace{0.3em}
$x_1^{(ps,j)}$&0.75&-0.2&-0.2&-0.2&-0.2&-0.2&-0.2&-0.2&-0.2&-0.2&-0.2&-0.2\\  
\vspace{0.3em}
$x_2^{(ps,j)}$&0.4&-0.3&-0.3&-0.3&-0.3&-0.3&-0.3&-0.3&-0.3&-0.3&-0.3&-0.3\\   
\vspace{0.3em} 
$x_3^{(ps,j)}$&0&0.05&0.05&0.05&0.05&0.05&0.05&0.05&0.05&0.05&0.05&0.05\\  
\vspace{0.3em}
 $\mathbbm{1}(x_2^{(es)} > 80) \cdot x_2^{(es)}$&0.5&0&0&0&0&0&0&0&0&0&0&0\\   
 \vspace{0.3em}
 $\mathbbm{1}(x_2^{(es)} < 40) \cdot x_{1,1}^{(es)}$&-0.25&0&0&0&0&0&0&0&0&0&0&0\\  
 \vspace{0.3em}
 $x_{3,1}^{(es)} \cdot x_{4,3}^{(es)}$&-0.5&0&0&0&0&0&0&0&0&0&0&0\\ 
 \vspace{0.3em}
 $x_{3,3}^{(es)} \cdot x_{4,1}^{(es)}$&0.5&0&0&0&0&0&0&0&0&0&0&0\\  
 \vspace{0.3em}
 $x_{1,1}^{(es)} \cdot x_{4,3}^{(es)}$&-0.3&0.03&0.03&0.03&0&0&0&0&0&0&0&0\\ 
 \vspace{0.3em}
 $x_1^{(ps,j)} \cdot x_2^{(ps,j)}$&0&0.06&0.06&0.06&0.06&0.06&0&0&0&0&0&0\\ 
\hline 
&$j=1$&$j=1$&$j=2$&$j=3$&$j=4$&$j=5$&$j=6$&$j=7$&$j=8$&$j=9$&$j=10$&$j=11$\\
\hline 
\end{tabular}
\end{adjustbox}

\caption{Coefficient specification for the simulation experiment with both linear and non-linear effects. The first column shows the covariates for which a coefficient is specified. The second column gives the parameter values for the occurrence intensities $\lambda(\boldsymbol{x}_i)$, i.e. $\boldsymbol{\beta}_{\lambda}$. The second to twelfth column indicate the coefficient vectors $\boldsymbol{\beta}_{p_1},\ldots,\boldsymbol{\beta}_{p_{11}}$ for the reporting probabilities $p_1(\boldsymbol{x}_i),\ldots,p_{11}(\boldsymbol{x}_i)$, respectively. }
\label{tab:spec2}
\end{table}

\subsection{Generalised linear model specification}

We use the same model specification in both experimental settings. In the setting with only linear effects, the model is correctly specified: although we include some redundant variables, we do not omit any variable with a non-zero coefficient. In the setting with linear and non-linear effects, the model is misspecified since we do not include higher order terms, because we do not know a priori which higher order terms are relevant and there are infinite options to include. For the occurrence GLM, we use the following specification:
\begin{align*}
    f^{\text{glm}}_{\text{occ}}(\boldsymbol{x})&= \beta_{0} + \beta^{(es)}_{1} x_{1,1}^{(es)}+\beta^{(es)}_{2}x_{2}^{(es)}+ \beta^{(es)}_{3}x_{3,2}^{(es)} + \beta^{(es)}_{4}x_{3,3}^{(es)} +\beta^{(es)}_{5}x_{4,2}^{(es)} +\beta^{(es)}_{6}x_{4,3}^{(es)}\\
    &+\beta^{(ps,1)}_{1}x_{1}^{(ps,1)}
    +\beta^{(ps,1)}_{2}x_{2}^{(ps,1)}
    +\beta^{(ps,1)}_{3}x_{3}^{(ps,1)}.
\end{align*}
For the $j$th reporting probability, we use the following specification:
\begin{align*}
    f^{\text{glm}[j]}_{\text{rep}}(\boldsymbol{x})&= \beta_{0} + \beta^{(es)}_{1} x_{1,1}^{(es)}+\beta^{(es)}_{2}x_{2}^{(es)}+ \beta^{(es)}_{3}x_{3,2}^{(es)} + \beta^{(es)}_{4}x_{3,3}^{(es)} +\beta^{(es)}_{5}x_{4,2}^{(es)} +\beta^{(es)}_{6}x_{4,3}^{(es)}\\
    &+\sum_{z=1}^{11}\beta^{(ps,z)}_{1}x_{1}^{(ps,z)}
    +\beta^{(ps,z)}_{2}x_{2}^{(ps,z)}
    +\beta^{(ps,z)}_{3}x_{3}^{(ps,z)}.
\end{align*}

\subsection{Tuning setup}

Tables~\ref{tab:tuning1} and \ref{tab:tuning2} show the chosen tuning values for the occurrence and de reporting model in the experiment with only linear effects and the experiment with both linear and non-linear effects, respectively.

\begin{table}[H]
\centering
\begin{adjustbox}{width=0.9\textwidth}
    \begin{tabular}{m{12.5em} m{7em} m{0.3em} m{7.8em}m{7.5em}}
  \hline

 &  && \multicolumn{2}{c}{Optimal setting}\\
 \cline{4-5} 
 &Tuning values&&Occurrence model&Reporting model\\
 \hline
 \textbf{XGBoost}&&&&\\
 \quad \texttt{eta\_occ}&$[0.01,0.05,0.1]$&&0.1&/\\
 \quad \texttt{eta\_rep}& $[0.001,0.005,0.01]$&&/&0.01\\
 \quad $T^{(1)}$&$[20,40]$&&20&20\\
 \quad $T^{(k)}$ with $k=2,\ldots,K$&$[10,20,40]$&&10&20\\
 \quad \texttt{tree\_depth}&$[3,5,7]$&&3&3\\
 \quad \texttt{xgb\_patience}&$[15,30,45]$&&15&30\\

 \textbf{Neural network}&&&&\\
 \quad $q_{1}$&$[5,10,15]$&&10&5\\
 \quad $q_{2}$&$[5,10,15]$&&10&15\\
 \quad \texttt{learning\_rate}&$[0.00005,0.0001,\newline0.0005,0.001,0.005]$&&0.00005&0.0001\\
 \quad \texttt{n\_epoch}&$[50]$&&50&50\\
 \quad \texttt{batch\_size}&$[32,64,128,256]$&&32&32\\
 \quad \texttt{nn\_patience}&$[5,10,15]$&&10&10\\

\hline 
 
\end{tabular}
\end{adjustbox}

\caption{Tuning parameter choices for the experiment with only linear effects. The first and second column show the tuning parameters and their considered values, respectively. The third column indicates the chosen values for the occurrence model, while the fourth column shows the values chosen for the reporting model.}
\label{tab:tuning1}
\end{table}

\begin{table}[H]
\centering
\begin{adjustbox}{width=0.9\textwidth}
    \begin{tabular}{m{12.5em} m{7em} m{0.3em} m{7.8em}m{7.5em}}
  \hline

 &  && \multicolumn{2}{c}{Optimal setting}\\
 \cline{4-5} 
 &Tuning values&&Occurrence model&Reporting model\\
 \hline
 \textbf{XGBoost}&&&&\\
 \quad \texttt{eta\_occ}&$[0.01,0.05,0.1]$&&0.05&/\\
 \quad \texttt{eta\_rep}& $[0.001,0.005,0.01]$&&/&0.01\\
 \quad $T^{(1)}$&$[20,40]$&&20&20\\
 \quad $T^{(k)}$ with $k=2,\ldots,K$&$[10,20,40]$&&40&10\\
 \quad \texttt{tree\_depth}&$[3,5,7]$&&3&3\\
 \quad \texttt{xgb\_patience}&$[15,30,45]$&&15&15\\

 \textbf{Neural network}&&&&\\
 \quad $q_{1}$&$[5,10,15]$&&5&15\\
 \quad $q_{2}$&$[5,10,15]$&&5&10\\
 \quad \texttt{learning\_rate}&$[0.00005,0.0001,\newline0.0005,0.001,0.005]$&&0.005&0.0001\\
 \quad \texttt{n\_epoch}&$[50]$&&50&50\\
 \quad \texttt{batch\_size}&$[32,64,128,256]$&&64&32\\
 \quad \texttt{nn\_patience}&$[5,10,15]$&&15&5\\

\hline 
 
\end{tabular}
\end{adjustbox}
\caption{Tuning parameter choices for the experiment with both linear as well as non-linear effects. The first and second column show the tuning parameters and their considered values, respectively. The third column indicates the chosen values for the occurrence model, while the fourth column shows the values chosen for the reporting model.}
\label{tab:tuning2}
\end{table}

\section{Application to real data} \label{app:covid}
\begin{table}[H]
\centering
\begin{tabular}{m{12.5em} m{7em} m{0.3em} m{7.8em}m{7.5em}}
  \hline

 &  && \multicolumn{2}{c}{Optimal setting}\\
 \cline{4-5} 
 &Tuning values&&Occurrence model&Reporting model\\
 \hline
 \textbf{XGBoost}&&&&\\
 \quad \texttt{eta\_occ}&$[0.01,0.05,0.1]$&&0.05&/\\
 \quad \texttt{eta\_rep}& $[0.001,0.005,0.01]$&&/&0.1\\
 \quad $T^{(1)}$&$[20,40]$&&20&20\\
 \quad $T^{(k)}$ with $k=2,\ldots,K$&$[10,20,40]$&&40&10\\
 \quad \texttt{tree\_depth}&$[3,5,7]$&&3&5\\
 \quad \texttt{xgb\_patience}&$[15,30,45]$&&30&30\\

 \textbf{Neural network}&&&&\\
 \quad $q_{1}$&$[5,10,15]$&&5&15\\
 \quad $q_{2}$&$[5,10,15]$&&5&10\\
 \quad \texttt{learning\_rate}&$[0.00005,0.0001,\newline0.0005,0.001,0.005]$&&0.0001&0.0005\\
 \quad \texttt{n\_epoch}&$[50]$&&50&50\\
 \quad \texttt{batch\_size}&$[32,64,128,256]$&&32&64\\
 \quad \texttt{nn\_patience}&$[5,10,15]$&&15&10\\

\hline 
 
\end{tabular}
\caption{Tuning parameter choices for the application to Argentinian Covid-19 data. The first and second column show the tuning parameters and their considered values, respectively. The third column indicates the chosen values for the occurrence model, while the fourth column shows the values chosen for the reporting model.}
\label{tab:covidtuning}
\end{table}

\begin{figure}[H] 
\centering
\subfloat[Tuesday]{

\begin{adjustbox}{width=0.45\textwidth}
    \includegraphics[width = \textwidth]{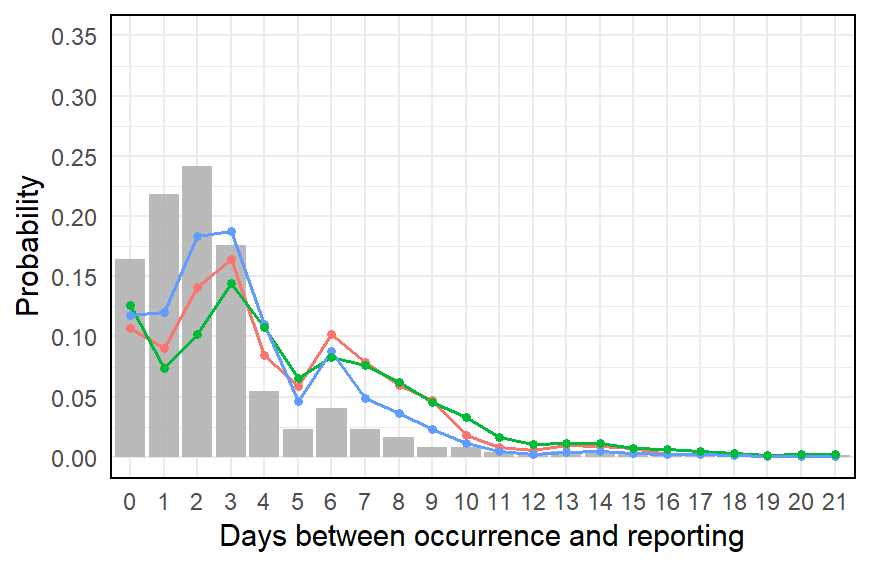}
\end{adjustbox}

}
\hspace{1.5em}
\subfloat[Friday]{

\begin{adjustbox}{width=0.45\textwidth}
\includegraphics[width = \textwidth]{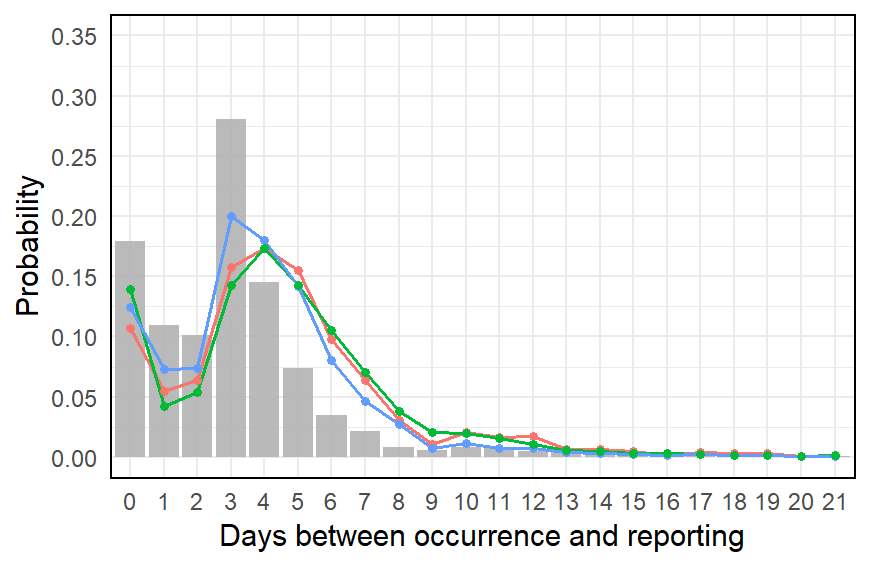}

\end{adjustbox}

}
\caption{Out-of-sample reporting probabilities for cases with symptom onset on a Tuesday (left panel) and Friday (right panel). The grey bars show the empirical probabilities in the test set. The red, green and blue lines correspond to the estimated reporting probabilities from the GLMs, neural networks and XGBoost models, respectively.}   
\label{fig:coviddayofweek}
\end{figure}

\begin{landscape}
    \begin{figure}
        \centering
        \includegraphics[width = \linewidth]{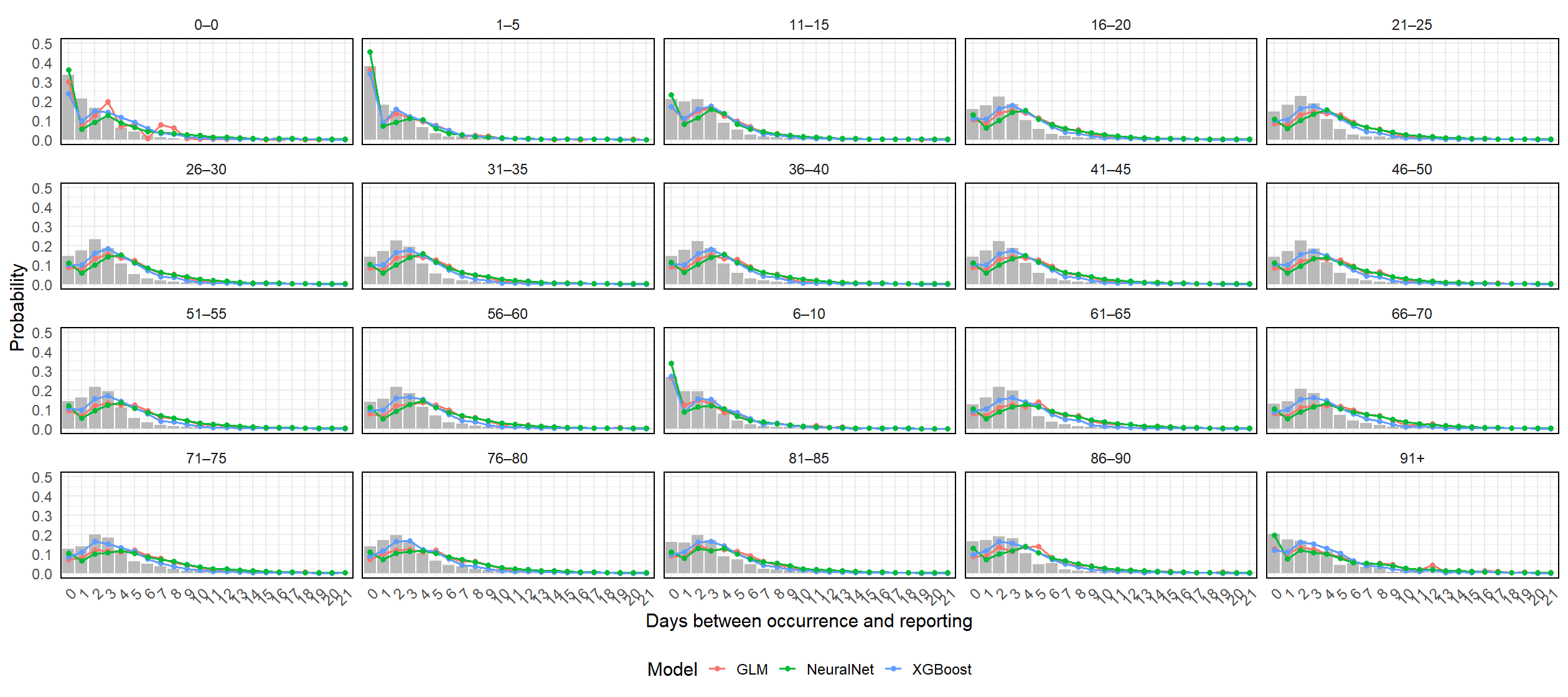}
        \caption{Out-of-sample reporting probabilities for different age brackets. The grey bars show the empirical probabilities in the test set. The red, green and blue lines correspond to the estimated reporting probabilities from the GLMs, neural networks and XGBoost models, respectively.}
        \label{fig:covidage_full}
    \end{figure}
\end{landscape}

\begin{landscape}
    \begin{figure}
        \centering
        \includegraphics[width = \linewidth]{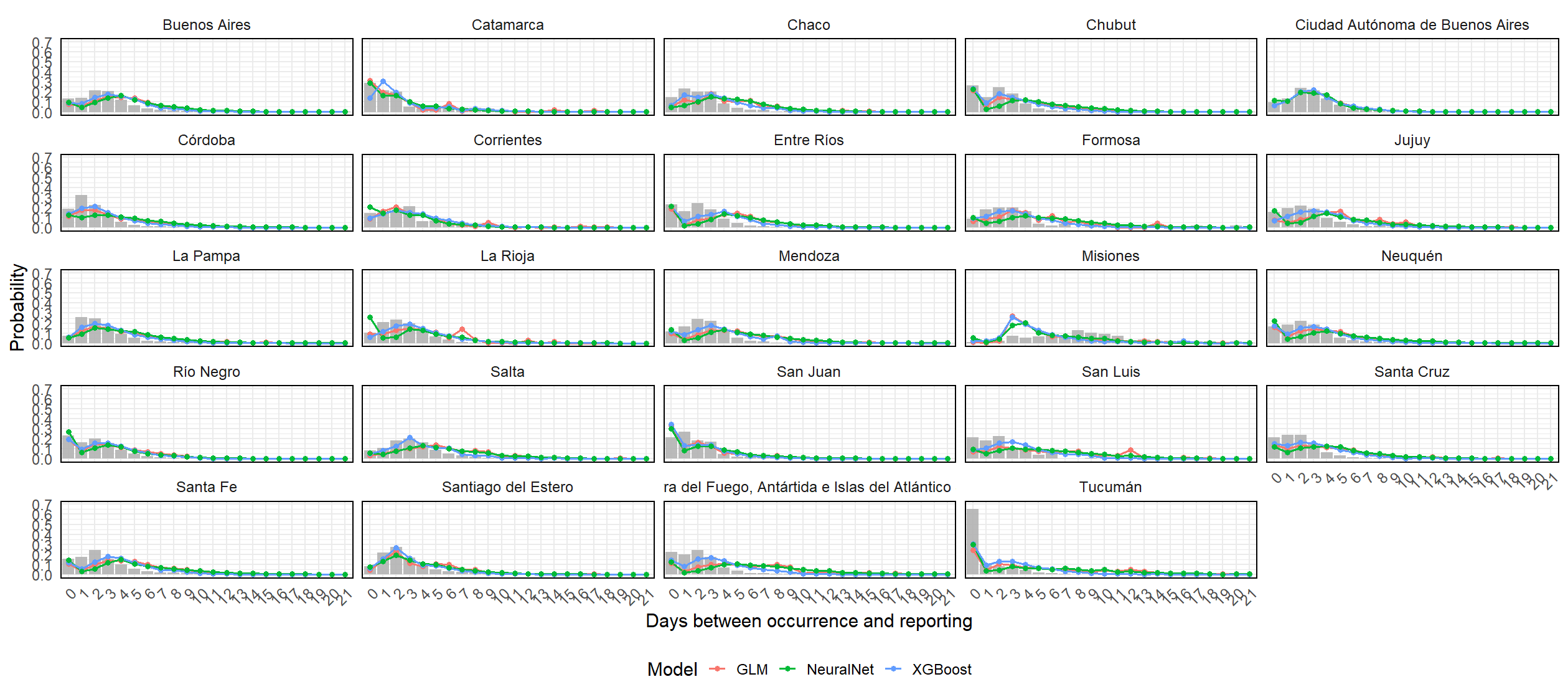}
        \caption{Out-of-sample reporting probabilities across different regions in Argentina. The grey bars show the empirical probabilities in the test set. The red, green and blue lines correspond to the estimated reporting probabilities from the GLMs, neural networks and XGBoost models, respectively.}
        \label{fig:covidlocation_full}
    \end{figure}
\end{landscape}

\end{document}